\documentclass{article}

\usepackage{microtype}
\usepackage{graphicx}
\usepackage{booktabs} 

\usepackage{hyperref}

\usepackage{icml2025}

\usepackage{amsmath}
\usepackage{amssymb}
\usepackage{mathtools}
\usepackage{amsthm}

\usepackage[capitalize,noabbrev]{cleveref}

\usepackage{amsmath,amsfonts,bm}

\def\Figref#1{Fig.~\ref{#1}}

\def\Secref#1{Sec.~\ref{#1}}

\def\eqref#1{eq.~\ref{#1}}
\def\Eqref#1{Eq.~\ref{#1}}

\def\1{\bm{1}}

\DeclareMathAlphabet{\mathsfit}{\encodingdefault}{\sfdefault}{m}{sl}
\SetMathAlphabet{\mathsfit}{bold}{\encodingdefault}{\sfdefault}{bx}{n}

\usepackage{hyperref}
\usepackage{cleveref}
\Crefname{lemma}{Lemma}{Lemmas}  
\usepackage{url}
\usepackage{subcaption}  
\usepackage{amsmath}
\usepackage{amsfonts}
\usepackage{amssymb}
\usepackage{hyperref}
\usepackage{float}
 
\usepackage{amsmath,amsthm}

\usepackage{lipsum} 
\newtheorem{lemma}{Lemma}
\newtheorem{assumption}{Assumption}

\usepackage{array}
\usepackage{multirow}

\usepackage[utf8]{inputenc} %
\usepackage[T1]{fontenc}    %
\usepackage{hyperref}       %
\usepackage{url}            %
\usepackage{booktabs}       %
\usepackage{amsfonts}       %
\usepackage{nicefrac}       %
\usepackage{microtype}      %
\usepackage{xcolor}         %
\usepackage{lipsum}
\usepackage{graphicx}
\usepackage{wrapfig}
\usepackage[size=small]{caption}
\usepackage{mathtools}
\usepackage{amssymb}
\usepackage{amsthm}
\usepackage{soul}
 
\usepackage[inline]{enumitem}

  \renewcommand{\P}{{\cal {P}}}

\newcommand{\prompt}{x^{(i)}}
\newcommand{\win}{y_w^{(i)}}
\newcommand{\lose}{y_l^{(i)}}
\newcommand{\actionspace}{\mathcal{Y}}

\newcommand{\piref}{\pi_\text{ref}}

\newcommand{\eqdef}{\stackrel{\rm def}{=}}

\theoremstyle{plain}

\newtheorem{proposition}{Proposition}
\theoremstyle{definition}

\theoremstyle{remark}

\usepackage[textsize=tiny]{todonotes}

\icmltitlerunning{Simultaneous Reward Distillation and Preference Learning}

\begin{document}

\twocolumn[
\icmltitle{Simultaneous Reward Distillation and Preference Learning: Get You a Language Model Who Can Do Both}



\icmlsetsymbol{equal}{*}

\begin{icmlauthorlist}
    \icmlauthor{Abhijnan Nath}{csu}
    \icmlauthor{Changsoo Jung}{csu}
    \icmlauthor{Ethan Seefried}{csu}
    \icmlauthor{Nikhil Krishnaswamy}{csu}
\end{icmlauthorlist}

\icmlaffiliation{csu}{Situated Grounding and Natural Language (SIGNAL) Lab, Colorado State University, Fort Collins, CO, USA}

\icmlcorrespondingauthor{Abhijnan Nath and Nikhil Krishnaswamy }{{abhijnan.nath, nikhil.krishnaswamy}@colostate.edu}

\icmlkeywords{Machine Learning, ICML}

\vskip 0.3in
]



\printAffiliationsAndNotice{} 
\begin{abstract}

Traditional RLHF-based LLM alignment methods explicitly maximize the expected rewards from a separate reward model. More recent supervised alignment methods like Direct Preference Optimization (DPO) circumvent this phase to avoid problems including model drift and reward overfitting. Although popular due to its simplicity, DPO and similar direct alignment methods which rely heavily on the Bradley-Terry-based pairwise preference formulation can still lead to degenerate policies when challenged by non-deterministic or noisy preference labels, for example human scoring of two candidate outputs with low confidence. This paper introduces \textbf{DRDO (Direct Reward Distillation and policy-Optimization)}, which simultaneously models rewards \textit{and} preferences to avoid such degeneracy. DRDO directly mimics rewards assigned by an oracle while learning human preferences with a novel preference likelihood formulation. Results on the Ultrafeedback and TL;DR datasets demonstrate that DRDO-trained policies surpass methods such as DPO and e-DPO in terms of expected rewards and are more robust, on average, to noisy preference signals as well as out-of-distribution (OOD) settings.
\end{abstract}

\vspace*{-2mm}
\section{Introduction}
\label{sec:intro}
\vspace*{-2mm}

\begin{figure*}[h]
    \centering
    \includegraphics[width=\textwidth]{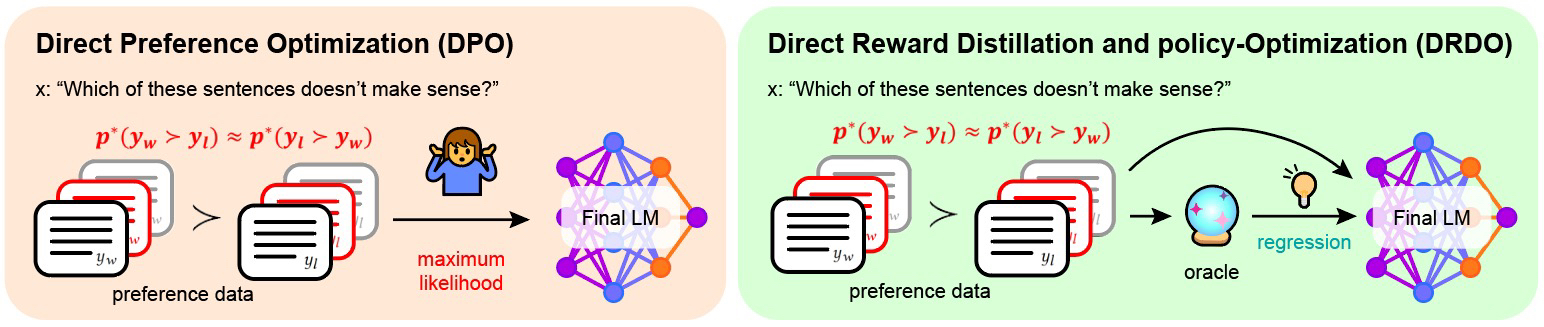}
    \vspace*{-2mm}
    \caption{Unlike popular supervised preference alignment algorithms like Direct Preference Optimization (DPO;~\citet{rafailov2024direct}) that learns rewards implicitly, \textbf{DRDO directly optimizes for explicit rewards from an Oracle while simultaneously learning diverse kinds of preference signals during alignment.} Optimized with a simple regression loss based on difference of rewards assigned by the Oracle and the introduction of a focal-log-unlikelihood component (see~\Secref{sec:drdo}), DRDO avoids DPO's particular challenges at learning non-deterministic preference pairs, thereby bridging the gap between the preference distribution estimated from the data and the true preference distribution $p^*$. Additionally, DRDO does not require an additional reference model during training and can leverage reward signals even when preference labels are not directly accessible.}
    \label{fig:DPO_DRDO}
    \vspace*{-2mm}
\end{figure*}

In the development of large language models (LLMs), robust modeling of human preferences is essential for producing models that users actually find useful. Inherent in this problem is that while popular approaches like Direct Preference Optimization (DPO) and its variants implicitly assume that pairs of preferred and dispreferred samples in preference data have an unambiguous winner, this does not reflect the reality of actual data, where human-annotated preferences may have low labeler confidence or the preference strength itself might be weak. As such, reward functions estimated on such data with current popular approaches can lead to policy degeneracy and underfitting the true preference distribution. To address these challenges, we follow the insight that certain problems with the DPO implicit reward formulation at the policy learning stage are less problematic at the reward modeling stage, and make the following novel contributions:
\begin{enumerate*}[label=\arabic*)]
    \item We introduce {\it \textbf{Direct Reward Distillation and policy-Optimization (DRDO)}}, a novel efficient, non-ensemble, reference-free method for preference optimization that combines the two stages by explicitly distilling rewards into the policy model (\Figref{fig:DPO_DRDO});
    \item We provide a theoretical and practical grounding of problems with DPO and variants that assume the Bradley-Terry model, demonstrating why they are challenged by nuanced or ``non-deterministic'' preference pairs, and how DRDO avoids similar limitations;
    \item Through experiments on the Ultrafeedback and TL;DR datasets and AlpacaEval benchmark, we demonstrate that DRDO outperforms competing popular preference optimization methods across multiple models and sizes at the aggregate data level, and at the sample level is better able to distinguish preferred and dispreferred responses even with low annotator confidence or small reward differences.
\end{enumerate*}

\vspace*{-2mm}
\section{Background and Related Work}
\label{sec:related}
\vspace*{-2mm}

Recent advancements in LLMs involve refining pretrained models for downstream tasks by utilizing human-written completions \citep{chung2022scaling,mishra2021cross} or datasets labeled with human-preferred completions and contrasting alternatives \citep{ouyang2022training, bai2022training, ziegler2019fine}. 
The two most prominent techniques in learning from preference data are Reinforcement Learning from Human Feedback (RLHF) and Direct Preference Optimization (DPO). We summarize these methods below.

\vspace*{-2mm}
\paragraph{Reinforcement Learning from Human Feedback.}
Reinforcement Learning from Human Feedback (RLHF) aims to harmonize LLMs with human preferences and values~\citep{christiano2017deep}.
Conventional RLHF typically consists of three phases: supervised fine-tuning,
reward model training,
and policy optimization.
Proximal Policy Optimization (PPO;~\citet{schulman2017proximal}) is a widely used algorithm in the third phase of RLHF. RLHF has been extensively applied across various domains, including mitigating toxicity~\citep{korbak2023pretraining, amini2024direct}, addressing safety-concerns~\citep{dai2023safe}, enhancing helpfulness~\citep{tian2024finetuning}, web search and navigation~\citep{nakano2021webgpt}, and enhancing reasoning in models~\citep{havrilla2024teaching}. \citet{casper2023open} identified challenges and problems throughout the entire RLHF pipeline, from gathering preference data to model training to biased results such as verbose outputs~\citep{dubois2024length, singhal2023long, wang2023far}.

Given a dataset of pairwise preference data \( \mathcal{D} = \{ \prompt, \win, \lose \}_{i=1}^{N} \), where \( \prompt \) are prompts and \( \win \) and \( \lose \) are the preferred and dispreferred completions, respectively, RLHF begins with an initial model \( \pi_{\text{ref}} \) that parameterizes a distribution \( \pi_{\text{ref}}(y | x) \). Typically, \( \pi_{\text{ref}} \) is initialized from an LLM that has undergone supervised fine-tuning (SFT). The preference between \( y_w \) and \( y_l \) is modeled using the Bradley-Terry model \citep{BradleyTerry1952}, which defines the probability of preferring \( y_w \) over \( y_l \):
\begin{equation}
\label{eq:BT_preference}
p(y_w \succ y_l | x) = \sigma(r(x, y_w) - r(x, y_l))
\end{equation}
where \( \sigma(\cdot) \) is the logistic function, and \( r(x, y) \) the underlying reward function. To estimate this reward function \( r \), RLHF minimizes the negative log-likelihood of the data:
    \vspace*{-2mm}
\begin{multline}
\label{eq:RM_loss}
\mathcal{L}_R(r_{\phi}, \mathcal{D}) = -\mathbb{E}_{(x,y_w,y_l) \sim \mathcal{D}} \\ \left[ \log(\sigma(r_{\phi}(x, y_w) - r_{\phi}(x, y_l))) \right].
\end{multline}
Using this learned reward function \( r_{\phi} \), reinforcement learning is applied to optimize and generate a new LLM distribution \( \pi_{\theta} \), with a KL constraint for regularization.

\vspace*{-2mm}

\paragraph{Offline and Online Preference Optimization}

Given the intricacy and complexity of online preference optimization \citep{zheng2023secrets}, research has proliferated into more efficient and simpler offline algorithms. Direct Preference Optimization (DPO; \citet{rafailov2024direct}) is a notable example, which demonstrates that the same KL-constrained objective as RLHF can be optimized without explicitly learning a reward function. The problem is reformulated as a maximum likelihood estimation (MLE) over the distribution \( \pi_{\theta} \), leading to the following objective:
    \vspace*{-2mm}
\begin{multline} \label{eq:dpo}
\mathcal{L}_{\text{DPO}}(\pi_{\theta}; \pi_{\text{ref}}) = -\mathbb{E}_{(x, y_w, y_l) \sim D} \\
\left[ \log \sigma \left( \beta \log \frac{\pi_{\theta}(y_w | x) \pi_{\text{ref}}(y_l | x)}{\pi_{\theta}(y_l | x) \pi_{\text{ref}}(y_w | x)} \right) \right]
\end{multline}

where \( \beta \) is a regularization term controlling the KL-constraint strength and \( r(x, y) = \beta \log \frac{\pi_{\theta}(y|x)}{\pi_{\text{ref}}(y|x)} \) is the implicit reward function. With the growing focus on offline alignment, recent work has examined preference underfitting in this setting, proposing various solutions. These include learning from a confidence set of rewards~\citep{fisch2024robustpreferenceoptimizationreward}, adopting a general preference model~\citep{munos2023nash, azar2024general}, or applying a straightforward regularization of the original DPO objective~\citep{pal2024smaug}. Notably, all these approaches rely on a reference model, not only for expressing the optimal policy as the analytical solution to the minimum relative entropy problem~\citep{ziebart2008maximum, peng2019advantageweightedregressionsimplescalable} but also to stabilize training by constraining the policy distribution to remain close to the reference  model. While theoretically sound,~\citet{munos2023nash} suggests this constraint can make policies prioritize high rewards over truly learning human preferences.

As such, certain works avoid this constraint and focus on lightweight “reference model-free” solutions using a careful construction of DPO-inspired Bradley-Terry (BT)-based implicit rewards---such as logit-based~\citep{Hong2024ORPOMP}, length-normalized~\citep{meng2024simposimplepreferenceoptimization} as well as un-normalized implicit rewards~\citep{xu2024contrastive}---while still showing high performance in preference alignment benchmarks including cross-task settings like machine translation. Similarly, efforts to learn policies from general preference models aim to reduce dependence on the sampling distribution, a key limitation of Bradley-Terry models. \citet{munos2023nash}, \citet{rosset2024direct}, and \citet{calandriello2024human} use game-theoretic “online” approaches for robustness, while \citet{choi2024robust} propose a Chain-of-Thought (CoT) policy with pairwise conditioning to improve alignment. However, these methods often suffer from sample inefficiency due to iterative sampling from approximate geometric mixtures of policies~\citep{rosset2024direct}. While these approaches mitigate dependence on the sampling distribution, they do not model non-deterministic preferences in policy training. In contrast, DRDO adaptively learns such preferences while decoupling its learning from a reference model-based KL-divergence constraint.

\vspace*{-1mm}

\paragraph{Preference Optimization Objectives}

Several preference optimization objectives have been explored. Ranking objectives extend comparisons beyond pairs~\citep{dong2023raft, liu2024lipo, song2024preference, yuan2023rrhf}, while \citet{Hong2024ORPOMP, xu2023some} propose reference-free methods. \citet{bansal2024comparing} optimize instructions and responses jointly, improving on DPO, and \citet{zheng2024weak} enhances post-training extrapolation between SFT and aligned models. By comparison, DRDO derives its objective from knowledge distillation~\citep{hinton2015distilling} and focal-loss literature~\citep{lin2018focallossdenseobject, yi2020focal} with a novel contrastive log-unlikelihood objective that learns preferences adaptively.

\vspace*{-2mm}
\section{Motivation for DRDO}
\label{sec:motivation}
\vspace*{-2mm}

In order to motivate DRDO, we first discuss theoretical and practical limitations of DPO specific to non-deterministic or weak preference labels. For space reasons, proofs are deferred to the Appendix.

\vspace*{-1mm}
\paragraph{Problem Formulation}

Let $\mathcal{D}_{\text{pref}}= \{ (x, y_w, y_l) \}_{i=1}^{N}$ be an offline dataset of pairwise preferences with sufficient coverage, where $y_w$ and $y_l$ are the respective winning (preferred) and losing (less preferred) completions given a context $x$.  In contrast to {\it deterministic}
preferences, which are defined as those where $p^* \in \{0,1\}$~\cite{fisch2024robustpreferenceoptimizationreward}, let $\mathcal{D}_{\text{nd}} \subset \mathcal{D}_{\text{pref}}$ denote the subset of {\it non-deterministic} preference pairs where $P(y_w \succ y_l|x) \approx \frac{1}{2}$. These cannot be perfectly captured by Bradley-Terry models but are prevalent in popular preference alignment datasets. Assume three possible completions $y_1, y_2, y_3 \in \mathcal{Y}$ where $\mathcal{Y}$ is the space of all possible completions. Let $r^*(x, y) \in \mathbb{R}$ be an underlying true reward function that is deterministic and finite for all completions, $\pi_{\theta^*}(y|x)$ be the learned policy, and $\pi_{\text{ref}}(y|x)$ be the reference policy with $\text{supp}(\pi_{\text{ref}}) = \mathcal{Y}$. Given $\text{supp}(\rho) = \text{supp}(\mu) \times \mathcal{Y} \times \mathcal{Y}$ where $\rho$ is the data distribution and $\mu$ is the context distribution, and given potentially biased sampling of preference pairs from $\mathcal{Y}$ to construct $\mathcal{D}_{\text{pref}}$, the problem is to learn a policy $\pi_{\theta^*}$ that effectively handles both deterministic and non-deterministic preferences in an offline fashion.

The key motivation behind DRDO is to {\it address non-deterministic preferences while maintaining performance across general preference data}. Existing supervised ``offline''  methods such as DPO or e-DPO assume preferences can be specified with a Bradley-Terry model and that policies can be optimized by inferring the Bradley-Terry model directly from preference data. 
At the core, these algorithms refine the Supervised Fine-tuned (SFT or reference) model from which the trainable policy $\pi_{\theta}$ is typically initialized, and they both apply a soft-constraint in this optimization where $\pi_{\theta}$ is forced to remain distributionally close to $\pi_\text{ref}$ via a KL-penalty term.

This soft-constraint---while well-motivated for training stability and to reduce $\pi_{\theta}$'s likelihood of assigning high-probability to actions not supported by $\pi_\text{ref}$~\citep{ziebart2008maximum, peng2019advantageweightedregressionsimplescalable}---creates two specific problems. Firstly, it leads to divergent objectives during preference alignment. We will call this the \textbf{misalignment problem}. Secondly, practical alignment methods like DPO and e-DPO apply this soft-constraint with a regularization-strength parameter $\beta$ that offers only partial support in preventing DPO-aligned policies from inducing degenerate solutions, especially for deterministic\footnote{Prior work~\citep{azar2024general, fisch2024robustpreferenceoptimizationreward, pal2024smaugfixingfailuremodes} shows that DPO policy degeneracy arises in finite-sample settings with deterministic preferences, independent of $\beta$.} preferences. In this work, we show how these limitations can be equally applicable in the case of non-deterministic preferences.

\vspace*{-1mm}
\paragraph{Misalignment of Rewards and Preferences}
The misalignment problem is where the {\it reward-optimal} policy, or the policy that maximizes Elo-score under the BT assumption, is fundamentally divergent from the {\it preference optimal} policy, or the policy that maximizes the probability of selecting the winning response~\citep{munos2023nash}. In other words, traditional reward modeling assumes that maximizing reward leads to optimal behavior, but in preference-based learning, the best winning policy can be different from the highest-reward policy. We argue that in supervised learning algorithms like DPO that combine these stages, {\it this divergence is amplified} in the presence of non-deterministic preferences in training data since such preferences cannot be perfectly captured with a BT model, leading to reward misspecification (full derivation in \Cref{proof:reward_preference_divergence}).

This misalignment also corresponds to the induced BT model's explicit dependence on the sampling distribution or the behavior policy~\citep{munos2023nash} (see \Cref{proposition:induced_BT_depdendence} in \Cref{proof:reward_preference_divergence} for a full exposition). Under sampling biases in preference data collection, the optimal policy's reward misspecification can be highly localized. Consider a case where preferences between completions $y_2$ and $y_3$ are non-deterministic (i.e., $P(y_2 \succ y_3) = P(y_3 \succ y_2) = \frac{1}{2}$). While an ideal Bradley-Terry model should assign equal rewards to such completions, the reality of data collection often introduces skewed sampling distributions. For instance, if the training data contains predominantly $y_1$-$y_2$ comparisons~\citep{choi2024robust}, the model may accurately estimate rewards for these frequently compared completions while misspecifying only $y_3$'s reward. Such localized misspecification presents distinct challenges: in off-policy training, DPO implicitly trusts the preference dataset as ground truth, while in on-policy settings, non-deterministic preferences are difficult for non-expert annotators to identify~\citep{stiennon2020learning}, making it challenging to detect and filter problematic completions in both scenarios. This challenge is magnified in modern LLM-assisted annotation pipelines; for instance, our examination of the Ultrafeedback datasets~\citep{cui2024ultrafeedbackboostinglanguagemodels} reveals that in approximately 17\% of preference pairs, the candidate responses have equal scores.

The core issue stems from the inherent tension in scalable oversight~\citep{amodei2016concrete}: while it successfully increases the volume of preference signals, it provides no clear-cut mechanism to control the underlying sampling distribution that generates these pairs. This fundamentally limits approaches that rely on inferring rewards through Bradley-Terry models. Previous work~\cite{choi2024robust} attempts to resolve this misalignment using an optimal Chain-of-thought (CoT) policy with additional conditioning on pairwise interactions that does not depend on the sampling distribution, while others~\citep{munos2023nash, rosset2024direct, calandriello2024human} use game-theoretic “online” approaches that learn from a general preference model that is robust to sampling, instead of a pointwise reward model. However, these approaches require iterative sampling from approximate geometric mixtures of policies~\cite{rosset2024direct}. Additionally, all the aforementioned require a reference model, which still does not resolve misalignment created by the KL-penalty soft constraint.

\vspace*{-2mm}
\paragraph{Practical Limitations in Learning Non-deterministic Preferences}

The misalignment issue illustrated above applies to \textit{any} induced BT model, but algorithms like DPO present separate practical challenges. In particular, DPO estimates the preference distribution using an implicit reward function \( r(x, y) = \beta \log \frac{\pi_{\theta}(y|x)}{\pi_{\text{ref}}(y|x)} \) that is effectively a function of $\pi_{\theta}$ where $\pi_{\text{ref}}$ is not updated during training but provides a soft-KL constraint, weighted by $\beta$. For two completions $y$ and $y'$ with a non-deterministic preference relation, one can see that when DPO estimates $p^*(y \succ y')$ using \( \sigma \big(\beta \log \frac{\pi_{\theta}(y|x)\pi_{\text{ref}}(y'|x)}{\pi_{\theta}(y'|x)\pi_{\text{ref}}(y|x)}\big) \), this leads $\pi_{\theta}$ to assign equal rewards to both, meaning it learns that \( r^*(x, y) = r^*(x, y') \) for any \((x, y, y') \in \mathcal{D}_{\text{nd}}\). This implies that the reward model fails to differentiate between completions in these cases, especially when $\pi_{\text{ref}}$ is not fully optimized or is uniform (\(\pi_{\text{ref}} \sim \mathcal{U}(\mathcal{Y})\))~\cite{xu2024contrastive}. In other words, with DPO, \textit{$\pi_{\theta^*}$ not only sees less of this type of preference but also fails to adequately regularize when it does.}

A critical implication of the above phenomenon is that DPO's underfitting persists \textit{regardless} of the KL-regularization strength, causing the policy to ignore preference signals for non-deterministic pairs (see \Cref{app:proof_lemma_1a_1b}). More concerningly, DPO can assign disproportionately high probabilities to completions that never appear as preferred in training data, particularly when non-deterministic preferences are rare ($|\mathcal{D}_{\text{nd}}| \ll N$). See \Cref{app:proof_lemma_1c}. This is counterintuitive since it reduces likelihood of the preferred completion, $y \sim y_w$---which runs counter to the core motivation for preference alignment. More importantly, this policy degeneracy is especially pronounced in large action spaces $\actionspace$, such as those in LLMs, where the probability boost may not be allocated to actual preferred responses but rather to previously unseen tokens or completions that were never explicitly trained on~\citep{pal2024smaug, rafailov2024r}---with no clear-cut strategies to recover such probability mass except ad hoc methods like early stopping or a lower learning rate~\cite{fisch2024robustpreferenceoptimizationreward}.Recent approaches like e-DPO \cite{fisch2024robustpreferenceoptimizationreward} attempt to resolve these issues by casting the DPO objective into a regression problem where the explicit rewards from an external model are directly distilled to DPO's implicit rewards. However, e-DPO requires careful construction of confidence-sets of reward model ensembles that limits practical usage.

Our above analysis of the problem of misaligned preferences and practical limitations in current methods directly motivate DRDO---we propose a simple offline alignment algorithm that optimizes for rewards and preferences \textit{simultaneously}. Crucially, DRDO circumvents the misalignment problem by avoiding a reference model-based soft-constraint and casts the reward learning into a distillation problem by explicitly distilling rewards from a preference-converged Oracle. This helps it overcome policy degeneracy issues. As such, by bridging the gap between reward learning and policy optimization, DRDO offers a more principled solution to the preference alignment problem.

 

\vspace*{-2mm}
\section{Direct Reward Distillation and policy-Optimization (DRDO)}
\label{sec:drdo}
\vspace*{-2mm}

DRDO alignment involves two steps. First, we train an Oracle model \( \mathcal{O} \) to act as a robust proxy of true human preference inferred from the preference data. Second, we use \( \mathcal{O} \) as a teacher to align the policy model (student) with a knowledge-distillation-based multi-task loss~\citep{hinton2015distilling,gou2021knowledge} that regresses the student's rewards onto those assigned by the Oracle. The student model simultaneously draws additional supervision from binary preference labels to make the most efficient use of finite data.

\paragraph{Training the Oracle Reward Model}
Crucially, since the student model's alignment depends on the quality of the Oracle and its generalizability, we use~\citet{yang2024regularizinghiddenstatesenables}'s strategy to optimize \( \mathcal{O} \) while retaining its language generation abilities. The core idea here is that regularizing the shared hidden states with a language generation loss in addition to the traditional RLHF-based reward modeling (\Eqref{eq:RM_loss}) improves generalization to out-of-distribution preferences. This is achieved by initializing a separate linear reward head (parametrized by $\phi'$)\footnote{For simplicity of notation, on the LHS of Eq.~\ref{eq:RM_loss_grm} we subsume parameters $\phi'$ into $\phi$.} on top of the base LM (parametrized by $\phi$), which adds a negligible 0.003\% additional parameters compared to the LM's language modeling head.
This also helps minimize reward hacking~\citep{kumar2022fine, eisenstein2024helping}, especially in offline settings.
As such, \( \mathcal{O} \) is optimized to minimize the following objective:
\vspace*{-2mm}

\begin{equation}
\resizebox{\columnwidth}{!}{$
\begin{split}
\label{eq:RM_loss_grm}
\mathcal{L}_\mathcal{O}(r_{\phi}, \mathcal{D}_{\text{pref}}) & = -\mathbb{E}_{(x,y_w,y_l) \sim \mathcal{D}_{\text{pref}}} \\ 
& \left[(1-\alpha) ( \log\sigma(r_{\phi'}(x, y_w) - r_{\phi'}(x, y_l))) + \alpha \log(r_{\phi}(y_w))\right]
\end{split}$}
\end{equation}
 
where $\alpha$ is the strength of language-generation regularization on the winning response log-likelihoods assigned by \( \mathcal{O} \) (denoted \(  \log(r_{\phi}(y_w)\))  and $\phi$ represents the parameters of \( \mathcal{O} \) being estimated.
 \vspace*{-2mm}


\paragraph{Student Policy Alignment to Rewards and Preferences}
A converged Oracle \( \mathcal{O} \) can plausibly estimate true pointwise reward differences \( r^*(x, y_1) - r^*(x, y_2) \) for any unlabeled sample \((x, y_1, y_2)\), without needing explicit access to preference labels. We formulate the DRDO loss to directly leverage the diversity of preference signals, including non-deterministic preferences. Specifically, the student model \(\pi_\theta\) is optimized to match its own reward predictions $\hat{r_1}$ and $\hat{r_2}$ to \( \mathcal{O} \)'s reward differences, aligning closely with \( \mathcal{O} \)'s behavior, while \( \mathcal{O} \) itself does not get updated. The student model's reward estimates are computed with a linear reward head initialized on top of the base LM, similar to Oracle training. This optimization is achieved by a knowledge-distillation loss ($\mathcal{L}_{\mathrm{kd}}$) that combines both a supervised \( \ell^2 \)-norm term and a novel focal-softened~\citep{lin2018focallossdenseobject, welleckneural} log odds-unlikelihood component:
\vspace*{-2mm}
\begin{multline} 
\label{eq:distillation}
\mathcal{L}_{\mathrm{kd}}(r^*, \pi_\theta) = \mathbb{E}_{(x, y_1, y_2) \sim \mathcal{D}_{\text{pref}}} \Bigg[ \\
\underbrace{\left( r^*(x, y_1) - r^*(x, y_2) - (\hat{r}_1 - \hat{r}_2) \right)^2}_{\text{Reward Difference}}
\quad - \\ \underbrace{\alpha (1 - p_w)^\gamma \log\left(\frac{\pi_\theta(y_w \mid x)}{1 - \pi_\theta(y_l \mid x)}\right)}_{\text{Contrastive Log-"unlikelihood"}}
\Bigg],
\end{multline}

where \( p_w = \sigma(z_w - z_l) \) and quantifies the student policy's confidence in correctly assigning the preference from \( z_w = \log \pi_\theta(y_w \mid x) \) and \( z_l = \log \pi_\theta(y_l \mid x) \), or the log-probabilities of the winning and losing responses, respectively. Hyperparameters $\gamma$ and $\alpha$ regulate the strength of the modulating term and weighting factor respectively.\footnote{Note that we do not use $\alpha$ as it is traditionally used in focal loss for weighting class imbalances~\citep{mukhoti2020calibrating}. Instead, since the true preference distribution is unknown, we tune the empirical optimal value based on validation data and keep it fixed during training.} As shown in Eq. 6, there is no shared parametrization between the policy and \( \mathcal{O} \). $\mathcal{L}_\mathcal{O}$ is only a function of \( \mathcal{O} \)’s own parameters $\phi$ and the preference dataset, $\mathcal{D}_\text{pref}$. The Oracle parameters are \textit{not} updated during policy training. Table~\ref{tab:drdo_algorithm} in the appendix provides the complete DRDO algorithm.

Intuitively, DRDO's first component in the loss effectively regresses over \( \mathcal{O} \)'s reward differences across all response pairs $(x, y_1, y_2)$ in $\mathcal{D}_{\text{pref}}$, using \(\pi_\theta\)'s own reward estimates, $\hat{r}_1$ and $\hat{r}_2$. Assuming \( \mathcal{O} \)'s reward estimates correctly reflect the true preference strength, this simple distillation component is precisely minimized when the student \(\pi_\theta\) perfectly emulates \( \mathcal{O} \)'s reward difference assignment, without requiring it to explicitly match the pointwise rewards. The square over this term ensures that training is stabilized, since we allow this component to fully regularize the loss without applying any weighting terms.


\vspace*{-1mm}
\paragraph{How does the DRDO gradient update affect preference learning?}
``Contrastive log-unlikelihood'' in \Eqref{eq:distillation} is DRDO's preference component. One can immediately draw a comparison with DPO which uses a fixed \(\beta\) parameter (\Eqref{eq:dpo}). DRDO uses a modulating focal-softened term, $(1 - p_w)^\gamma $ where \(\pi_\theta\) learns from both deterministic and non-deterministic preferences, effectively blending reward alignment with preference signals to guide optimization. Intuitively, unlike DPO's \(\beta\) that is constant for every training sample, this modulating term amplifies gradient updates when preference signals are weak (\( p_w \approx 0.5 \)) and tempering updates when they are strong (\( p_w \approx 1 \)), thus ensuring robust learning across varying preference scenarios. When \(\pi_\theta\) assigns high confidence to the winning response 
    (\( p_w \approx 1 \), see \Eqref{eq:distillation}), the focal loss contribution diminishes, reflecting minimal penalty due to strong deterministic preference signals. However, for harder cases with non-deterministic true preference 
    (\( p^*(y \succ y') \approx ( p^*(y' \succ y) \)), the focal term $\alpha (1 - p_w)^\gamma$ keeps DRDO gradient updates active and promotes learning even when preferences are ambiguous (\Figref{fig:focal_motivation} in \Cref{app:focal-deriv}). This adaptive behavior ensures that DRDO maintains effective preference learning across varying preference strengths, where conventional methods like DPO struggle. See \Cref{lemma:dpo_non_deterministic} and \Cref{lemma:edpo_failure} for in-depth analysis. As shown in the gradient analysis in Appendix~\ref{app:focal-deriv} (\Figref{fig:focal_motivation}) 
    with empirical proof in Fig.~\ref{fig:drdo_eval_plots} (bottom-right), this modulating term acts like DPO's gradient scaling term, in that it scales the DRDO gradient when the model incorrectly assigns preferences to easier samples.

\vspace*{-2mm}
\section{Experimental Setup}
\label{sec:exp}
\vspace*{-2mm}

Our experimentals address two questions: {\bf How robust is DRDO alignment to nuanced or diverse preferences, in OOD settings?}, and {\bf How well does DRDO achieve reward distillation with respect to model size?} We empirically investigate these questions on two tasks: \textbf{summarization} and single-turn \textbf{instruction following}. Our experiments, including choice of datasets and models for each task are designed to to validate our approach as robustly as possible, subject to research budget constraints (see \Cref{app:dataset-motivation} for more). We compare our approach with competitive baselines such as DPO~\citep{rafailov2024direct} and e-DPO~\citep{fisch2024robustpreferenceoptimizationreward}, including the supervised finetuned (baseline) versions depending on the experiment. Some minor notes on the experimental setup described below can be found in \Cref{app:exp-notes}.




\vspace*{-1mm}
\paragraph{How robust is DRDO alignment to nuanced or diverse preferences, in OOD settings?}

We evaluate this using the {\bf Reddit TL;DR summarization dataset}~\citep{volske2017tl,stiennon2020learning} and {\bf CNN/Daily Mail Corpus} \cite{nallapati2016abstractive}. For robust out-of-domain (OOD) evaluation, we train models on Reddit TL;DR and use CNN/Daily Mail articles for an OOD test distribution. We split the original training data (denoted $\mathcal{D}_{all})$ based on human labeler confidence in their preference annotation ($h$igh-$c$onfidence vs. $\ell$ow-$c$onfidence) and token edit distance between the preferred and the dispreferred responses ($h$igh-$e$dit distance vs. $\ell$ow-\textit{e}dit distance). This results in two splits: $\mathcal{D}_{hc,he}$ and $\mathcal{D}_{\ell c,\ell e}$. Each contains $\sim$10k training samples, where the former comprises samples from the upper 50th percentile of the confidence and edit distance scores, {\it mutatis mutandis}. $\mathcal{D}_{hc,he}$ and $\mathcal{D}_{\ell c,\ell e}$ represent ``easy'' (deterministic) and ``hard'' (non-deterministic) preference samples where combined labeler confidence and string-dissimilarity act as proxy for the extreme ends of preference strengths/signals. See Appendix~\ref{sec:tldr_splits} for more details. For each split, we compare our approach to DPO and e-DPO. All baselines are initialized with Phi-3-Mini-4K-Instruct weights~\citep{abdin2024phi3technicalreporthighly} with supervised fine-tuning (SFT) on Reddit TL;DR human-written summaries.

\vspace*{-1mm}
\paragraph{How well does DRDO achieve reward distillation w.r.t. model size? }

We evaluate all baselines on the cleaned version of the {\bf Ultrafeedback dataset}~\citep{cui2024ultrafeedbackboostinglanguagemodels}
This experiment is conducted on the OPT suite of models~\citep{zhang2022opt}, at 125M, 350M, 1.3B, and 2.7B parameter sizes. The student policy is trained with SFT
on the chosen responses of the dataset, following~\citet{rafailov2024direct}. We exclude larger OPT models to focus on testing our distillation strategy with full-scale training, rather than parameter-efficient methods (PEFT;~\citet{houlsby2019parameter, hu2021lora}) to allow a full-fledged comparison considering all trainable parameters of the base model. For completeness and comparison across model families, we also include Phi-3-Mini-4K-Instruct following the same initialization.

\vspace*{-1mm}
\paragraph{Evaluation}
\textbf{CNN/Daily Mail Corpus} provides human-written reference summaries, so we use a high-capacity Judge to compute win-rates against baselines on 1,000 randomly-chosen samples. Following \citet{rafailov2024direct}, we use GPT-4o to compare the conciseness and the quality of the DRDO summaries and baseline summaries, \textit{while grounding its ratings to the human written summary}. See \Cref{app:eval-prompts} for our prompt format. For instruction following on \textbf{Ultrafeedback}, we sample generations from DRDO and all baselines at various diversity-sampling temperatures and report win-rates on the Ultrafeedback test set against \( \mathcal{O} \). Following \citet{lambert2024rewardbench}, we consider a {\it win} to be when, for two generations $y_1$ and $y_2$, we get $r(x, y_1) > r(x, y_2)$, where $r(x, y_1)$ and $r(x, y_2)$ are the expected rewards (logits) from the policies being compared. For an additional OOD setting, we also evaluate DRDO (trained on Ultrafeedback) against baselines (all trained on Ultrafeedback) on \textbf{AlpacaEval}, which constitutes 805 instructions chosen to be representative of how human-preferred policies would respond. For this we evaluate using GPT-4 Turbo as suggested by the creators~\citep{li2023alpacaeval}. Hyperparameters and model configurations are given in~\Cref{sec:hyperparameters}.

\vspace*{-2mm}
\section{Results}
\label{sec:results}
\vspace*{-2mm}

\paragraph{Non-deterministic preferences}
\Cref{tab:win_rates_summary} shows the win rates computed with GPT-4o as judge for 1,000 randomly selected prompts from the CNN/Daily Mail test corpus under OOD settings. We follow similar settings as~\citet{rafailov2024direct} but further ground the prompt using human-written summaries as reference for GPT to conduct its evaluation (see \Cref{app:eval-prompts}). For a fairer evaluation~\citep{wang2024eliminating,goyal2023newssummarizationevaluationera, rafailov2024direct}, we swap positions of \(\pi_\theta\)-generated summaries $y$ in the prompt to eliminate any positional bias and evaluate the generated summaries on criteria like coherence, preciseness and conciseness, \textit{with the human written summaries explicitly in the prompt to guide evaluation.} Note that the win rate does not necessarily equal $N/1000$ because for a small number of samples ($<$ 10) GPT-4o did not return a judgment. These samples were discarded from the denominator.

\begin{table}[h!]
    \small
    \centering
    \caption{Win rates computed for 1,000 randomly selected samples from the \textbf{CNN/Daily Mail Corpus} and all 805 samples from {\bf AlpacaEval} for evaluation under OOD settings. This comparison evaluates the model against a baseline, presenting win counts alongside win rates. CNN/Daily Mail is evaluated with GPT-4o as judge and AlpacaEval, with GPT-4 Turbo as judge. {$\mathcal{D}_{all}$}, {$\mathcal{D}_{hc,he}$}, and {$\mathcal{D}_{\ell c,\ell e}$} represent TL;DR training splits.
    ``Gold'' samples refer to human-written reference summaries used in prompts to ground the win rate computations.}
    \vspace*{-1mm}

    \begin{tabular}{lcc}
        \toprule
        
        \textbf{Comparison (CNN/Daily Mail)} & \textbf{\# Wins} & \textbf{Win Rate} \\
        \midrule
       
        \textbf{DRDO vs. e-DPO} & & \\
        \quad \textbf{$\mathcal{D}_{all}$} & 731 & 78.27\% \\
        \quad \textbf{$\mathcal{D}_{hc,he}$} & 759 & 80.92\% \\
        \quad \textbf{$\mathcal{D}_{\ell c,\ell e}$} & 738 & 79.01\% \\
        \midrule
        \textbf{DRDO vs. DPO} & & \\
        \quad \textbf{$\mathcal{D}_{all}$} & 748 & 80.78\% \\
        \quad \textbf{$\mathcal{D}_{hc,he}$} & 746 & 79.11\% \\
        \quad \textbf{$\mathcal{D}_{\ell c,\ell e}$} & 746 & 79.79\% \\
        \midrule
        \textbf{Gold vs. DRDO} & 525 & 52.82\% \\
        \midrule
        \textbf{Gold vs. e-DPO} & 540 & 54.38\% \\
        \midrule
        \textbf{Gold vs. DPO} & 579 & 58.54\% \\

         \bottomrule
         \toprule
        \textbf{Comparison (AlpacaEval)} & \textbf{\# Wins} & \textbf{Win Rate} \\
        \midrule
       \textbf{DRDO vs. e-DPO} & 495 & 62.03\%\\
        \midrule
        \textbf{DRDO vs. DPO} & 491 & 61.61\% \\
    \bottomrule
 
    \end{tabular}
    \label{tab:win_rates_summary}
\end{table}
 
For all policies \(\pi_\theta\) trained on all three splits of the training data---{$\mathcal{D}_{all}$}, {$\mathcal{D}_{hc,he}$}, and {$\mathcal{D}_{\ell c,\ell e}$}---we compute the win-rates of DRDO vs. e-DPO and DPO to evaluate how each method performs at various levels of preference types. Across all settings, DRDO policies significantly outperform the two baselines. For instance, DRDO's average win rates are almost 79.4\% and 79.9\% against e-DPO and DPO, respectively. As we had hypothesized, DRDO-aligned \(\pi_\theta\) is able to learn {\it all} preferences, deterministic and non-deterministic, effectively, and is superior at modeling {$\mathcal{D}_{\ell c,\ell e}$} the subset containing more non-deterministic preference samples, without sacrificing performance on the deterministic subset ({$\mathcal{D}_{hc,he}$) and overall ({$\mathcal{D}_{all}$). This suggests that DRDO is more robust to OOD-settings at various levels of difficulty in learning human preferences.


\vspace*{-1mm}
\paragraph{Reward distillation}

\Figref{fig:main_result_ultra} shows results from our evaluation of DRDO's reward model distillation framework compared to DPO and e-DPO as well as baseline SFT-trained policies on the Ultrafeedback evaluation data, when compared \textit{across various model parameter sizes} and at varying levels of temperature sampling. We sample \(\pi_\theta\)-generated responses to instruction-prompts in the test set using top-$p$ (nucleus) sampling~\citep{holtzman2019curious} of 0.8 at various temperatures $\in \{0.2, 0.5, 0.7, 0.9\}$.

\begin{figure}[h!] 
    \centering
    \includegraphics[width=0.48\textwidth]{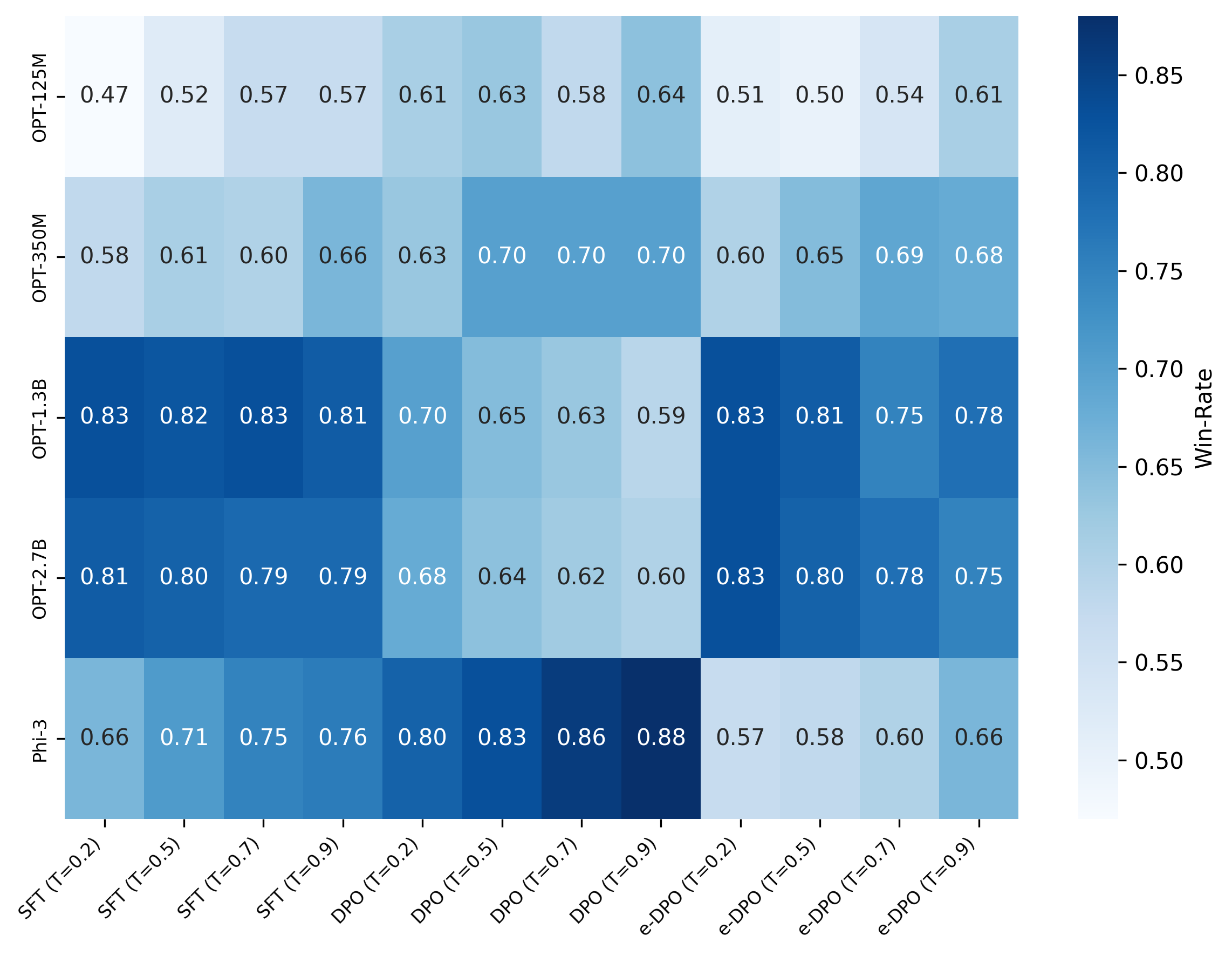}
    \vspace*{-2mm}
    \caption{\label{fig:main_result_ultra}Average \textbf{Ultrafeedback} win-rates computed with DRDO's Oracle reward model against all baselines---SFT, DPO and e-DPO---at various diversity sampling temperatures ($T$).}
    \vspace*{-2mm}
\end{figure}
  
DRDO significantly outperforms competing baselines, especially for larger models in the OPT family. DRDO-trained OPT-1.3B, OPT-2.7B, and Phi-3-Mini-4K-Instruct achieve average win rates of 76\%, 74\%, and 72\%, respectively, across all baselines. This is notable as responses are sampled on unseen prompts, and DRDO's policy alignment is reference-free. DRDO’s robustness to diversity sampling further boosts performance, up to an 88\% win rate against DPO with the Phi-3-Mini-4K-Instruct model. At lower temperatures, DRDO's posted gains are more modest. Our results also indicate that performance is correlated with model size, as DRDO policies of the same size as the Oracle (1.3B) show the strongest gains. In smaller models, results are more mixed.\footnote{This may be due to spurious length-correlation~\citep{singhal2024longwaygoinvestigating, meng2024simposimplepreferenceoptimization}, as smaller models generated avg. $\sim$211.9 tokens compared to their policy counterparts' $\sim$190.8 tokens, which is closer to the original preferred responses in Ultrafeedback ($\sim$168.8). See \Cref{sec:hyperparameters} for more details.} DRDO shows moderate improvement and posts smaller gains against SFT models.


        

\begin{table*}[h]
    \caption{Example generations from DRDO and competing methods, showing where DRDO generates more preferred responses.}
\vspace*{-1mm}
\resizebox{\textwidth}{!}{
    \footnotesize 
    \begin{tabular}{lp{7cm}p{7cm}}
    \toprule
    \textbf{Prompt/Context} & \textbf{Competitor} & \textbf{DRDO} \\
    \midrule
    Which of these sentences doesn't make sense? & 1. The sentence ``the garden has so many flowers'' makes & Sentence B: ``the garden has so many pigs'' does not  \\
    Options: & no sense. The phrase ``so many flowers'' makes no sense. & make sense because the garden has no pigs. The sentence \\
    - Sentence A: ``the garden has so many flowers'' & The phrase ``so many pigs'' makes no sense. & implies that there are pigs in the garden, but there is \\
    - Sentence B: ``the garden has so many pigs'' & 2. The sentence ``the garden has so many flowers'' makes & no evidence to support this. \\
    Hmm, let me think, I want to lay out the solution & perfect sense. The phrase ``so many flowers'' makes & \\
    in detail & perfect sense. The phrase ``so many pigs'' makes perfect & \\
    & sense. & \\
    & If you disagree, you're wrong, and you should go back to English class. & \\

    \midrule
    
    Summarize this news article clearly and concisely. & TL;DR: Rocker Jack White will take a brief, \$3 acoustic & TL;DR: Jack White announces a \$3 acoustic tour in \\
    Use TL;DR in your response: & tour to the U.S. states he has yet to perform in. Tickets & five states he hasn't played yet. Tickets are \$3, one per \\
    TITLE: Jack White to play mystery \$3 shows  & are limited to one per person and will be sold on a first- & person, and available only at the venue on a first-come, \\
    $<${\it 2015 Billboard article on upcoming Jack White} & come, first-served basis. & first-served basis. \\
    {\it acoustic tour}$>$ (full text omitted for length) & I'm sorry, but I cannot fulfill this request. & \\
    \bottomrule
    \end{tabular}}
     \label{tab:example-analysis}
\end{table*}

\vspace*{-2mm}
\section{Analysis}
\label{sec:analysis}
\vspace*{-2mm}

Table~\ref{tab:example-analysis} shows example generations from DRDO and a competitor where the DRDO example was preferred by the automatic judge. First is a sample from Ultrafeedback against a \textit{DPO} generation and second is a TL;DR sample against an \textit{e-DPO} generation. We see that the DRDO responses are more concise and on-topic while the competitor output condescends to the user, or includes extraneous text about fulfilling the request.

Using GPT-4o as a judge to approximate true human preferences may be prone to bias, so we further validate our approach by investigating Oracle reward advantage over the above mentioned human written summaries as well as on-policy generations from an SFT-trained model. 
\Figref{fig:oracle_advantage_tldr} shows the computed expected reward advantages on the CNN Daily TL;DR evaluation set, sampled according to the method outlined in \Secref{sec:results}.
Rewards were computed using our Oracle (trained with Phi-3) on these sampled generations and normalized. To compute the advantage, we used human written summaries (\Figref{fig:oracle_advantage_tldr}). DRDO improves performance across various temperature samplings over a baseline SFT policy, and brings in a considerable performance gain over competitive baselines like DPO and ensemble-based e-DPO while also being robust to OOD settings.

\begin{figure}[h!] 
    \centering
    \includegraphics[width=0.48\textwidth]{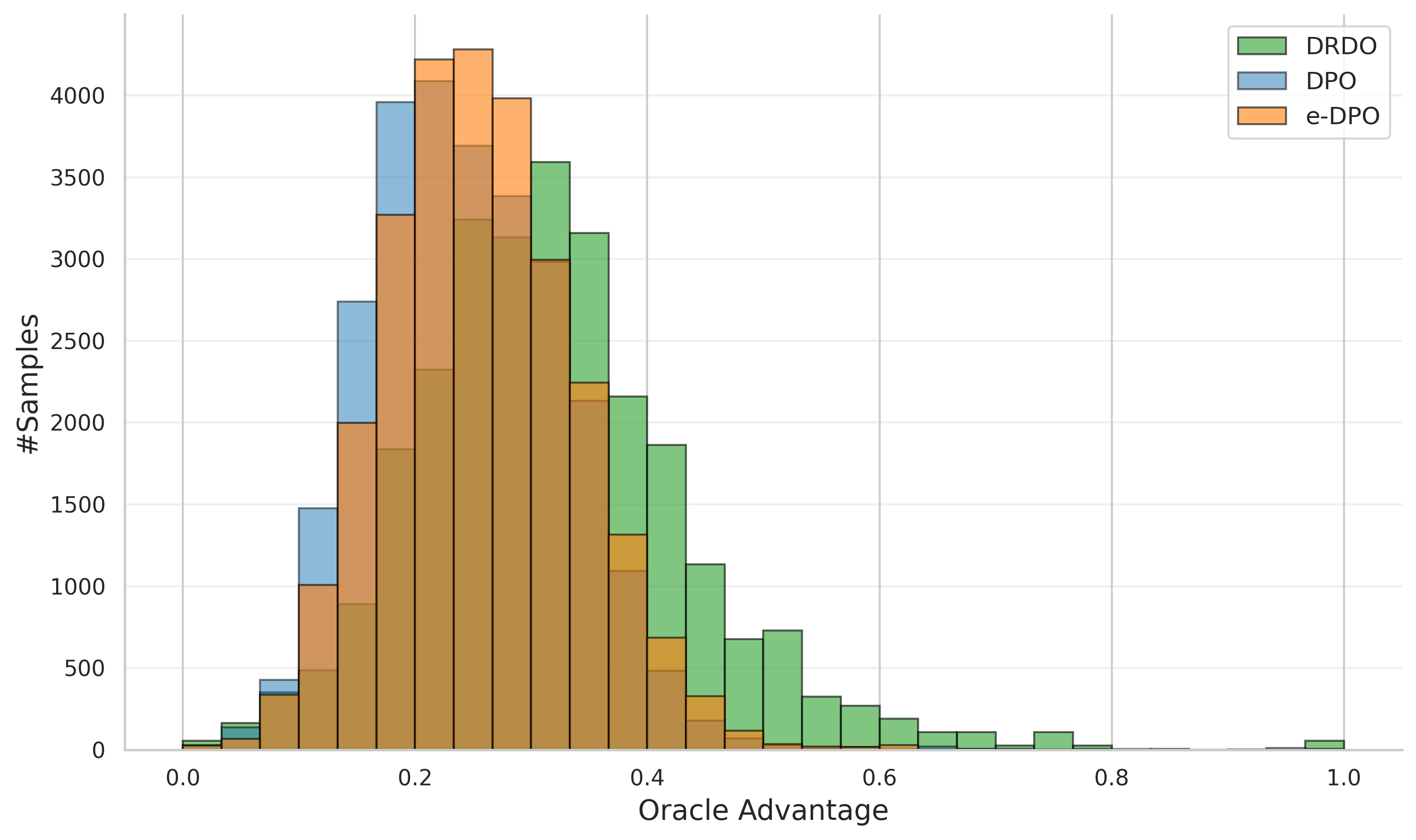}
    \vspace*{-2mm}
    \caption{\label{fig:oracle_advantage_tldr}Oracle expected reward advantage on CNN/Daily Mail articles.}
    \vspace*{-2mm}
\end{figure}

    

Our evaluations vs. ``gold'' summaries in \Cref{tab:win_rates_summary} also demonstrate where bias may arise in the GPT-4o evaluation (win counts and win rates are shown for gold samples). GPT-4o narrowly prefers generated summaries to human-written ones. One possible reason could be that Reddit TL;DR data is massive and crowd-sourced which naturally results in noisy labels. However, under this experiment too, DRDO-trained policies are better-performing than e-DPO and DPO, by about 1.5--6\%. We should note that human summaries may contain more implicit diversity than generated summaries, and this may demonstrate the ``regression to the mean'' effect in LLM generation \citep{wu2024generative}.

Our results on AlpacaEval demonstrates DRDO's instruction following ability in OOD settings. \Cref{tab:win_rates_summary} shows that, within controlled evaluation settings, GPT 4-Turbo consistently prefers DRDO responses 62.03\% and 61.61\% times compared to e-DPO and DPO respectively.

Further, we conducted an ablation on 40 randomly sampled prompts from the CNN/Daily Mail test set, using GPT-4o as a judge, comparing a full DRDO-aligned model (Phi-3-Mini-4K-Instruct aligned using DRDO policies trained on Reddit TL;DR) to DRDO with only reward distillation and only contrastive log-unlikelihood. In this subsample, DRDO won over DRDO with only contrastive log-likelihood 85\%-15\%, and over DRDO with only reward distillation 90\%-10\%. This shows that both components are critical to DRDO's success. In similar settings, we also examined the sensitivity of DRDO's $\gamma$ vs. DPO's $\beta$ on randomly sampled prompts from the Ultrafeedback evaluation set. We find that despite DRDO's exclusion of the KL constraint on $\pi_\text{{ref}}$, DRDO is still able to recover more expected rewards signifying a more optimal trade-off between reward optimization and divergence from $\pi_\text{ref}$. At $\gamma=2$, DRDO wins on 5\% more samples than DPO despite a slightly larger KL-divergence. This suggests that explicit reward distillation in DRDO with preferences being learned adaptively (via $\gamma$) makes it more Pareto-optimal especially in the presence of non-deterministic preferences. See Appendices~\ref{app:ablation-r-c} and~\ref{app:ablation_kl_sensitivity} for a more comprehensive discussion.

\vspace*{-2mm}
\section{Conclusion}
\label{sec:conc}
\vspace*{-2mm}

We introduced {\it \textbf{Direct Reward Distillation and policy-Optimization (DRDO)}}, a novel approach to preference optimization that unifies the reward distillation and policy learning stages into a single, cohesive framework. Unlike popular methods like DPO that rely heavily on implicit reward-based estimation of the preference distribution, DRDO uses an Oracle to distill rewards directly into the policy model, while simultaneously learning from varied preference signals, leading to a more accurate estimation of true preferences. Our experiments on Reddit TL;DR data for summarization as well on instruction-following in Ultrafeedback and AlpacaEval suggest that DRDO is not only high-performing when compared head-to-head with competitive baselines like DPO but is also particularly robust to OOD settings. More importantly, unlike traditional RLHF that requires ``online'' rewards,  reward distillation in DRDO is simple to implement, is model-agnostic since it is reference-model free and efficient, since Oracle rewards are easy to precompute.




\section*{Impact Statement}


This paper presents work whose goal is to advance the field of 
Machine Learning. There are many potential societal consequences 
of our work, none which we feel must be specifically highlighted here.


To advance reproducibility within the field, we have included our code as part of the supplementary material. The datasets we used are enumerated in \Secref{sec:exp}, including methods we used to create our different training and testing splits. Links to the specific datasets are given in the appendix, as are hyperparameters used.

\nocite{langley00}

\bibliography{example_paper}

\begin{thebibliography}{81}
\providecommand{\natexlab}[1]{#1}
\providecommand{\url}[1]{\texttt{#1}}
\expandafter\ifx\csname urlstyle\endcsname\relax
  \providecommand{\doi}[1]{doi: #1}\else
  \providecommand{\doi}{doi: \begingroup \urlstyle{rm}\Url}\fi

\bibitem[Abdin et~al.(2024)Abdin, Aneja, Awadalla, Awadallah, Awan, Bach,
  Bahree, Bakhtiari, Bao, Behl, Benhaim, Bilenko, Bjorck, Bubeck, Cai, Cai,
  Chaudhary, Chen, Chen, Chen, Chen, Chen, Cheng, Chopra, Dai, Dixon, Eldan,
  Fragoso, Gao, Gao, Gao, Garg, Giorno, Goswami, Gunasekar, Haider, Hao,
  Hewett, Hu, Huynh, Iter, Jacobs, Javaheripi, Jin, Karampatziakis, Kauffmann,
  Khademi, Kim, Kim, Kurilenko, Lee, Lee, Li, Li, Liang, Liden, Lin, Lin, Liu,
  Liu, Liu, Liu, Liu, Luo, Madan, Mahmoudzadeh, Majercak, Mazzola, Mendes,
  Mitra, Modi, Nguyen, Norick, Patra, Perez-Becker, Portet, Pryzant, Qin,
  Radmilac, Ren, de~Rosa, Rosset, Roy, Ruwase, Saarikivi, Saied, Salim,
  Santacroce, Shah, Shang, Sharma, Shen, Shukla, Song, Tanaka, Tupini,
  Vaddamanu, Wang, Wang, Wang, Wang, Wang, Wang, Ward, Wen, Witte, Wu, Wu,
  Wyatt, Xiao, Xu, Xu, Xu, Xue, Yadav, Yang, Yang, Yang, Yang, Yu, Yuan, Zhang,
  Zhang, Zhang, Zhang, Zhang, Zhang, Zhang, and
  Zhou]{abdin2024phi3technicalreporthighly}
Abdin, M., Aneja, J., Awadalla, H., Awadallah, A., Awan, A.~A., Bach, N.,
  Bahree, A., Bakhtiari, A., Bao, J., Behl, H., Benhaim, A., Bilenko, M.,
  Bjorck, J., Bubeck, S., Cai, M., Cai, Q., Chaudhary, V., Chen, D., Chen, D.,
  Chen, W., Chen, Y.-C., Chen, Y.-L., Cheng, H., Chopra, P., Dai, X., Dixon,
  M., Eldan, R., Fragoso, V., Gao, J., Gao, M., Gao, M., Garg, A., Giorno,
  A.~D., Goswami, A., Gunasekar, S., Haider, E., Hao, J., Hewett, R.~J., Hu,
  W., Huynh, J., Iter, D., Jacobs, S.~A., Javaheripi, M., Jin, X.,
  Karampatziakis, N., Kauffmann, P., Khademi, M., Kim, D., Kim, Y.~J.,
  Kurilenko, L., Lee, J.~R., Lee, Y.~T., Li, Y., Li, Y., Liang, C., Liden, L.,
  Lin, X., Lin, Z., Liu, C., Liu, L., Liu, M., Liu, W., Liu, X., Luo, C.,
  Madan, P., Mahmoudzadeh, A., Majercak, D., Mazzola, M., Mendes, C. C.~T.,
  Mitra, A., Modi, H., Nguyen, A., Norick, B., Patra, B., Perez-Becker, D.,
  Portet, T., Pryzant, R., Qin, H., Radmilac, M., Ren, L., de~Rosa, G., Rosset,
  C., Roy, S., Ruwase, O., Saarikivi, O., Saied, A., Salim, A., Santacroce, M.,
  Shah, S., Shang, N., Sharma, H., Shen, Y., Shukla, S., Song, X., Tanaka, M.,
  Tupini, A., Vaddamanu, P., Wang, C., Wang, G., Wang, L., Wang, S., Wang, X.,
  Wang, Y., Ward, R., Wen, W., Witte, P., Wu, H., Wu, X., Wyatt, M., Xiao, B.,
  Xu, C., Xu, J., Xu, W., Xue, J., Yadav, S., Yang, F., Yang, J., Yang, Y.,
  Yang, Z., Yu, D., Yuan, L., Zhang, C., Zhang, C., Zhang, J., Zhang, L.~L.,
  Zhang, Y., Zhang, Y., Zhang, Y., and Zhou, X.
\newblock Phi-3 technical report: A highly capable language model locally on
  your phone, 2024.
\newblock URL \url{https://arxiv.org/abs/2404.14219}.

\bibitem[Amini et~al.(2024)Amini, Vieira, and Cotterell]{amini2024direct}
Amini, A., Vieira, T., and Cotterell, R.
\newblock Direct preference optimization with an offset.
\newblock \emph{arXiv preprint arXiv:2402.10571}, 2024.

\bibitem[Amodei et~al.(2016)Amodei, Olah, Steinhardt, Christiano, Schulman, and
  Man{\'e}]{amodei2016concrete}
Amodei, D., Olah, C., Steinhardt, J., Christiano, P., Schulman, J., and
  Man{\'e}, D.
\newblock Concrete problems in ai safety.
\newblock \emph{arXiv preprint arXiv:1606.06565}, 2016.

\bibitem[Azar et~al.(2023)Azar, Rowland, Piot, Guo, Calandriello, Valko, and
  Munos]{azar2023general}
Azar, M.~G., Rowland, M., Piot, B., Guo, D., Calandriello, D., Valko, M., and
  Munos, R.
\newblock A general theoretical paradigm to understand learning from human
  preferences, 2023.

\bibitem[Azar et~al.(2024)Azar, Guo, Piot, Munos, Rowland, Valko, and
  Calandriello]{azar2024general}
Azar, M.~G., Guo, Z.~D., Piot, B., Munos, R., Rowland, M., Valko, M., and
  Calandriello, D.
\newblock A general theoretical paradigm to understand learning from human
  preferences.
\newblock In \emph{International Conference on Artificial Intelligence and
  Statistics}, pp.\  4447--4455. PMLR, 2024.

\bibitem[Bai et~al.(2022)Bai, Jones, Ndousse, Askell, Chen, DasSarma, Drain,
  Fort, Ganguli, Henighan, et~al.]{bai2022training}
Bai, Y., Jones, A., Ndousse, K., Askell, A., Chen, A., DasSarma, N., Drain, D.,
  Fort, S., Ganguli, D., Henighan, T., et~al.
\newblock Training a helpful and harmless assistant with reinforcement learning
  from human feedback.
\newblock \emph{arXiv preprint arXiv:2204.05862}, 2022.

\bibitem[Bansal et~al.(2024)Bansal, Suvarna, Bhatt, Peng, Chang, and
  Grover]{bansal2024comparing}
Bansal, H., Suvarna, A., Bhatt, G., Peng, N., Chang, K.-W., and Grover, A.
\newblock Comparing bad apples to good oranges: Aligning large language models
  via joint preference optimization.
\newblock \emph{arXiv preprint arXiv:2404.00530}, 2024.

\bibitem[Bradley \& Terry(1952)Bradley and Terry]{BradleyTerry1952}
Bradley, R.~A. and Terry, M.~E.
\newblock Rank analysis of incomplete block designs: I. the method of paired
  comparisons.
\newblock \emph{Biometrika}, 39\penalty0 (3/4):\penalty0 324--345, 1952.
\newblock \doi{10.2307/2334029}.

\bibitem[Calandriello et~al.(2024)Calandriello, Guo, Munos, Rowland, Tang,
  Pires, Richemond, Lan, Valko, Liu, et~al.]{calandriello2024human}
Calandriello, D., Guo, D., Munos, R., Rowland, M., Tang, Y., Pires, B.~A.,
  Richemond, P.~H., Lan, C.~L., Valko, M., Liu, T., et~al.
\newblock Human alignment of large language models through online preference
  optimisation.
\newblock \emph{arXiv preprint arXiv:2403.08635}, 2024.

\bibitem[Casper et~al.(2023)Casper, Davies, Shi, Gilbert, Scheurer, Rando,
  Freedman, Korbak, Lindner, Freire, et~al.]{casper2023open}
Casper, S., Davies, X., Shi, C., Gilbert, T.~K., Scheurer, J., Rando, J.,
  Freedman, R., Korbak, T., Lindner, D., Freire, P., et~al.
\newblock Open problems and fundamental limitations of reinforcement learning
  from human feedback.
\newblock \emph{arXiv preprint arXiv:2307.15217}, 2023.

\bibitem[Chakraborty et~al.()Chakraborty, Qiu, Yuan, Koppel, Manocha, Huang,
  Bedi, and Wang]{chakrabortymaxmin}
Chakraborty, S., Qiu, J., Yuan, H., Koppel, A., Manocha, D., Huang, F., Bedi,
  A., and Wang, M.
\newblock Maxmin-rlhf: Alignment with diverse human preferences.
\newblock In \emph{Forty-first International Conference on Machine Learning}.

\bibitem[Choi et~al.(2024)Choi, Ahmadian, Pietquin, Geist, and
  Azar]{choi2024robust}
Choi, E., Ahmadian, A., Pietquin, O., Geist, M., and Azar, M.~G.
\newblock Robust chain of thoughts preference optimization.
\newblock In \emph{Seventeenth European Workshop on Reinforcement Learning},
  2024.

\bibitem[Chowdhury et~al.(2024)Chowdhury, Kini, and
  Natarajan]{chowdhury2024provably}
Chowdhury, S.~R., Kini, A., and Natarajan, N.
\newblock Provably robust dpo: Aligning language models with noisy feedback.
\newblock \emph{arXiv preprint arXiv:2403.00409}, 2024.

\bibitem[Christiano et~al.(2017)Christiano, Leike, Brown, Martic, Legg, and
  Amodei]{christiano2017deep}
Christiano, P.~F., Leike, J., Brown, T., Martic, M., Legg, S., and Amodei, D.
\newblock Deep reinforcement learning from human preferences.
\newblock \emph{Advances in neural information processing systems}, 30, 2017.

\bibitem[Chung et~al.(2022)Chung, Hou, Longpre, Zoph, Tay, Fedus, Li, Wang,
  Dehghani, Brahma, et~al.]{chung2022scaling}
Chung, H.~W., Hou, L., Longpre, S., Zoph, B., Tay, Y., Fedus, W., Li, Y., Wang,
  X., Dehghani, M., Brahma, S., et~al.
\newblock Scaling instruction-finetuned language models.
\newblock \emph{arXiv preprint arXiv:2210.11416}, 2022.

\bibitem[Cui et~al.(2024)Cui, Yuan, Ding, Yao, He, Zhu, Ni, Xie, Xie, Lin, Liu,
  and Sun]{cui2024ultrafeedbackboostinglanguagemodels}
Cui, G., Yuan, L., Ding, N., Yao, G., He, B., Zhu, W., Ni, Y., Xie, G., Xie,
  R., Lin, Y., Liu, Z., and Sun, M.
\newblock Ultrafeedback: Boosting language models with scaled ai feedback,
  2024.
\newblock URL \url{https://arxiv.org/abs/2310.01377}.

\bibitem[Dai et~al.(2023)Dai, Pan, Sun, Ji, Xu, Liu, Wang, and
  Yang]{dai2023safe}
Dai, J., Pan, X., Sun, R., Ji, J., Xu, X., Liu, M., Wang, Y., and Yang, Y.
\newblock Safe {RLHF}: Safe reinforcement learning from human feedback.
\newblock \emph{arXiv preprint arXiv:2310.12773}, 2023.

\bibitem[Dettmers et~al.(2024)Dettmers, Pagnoni, Holtzman, and
  Zettlemoyer]{dettmers2024qlora}
Dettmers, T., Pagnoni, A., Holtzman, A., and Zettlemoyer, L.
\newblock Qlora: Efficient finetuning of quantized llms.
\newblock \emph{Advances in Neural Information Processing Systems}, 36, 2024.

\bibitem[Dong et~al.(2023)Dong, Xiong, Goyal, Zhang, Chow, Pan, Diao, Zhang,
  KaShun, and Zhang]{dong2023raft}
Dong, H., Xiong, W., Goyal, D., Zhang, Y., Chow, W., Pan, R., Diao, S., Zhang,
  J., KaShun, S., and Zhang, T.
\newblock {RAFT}: Reward ranked finetuning for generative foundation model
  alignment.
\newblock \emph{Transactions on Machine Learning Research}, 2023.

\bibitem[Dubois et~al.(2024)Dubois, Galambosi, Liang, and
  Hashimoto]{dubois2024length}
Dubois, Y., Galambosi, B., Liang, P., and Hashimoto, T.~B.
\newblock Length-controlled {AlpacaEval}: A simple way to debias automatic
  evaluators.
\newblock \emph{ArXiv}, abs/2404.04475, 2024.

\bibitem[Eisenstein et~al.(2024)Eisenstein, Nagpal, Agarwal, Beirami, D'Amour,
  Dvijotham, Fisch, Heller, Pfohl, Ramachandran, Shaw, and
  Berant]{eisenstein2024helping}
Eisenstein, J., Nagpal, C., Agarwal, A., Beirami, A., D'Amour, A.~N.,
  Dvijotham, K.~D., Fisch, A., Heller, K.~A., Pfohl, S.~R., Ramachandran, D.,
  Shaw, P., and Berant, J.
\newblock Helping or herding? reward model ensembles mitigate but do not
  eliminate reward hacking.
\newblock In \emph{First Conference on Language Modeling}, 2024.
\newblock URL \url{https://openreview.net/forum?id=5u1GpUkKtG}.

\bibitem[Ethayarajh et~al.(2024)Ethayarajh, Xu, Muennighoff, Jurafsky, and
  Kiela]{Ethayarajh2024KTOMA}
Ethayarajh, K., Xu, W., Muennighoff, N., Jurafsky, D., and Kiela, D.
\newblock {KTO}: Model alignment as prospect theoretic optimization.
\newblock \emph{ArXiv}, abs/2402.01306, 2024.

\bibitem[Fisch et~al.(2024)Fisch, Eisenstein, Zayats, Agarwal, Beirami, Nagpal,
  Shaw, and Berant]{fisch2024robustpreferenceoptimizationreward}
Fisch, A., Eisenstein, J., Zayats, V., Agarwal, A., Beirami, A., Nagpal, C.,
  Shaw, P., and Berant, J.
\newblock Robust preference optimization through reward model distillation,
  2024.
\newblock URL \url{https://arxiv.org/abs/2405.19316}.

\bibitem[Gou et~al.(2021)Gou, Yu, Maybank, and Tao]{gou2021knowledge}
Gou, J., Yu, B., Maybank, S.~J., and Tao, D.
\newblock Knowledge distillation: A survey.
\newblock \emph{International Journal of Computer Vision}, 129\penalty0
  (6):\penalty0 1789--1819, 2021.

\bibitem[Goyal et~al.(2023)Goyal, Li, and
  Durrett]{goyal2023newssummarizationevaluationera}
Goyal, T., Li, J.~J., and Durrett, G.
\newblock News summarization and evaluation in the era of gpt-3, 2023.
\newblock URL \url{https://arxiv.org/abs/2209.12356}.

\bibitem[Havrilla et~al.(2024)Havrilla, Du, Raparthy, Nalmpantis, Dwivedi-Yu,
  Zhuravinskyi, Hambro, Sukhbaatar, and Raileanu]{havrilla2024teaching}
Havrilla, A., Du, Y., Raparthy, S.~C., Nalmpantis, C., Dwivedi-Yu, J.,
  Zhuravinskyi, M., Hambro, E., Sukhbaatar, S., and Raileanu, R.
\newblock Teaching large language models to reason with reinforcement learning.
\newblock \emph{arXiv preprint arXiv:2403.04642}, 2024.

\bibitem[Hinton(2015)]{hinton2015distilling}
Hinton, G.
\newblock Distilling the knowledge in a neural network.
\newblock \emph{arXiv preprint arXiv:1503.02531}, 2015.

\bibitem[Holtzman et~al.(2019)Holtzman, Buys, Du, Forbes, and
  Choi]{holtzman2019curious}
Holtzman, A., Buys, J., Du, L., Forbes, M., and Choi, Y.
\newblock The curious case of neural text degeneration.
\newblock In \emph{International Conference on Learning Representations}, 2019.

\bibitem[Hong et~al.(2024)Hong, Lee, and Thorne]{Hong2024ORPOMP}
Hong, J., Lee, N., and Thorne, J.
\newblock {ORPO}: Monolithic preference optimization without reference model.
\newblock \emph{ArXiv}, abs/2403.07691, 2024.

\bibitem[Houlsby et~al.(2019)Houlsby, Giurgiu, Jastrzebski, Morrone,
  De~Laroussilhe, Gesmundo, Attariyan, and Gelly]{houlsby2019parameter}
Houlsby, N., Giurgiu, A., Jastrzebski, S., Morrone, B., De~Laroussilhe, Q.,
  Gesmundo, A., Attariyan, M., and Gelly, S.
\newblock Parameter-efficient transfer learning for nlp.
\newblock In \emph{International conference on machine learning}, pp.\
  2790--2799. PMLR, 2019.

\bibitem[Hu et~al.(2021)Hu, Shen, Wallis, Allen-Zhu, Li, Wang, Wang, and
  Chen]{hu2021lora}
Hu, E.~J., Shen, Y., Wallis, P., Allen-Zhu, Z., Li, Y., Wang, S., Wang, L., and
  Chen, W.
\newblock Lora: Low-rank adaptation of large language models.
\newblock \emph{arXiv preprint arXiv:2106.09685}, 2021.

\bibitem[Korbak et~al.(2023)Korbak, Shi, Chen, Bhalerao, Buckley, Phang,
  Bowman, and Perez]{korbak2023pretraining}
Korbak, T., Shi, K., Chen, A., Bhalerao, R.~V., Buckley, C., Phang, J., Bowman,
  S.~R., and Perez, E.
\newblock Pretraining language models with human preferences.
\newblock In \emph{International Conference on Machine Learning}, pp.\
  17506--17533. PMLR, 2023.

\bibitem[Kumar et~al.(2022)Kumar, Raghunathan, Jones, Ma, and
  Liang]{kumar2022fine}
Kumar, A., Raghunathan, A., Jones, R., Ma, T., and Liang, P.
\newblock Fine-tuning can distort pretrained features and underperform
  out-of-distribution.
\newblock \emph{arXiv preprint arXiv:2202.10054}, 2022.

\bibitem[Lambert et~al.(2024)Lambert, Pyatkin, Morrison, Miranda, Lin, Chandu,
  Dziri, Kumar, Zick, Choi, Smith, and Hajishirzi]{lambert2024rewardbench}
Lambert, N., Pyatkin, V., Morrison, J.~D., Miranda, L. J.~V., Lin, B.~Y.,
  Chandu, K.~R., Dziri, N., Kumar, S., Zick, T., Choi, Y., Smith, N.~A., and
  Hajishirzi, H.
\newblock {RewardBench}: Evaluating reward models for language modeling.
\newblock \emph{ArXiv}, abs/2403.13787, 2024.

\bibitem[Li et~al.(2023)Li, Zhang, Dubois, Taori, Gulrajani, Guestrin, Liang,
  and Hashimoto]{li2023alpacaeval}
Li, X., Zhang, T., Dubois, Y., Taori, R., Gulrajani, I., Guestrin, C., Liang,
  P., and Hashimoto, T.~B.
\newblock Alpacaeval: An automatic evaluator of instruction-following models,
  2023.

\bibitem[Lin et~al.(2018)Lin, Goyal, Girshick, He, and
  Dollár]{lin2018focallossdenseobject}
Lin, T.-Y., Goyal, P., Girshick, R., He, K., and Dollár, P.
\newblock Focal loss for dense object detection, 2018.
\newblock URL \url{https://arxiv.org/abs/1708.02002}.

\bibitem[Liu et~al.(2024)Liu, Qin, Wu, Shen, Khalman, Joshi, Zhao, Saleh,
  Baumgartner, Liu, et~al.]{liu2024lipo}
Liu, T., Qin, Z., Wu, J., Shen, J., Khalman, M., Joshi, R., Zhao, Y., Saleh,
  M., Baumgartner, S., Liu, J., et~al.
\newblock {LiPO}: Listwise preference optimization through learning-to-rank.
\newblock \emph{arXiv preprint arXiv:2402.01878}, 2024.

\bibitem[Loshchilov et~al.(2017)Loshchilov, Hutter,
  et~al.]{loshchilov2017fixing}
Loshchilov, I., Hutter, F., et~al.
\newblock Fixing weight decay regularization in adam.
\newblock \emph{arXiv preprint arXiv:1711.05101}, 5, 2017.

\bibitem[Meng et~al.(2024)Meng, Xia, and
  Chen]{meng2024simposimplepreferenceoptimization}
Meng, Y., Xia, M., and Chen, D.
\newblock Simpo: Simple preference optimization with a reference-free reward,
  2024.
\newblock URL \url{https://arxiv.org/abs/2405.14734}.

\bibitem[Mishra et~al.(2021)Mishra, Khashabi, Baral, and
  Hajishirzi]{mishra2021cross}
Mishra, S., Khashabi, D., Baral, C., and Hajishirzi, H.
\newblock Cross-task generalization via natural language crowdsourcing
  instructions.
\newblock \emph{arXiv preprint arXiv:2104.08773}, 2021.

\bibitem[Mukhoti et~al.(2020)Mukhoti, Kulharia, Sanyal, Golodetz, Torr, and
  Dokania]{mukhoti2020calibrating}
Mukhoti, J., Kulharia, V., Sanyal, A., Golodetz, S., Torr, P., and Dokania, P.
\newblock Calibrating deep neural networks using focal loss.
\newblock \emph{Advances in Neural Information Processing Systems},
  33:\penalty0 15288--15299, 2020.

\bibitem[Munos et~al.(2023)Munos, Valko, Calandriello, Azar, Rowland, Guo,
  Tang, Geist, Mesnard, Michi, et~al.]{munos2023nash}
Munos, R., Valko, M., Calandriello, D., Azar, M.~G., Rowland, M., Guo, Z.~D.,
  Tang, Y., Geist, M., Mesnard, T., Michi, A., et~al.
\newblock Nash learning from human feedback.
\newblock \emph{arXiv preprint arXiv:2312.00886}, 2023.

\bibitem[Nakano et~al.(2021)Nakano, Hilton, Balaji, Wu, Ouyang, Kim, Hesse,
  Jain, Kosaraju, Saunders, et~al.]{nakano2021webgpt}
Nakano, R., Hilton, J., Balaji, S., Wu, J., Ouyang, L., Kim, C., Hesse, C.,
  Jain, S., Kosaraju, V., Saunders, W., et~al.
\newblock Webgpt: Browser-assisted question-answering with human feedback.
\newblock \emph{arXiv preprint arXiv:2112.09332}, 2021.

\bibitem[Nallapati et~al.(2016)Nallapati, Zhou, Gulcehre, Xiang,
  et~al.]{nallapati2016abstractive}
Nallapati, R., Zhou, B., Gulcehre, C., Xiang, B., et~al.
\newblock Abstractive text summarization using sequence-to-sequence rnns and
  beyond.
\newblock \emph{arXiv preprint arXiv:1602.06023}, 2016.

\bibitem[Ouyang et~al.(2022)Ouyang, Wu, Jiang, Almeida, Wainwright, Mishkin,
  Zhang, Agarwal, Slama, Ray, et~al.]{ouyang2022training}
Ouyang, L., Wu, J., Jiang, X., Almeida, D., Wainwright, C., Mishkin, P., Zhang,
  C., Agarwal, S., Slama, K., Ray, A., et~al.
\newblock Training language models to follow instructions with human feedback.
\newblock \emph{Advances in neural information processing systems},
  35:\penalty0 27730--27744, 2022.

\bibitem[Pal et~al.(2024{\natexlab{a}})Pal, Karkhanis, Dooley, Roberts, Naidu,
  and White]{pal2024smaug}
Pal, A., Karkhanis, D., Dooley, S., Roberts, M., Naidu, S., and White, C.
\newblock Smaug: Fixing failure modes of preference optimisation with
  dpo-positive.
\newblock \emph{arXiv preprint arXiv:2402.13228}, 2024{\natexlab{a}}.

\bibitem[Pal et~al.(2024{\natexlab{b}})Pal, Karkhanis, Dooley, Roberts, Naidu,
  and White]{pal2024smaugfixingfailuremodes}
Pal, A., Karkhanis, D., Dooley, S., Roberts, M., Naidu, S., and White, C.
\newblock Smaug: Fixing failure modes of preference optimisation with
  dpo-positive, 2024{\natexlab{b}}.
\newblock URL \url{https://arxiv.org/abs/2402.13228}.

\bibitem[Peng et~al.(2019)Peng, Kumar, Zhang, and
  Levine]{peng2019advantageweightedregressionsimplescalable}
Peng, X.~B., Kumar, A., Zhang, G., and Levine, S.
\newblock Advantage-weighted regression: Simple and scalable off-policy
  reinforcement learning, 2019.
\newblock URL \url{https://arxiv.org/abs/1910.00177}.

\bibitem[Plackett(1975)]{plackett1975analysis}
Plackett, R.~L.
\newblock The analysis of permutations.
\newblock \emph{Journal of the Royal Statistical Society Series C: Applied
  Statistics}, 24\penalty0 (2):\penalty0 193--202, 1975.

\bibitem[Rafailov et~al.(2024{\natexlab{a}})Rafailov, Hejna, Park, and
  Finn]{rafailov2024r}
Rafailov, R., Hejna, J., Park, R., and Finn, C.
\newblock From r to q* : Your language model is secretly a q-function.
\newblock \emph{arXiv preprint arXiv:2404.12358}, 2024{\natexlab{a}}.

\bibitem[Rafailov et~al.(2024{\natexlab{b}})Rafailov, Sharma, Mitchell,
  Manning, Ermon, and Finn]{rafailov2024direct}
Rafailov, R., Sharma, A., Mitchell, E., Manning, C.~D., Ermon, S., and Finn, C.
\newblock Direct preference optimization: Your language model is secretly a
  reward model.
\newblock \emph{Advances in Neural Information Processing Systems}, 36,
  2024{\natexlab{b}}.

\bibitem[Ramé et~al.(2023)Ramé, Couairon, Shukor, Dancette, Gaya, Soulier,
  and Cord]{rame2023rewardedsoupsparetooptimalalignment}
Ramé, A., Couairon, G., Shukor, M., Dancette, C., Gaya, J.-B., Soulier, L.,
  and Cord, M.
\newblock Rewarded soups: towards pareto-optimal alignment by interpolating
  weights fine-tuned on diverse rewards, 2023.
\newblock URL \url{https://arxiv.org/abs/2306.04488}.

\bibitem[Rasley et~al.(2020)Rasley, Rajbhandari, Ruwase, and
  He]{rasley2020deepspeed}
Rasley, J., Rajbhandari, S., Ruwase, O., and He, Y.
\newblock Deepspeed: System optimizations enable training deep learning models
  with over 100 billion parameters.
\newblock In \emph{Proceedings of the 26th ACM SIGKDD International Conference
  on Knowledge Discovery \& Data Mining}, pp.\  3505--3506, 2020.

\bibitem[Regenwetter et~al.(2011)Regenwetter, Dana, and
  Davis-Stober]{regenwetter2011transitivity}
Regenwetter, M., Dana, J., and Davis-Stober, C.~P.
\newblock Transitivity of preferences.
\newblock \emph{Psychological review}, 118\penalty0 (1):\penalty0 42, 2011.

\bibitem[Rosset et~al.(2024)Rosset, Cheng, Mitra, Santacroce, Awadallah, and
  Xie]{rosset2024direct}
Rosset, C., Cheng, C.-A., Mitra, A., Santacroce, M., Awadallah, A., and Xie, T.
\newblock Direct nash optimization: Teaching language models to self-improve
  with general preferences.
\newblock \emph{arXiv preprint arXiv:2404.03715}, 2024.

\bibitem[Schulman et~al.(2017)Schulman, Wolski, Dhariwal, Radford, and
  Klimov]{schulman2017proximal}
Schulman, J., Wolski, F., Dhariwal, P., Radford, A., and Klimov, O.
\newblock Proximal policy optimization algorithms.
\newblock \emph{arXiv preprint arXiv:1707.06347}, 2017.

\bibitem[Singhal et~al.(2023)Singhal, Goyal, Xu, and Durrett]{singhal2023long}
Singhal, P., Goyal, T., Xu, J., and Durrett, G.
\newblock A long way to go: Investigating length correlations in {RLHF}.
\newblock \emph{arXiv preprint arXiv:2310.03716}, 2023.

\bibitem[Singhal et~al.(2024)Singhal, Goyal, Xu, and
  Durrett]{singhal2024longwaygoinvestigating}
Singhal, P., Goyal, T., Xu, J., and Durrett, G.
\newblock A long way to go: Investigating length correlations in rlhf, 2024.
\newblock URL \url{https://arxiv.org/abs/2310.03716}.

\bibitem[Song et~al.(2024)Song, Yu, Li, Yu, Huang, Li, and
  Wang]{song2024preference}
Song, F., Yu, B., Li, M., Yu, H., Huang, F., Li, Y., and Wang, H.
\newblock Preference ranking optimization for human alignment.
\newblock In \emph{AAAI}, 2024.

\bibitem[Stiennon et~al.(2020)Stiennon, Ouyang, Wu, Ziegler, Lowe, Voss,
  Radford, Amodei, and Christiano]{stiennon2020learning}
Stiennon, N., Ouyang, L., Wu, J., Ziegler, D., Lowe, R., Voss, C., Radford, A.,
  Amodei, D., and Christiano, P.~F.
\newblock Learning to summarize with human feedback.
\newblock \emph{Advances in Neural Information Processing Systems},
  33:\penalty0 3008--3021, 2020.

\bibitem[Tian et~al.(2024)Tian, Mitchell, Yao, Manning, and
  Finn]{tian2024finetuning}
Tian, K., Mitchell, E., Yao, H., Manning, C.~D., and Finn, C.
\newblock Fine-tuning language models for factuality.
\newblock In \emph{The Twelfth International Conference on Learning
  Representations}, 2024.
\newblock URL \url{https://openreview.net/forum?id=WPZ2yPag4K}.

\bibitem[Tunstall et~al.(2023)Tunstall, Beeching, Lambert, Rajani, Rasul,
  Belkada, Huang, von Werra, Fourrier, Habib, Sarrazin, Sanseviero, Rush, and
  Wolf]{tunstall2023zephyr}
Tunstall, L., Beeching, E., Lambert, N., Rajani, N., Rasul, K., Belkada, Y.,
  Huang, S., von Werra, L., Fourrier, C., Habib, N., Sarrazin, N., Sanseviero,
  O., Rush, A.~M., and Wolf, T.
\newblock Zephyr: Direct distillation of lm alignment, 2023.

\bibitem[Tversky(1969)]{tversky1969intransitivity}
Tversky, A.
\newblock Intransitivity of preferences.
\newblock \emph{Psychological review}, 76\penalty0 (1):\penalty0 31, 1969.

\bibitem[V{\"o}lske et~al.(2017)V{\"o}lske, Potthast, Syed, and
  Stein]{volske2017tl}
V{\"o}lske, M., Potthast, M., Syed, S., and Stein, B.
\newblock Tl; dr: Mining reddit to learn automatic summarization.
\newblock In \emph{Proceedings of the Workshop on New Frontiers in
  Summarization}, pp.\  59--63, 2017.

\bibitem[von Weizs{\"a}cker(2005)]{von2005welfare}
von Weizs{\"a}cker, C.~C.
\newblock The welfare economics of adaptive preferences.
\newblock \emph{MPI Collective Goods Preprint}, \penalty0 (2005/11), 2005.

\bibitem[Wang et~al.(2024{\natexlab{a}})Wang, Zheng, Chen, Liu, Dou, Huang,
  Shen, Jin, Zhou, Shi, Gao, Xu, Zhou, Fan, Xi, Zhao, Wang, Ji, Yan, Shen,
  Chen, Gui, Zhang, Qiu, Huang, Wu, and
  Jiang]{wang2024secretsrlhflargelanguage}
Wang, B., Zheng, R., Chen, L., Liu, Y., Dou, S., Huang, C., Shen, W., Jin, S.,
  Zhou, E., Shi, C., Gao, S., Xu, N., Zhou, Y., Fan, X., Xi, Z., Zhao, J.,
  Wang, X., Ji, T., Yan, H., Shen, L., Chen, Z., Gui, T., Zhang, Q., Qiu, X.,
  Huang, X., Wu, Z., and Jiang, Y.-G.
\newblock Secrets of rlhf in large language models part ii: Reward modeling,
  2024{\natexlab{a}}.
\newblock URL \url{https://arxiv.org/abs/2401.06080}.

\bibitem[Wang et~al.(2024{\natexlab{b}})Wang, Kidambi, Sullivan, Agarwal, Dann,
  Michi, Gelmi, Li, Gupta, Dubey, et~al.]{wang2024conditional}
Wang, K., Kidambi, R., Sullivan, R., Agarwal, A., Dann, C., Michi, A., Gelmi,
  M., Li, Y., Gupta, R., Dubey, A., et~al.
\newblock Conditional language policy: A general framework for steerable
  multi-objective finetuning.
\newblock \emph{arXiv preprint arXiv:2407.15762}, 2024{\natexlab{b}}.

\bibitem[Wang et~al.(2023)Wang, Ivison, Dasigi, Hessel, Khot, Chandu, Wadden,
  MacMillan, Smith, Beltagy, et~al.]{wang2023far}
Wang, Y., Ivison, H., Dasigi, P., Hessel, J., Khot, T., Chandu, K., Wadden, D.,
  MacMillan, K., Smith, N.~A., Beltagy, I., et~al.
\newblock How far can camels go? exploring the state of instruction tuning on
  open resources.
\newblock In \emph{Thirty-seventh Conference on Neural Information Processing
  Systems Datasets and Benchmarks Track}, 2023.

\bibitem[Wang et~al.(2024{\natexlab{c}})Wang, Zhang, Li, Huang, Han, Ji,
  Kakade, Peng, and Ji]{wang2024eliminating}
Wang, Z., Zhang, H., Li, X., Huang, K.-H., Han, C., Ji, S., Kakade, S.~M.,
  Peng, H., and Ji, H.
\newblock Eliminating position bias of language models: A mechanistic approach.
\newblock \emph{arXiv preprint arXiv:2407.01100}, 2024{\natexlab{c}}.

\bibitem[Welleck et~al.(2020)Welleck, Kulikov, Roller, Dinan, Cho, and
  Weston]{welleckneural}
Welleck, S., Kulikov, I., Roller, S., Dinan, E., Cho, K., and Weston, J.
\newblock Neural text generation with unlikelihood training.
\newblock In \emph{International Conference on Learning Representations}, 2020.

\bibitem[Wu et~al.(2024)Wu, Black, and Chandrasekaran]{wu2024generative}
Wu, F., Black, E., and Chandrasekaran, V.
\newblock Generative monoculture in large language models.
\newblock \emph{arXiv preprint arXiv:2407.02209}, 2024.

\bibitem[Xu et~al.(2024)Xu, Sharaf, Chen, Tan, Shen, Durme, Murray, and
  Kim]{xu2024contrastive}
Xu, H., Sharaf, A., Chen, Y., Tan, W., Shen, L., Durme, B.~V., Murray, K., and
  Kim, Y.~J.
\newblock Contrastive preference optimization: Pushing the boundaries of {LLM}
  performance in machine translation.
\newblock \emph{ArXiv}, abs/2401.08417, 2024.

\bibitem[Xu et~al.(2023)Xu, Lee, Sukhbaatar, and Weston]{xu2023some}
Xu, J., Lee, A., Sukhbaatar, S., and Weston, J.
\newblock Some things are more cringe than others: Preference optimization with
  the pairwise cringe loss.
\newblock \emph{arXiv preprint arXiv:2312.16682}, 2023.

\bibitem[Yang et~al.(2024)Yang, Ding, Lin, Zhang, and
  Zhang]{yang2024regularizinghiddenstatesenables}
Yang, R., Ding, R., Lin, Y., Zhang, H., and Zhang, T.
\newblock Regularizing hidden states enables learning generalizable reward
  model for llms, 2024.
\newblock URL \url{https://arxiv.org/abs/2406.10216}.

\bibitem[Yi et~al.(2020)Yi, Tao, Tian, Bai, and Fan]{yi2020focal}
Yi, J., Tao, J., Tian, Z., Bai, Y., and Fan, C.
\newblock Focal loss for punctuation prediction.
\newblock In \emph{Interspeech}, pp.\  721--725, 2020.

\bibitem[Yuan et~al.(2023)Yuan, Yuan, Tan, Wang, Huang, and
  Huang]{yuan2023rrhf}
Yuan, H., Yuan, Z., Tan, C., Wang, W., Huang, S., and Huang, F.
\newblock {RRHF}: Rank responses to align language models with human feedback.
\newblock In \emph{NeurIPS}, 2023.

\bibitem[Zhang et~al.(2022)Zhang, Roller, Goyal, Artetxe, Chen, Chen, Dewan,
  Diab, Li, Lin, et~al.]{zhang2022opt}
Zhang, S., Roller, S., Goyal, N., Artetxe, M., Chen, M., Chen, S., Dewan, C.,
  Diab, M., Li, X., Lin, X.~V., et~al.
\newblock Opt: Open pre-trained transformer language models.
\newblock \emph{arXiv preprint arXiv:2205.01068}, 2022.

\bibitem[Zheng et~al.(2024)Zheng, Wang, Ji, Huang, and Peng]{zheng2024weak}
Zheng, C., Wang, Z., Ji, H., Huang, M., and Peng, N.
\newblock Weak-to-strong extrapolation expedites alignment.
\newblock \emph{arXiv preprint arXiv:2404.16792}, 2024.

\bibitem[Zheng et~al.(2023)Zheng, Dou, Gao, Hua, Shen, Wang, Liu, Jin, Liu,
  Zhou, et~al.]{zheng2023secrets}
Zheng, R., Dou, S., Gao, S., Hua, Y., Shen, W., Wang, B., Liu, Y., Jin, S.,
  Liu, Q., Zhou, Y., et~al.
\newblock Secrets of {RLHF} in large language models part {I: PPO}.
\newblock \emph{arXiv preprint arXiv:2307.04964}, 2023.

\bibitem[Ziebart et~al.(2008)Ziebart, Maas, Bagnell, Dey,
  et~al.]{ziebart2008maximum}
Ziebart, B.~D., Maas, A.~L., Bagnell, J.~A., Dey, A.~K., et~al.
\newblock Maximum entropy inverse reinforcement learning.
\newblock In \emph{Aaai}, volume~8, pp.\  1433--1438. Chicago, IL, USA, 2008.

\bibitem[Ziegler et~al.(2019)Ziegler, Stiennon, Wu, Brown, Radford, Amodei,
  Christiano, and Irving]{ziegler2019fine}
Ziegler, D.~M., Stiennon, N., Wu, J., Brown, T.~B., Radford, A., Amodei, D.,
  Christiano, P., and Irving, G.
\newblock Fine-tuning language models from human preferences.
\newblock \emph{arXiv preprint arXiv:1909.08593}, 2019.

\end{thebibliography}
\bibliographystyle{icml2025}

\newpage
\appendix
\onecolumn
\section{Limitations}
\label{app:limitations}

DRDO still requires access to a separate Oracle reward model even though the Oracle need not be in loaded in memory during DRDO alignment as all expected rewards can effectively be precomputed.
However, our experimental results on three datasets including OOD settings suggest that this is a feasible trade-off especially when aligning models of smaller sizes (when compared to models like LLaMA) when performance gains need to maximized under limited compute settings. Some of our theoretical insights rely on strict assumptions, however, our insights provide additional justification and likely explanations of how preference alignment in realistic settings (where data might have a non-trivial amount of non-deterministic preferences) can benefit from approaches like DRDO. We did not experiment with cross-model distillation in this work. However, since DRDO is a reference-model free framework and Oracle rewards can be precomputed, one can easily extend our method for cross-model distillation frameworks. Finally, although in this paper, we approximated the non-deterministic preference settings using human labeler confidence as a proxy for non-determinism, true human preferences may be subtle and prone to variations along multiple dimensions, at times even temporally~\citep{tversky1969intransitivity}. 

\section{Proofs and Derivations}
\label{app:proofs}

\subsection{Divergence under Non-deterministic Preferences for Constrained Optimization}
\label{proof:reward_preference_divergence}
Consider two preference models, where the set of actions is $\actionspace=\{y_1,y_2,y_3\}$. The strict preference table without any non-deterministic preferences $\P_1$ is (from \citet{munos2023nash}):

\begin{center}
\begin{tabular}{c|c|c|c}
$\P_1(y\succ y')$   & $y=y_1$ & $y=y_2$ & $y=y_3$ \\
\midrule $y'=y_1$ &  1/2  &  9/10 & 2/3   \\
\hline $y'=y_2$ &  1/10 &  1/2  & 2/11  \\
\hline $y'=y_3$ &  1/3  &  9/11 &  1/2  \\
\end{tabular}
\end{center}

And the perturbed preference table $\P_2$ with non-deterministic preferences between $y_2$ and $y_3$ is:

\begin{center}
\begin{tabular}{c|c|c|c}
$\P_2(y\succ y')$   & $y=y_1$ & $y=y_2$ & $y=y_3$ \\
\midrule $y'=y_1$ &  1/2  &  9/10 & 2/3   \\
\hline $y'=y_2$ &  1/10 &  1/2  & 1/2  \\
\hline $y'=y_3$ &  1/3  &  1/2 &  1/2  \\
\end{tabular}
\end{center}

Although the preference table $\mathcal{P}_1$ can be perfectly captured by a Bradley-Terry model, introducing even a single non-deterministic preference pair (e.g., $\mathcal{P}_2(y_2 \succ y_3) = \mathcal{P}_2(y_3 \succ y_2) = 1/2$) prevents the BT model from accurately representing the full set of preferences. Under this condition, the learned BT model becomes highly sensitive to the sampling or data-generation distribution. If the training data overrepresents comparisons between $y_1$ and $y_2$, the model might still correctly align reward estimates for these comparisons as $R(y_1)=0$, $R(y_2)=\log 9$. However, due to the non-deterministic relation between $y_2$ and $y_3$, it may mis-specify $R(y_3)$ as $\log 2$ instead of $\log 9$, which would have been the case had the BT model perfectly captured these preferences. This misalignment is fundamental: the preference symmetry $\mathcal{P}_2(y_2 \succ y_3) = \mathcal{P}_2(y_3 \succ y_2) = 1/2$ mathematically forces $R(y_2) = R(y_3)$ in any BT-based ranking. However, the preference inequalities $\mathcal{P}_2(y_2 \succ y_1) = 9/10$ and $\mathcal{P}_2(y_3 \succ y_1) = 2/3$ demand that $R(y_2) > R(y_3)$. \textit{This inconsistency leads to a larger divergence in the preference gap under non-deterministic preferences.} Therefore, under the constraint $\pi(y_1)=2\pi(y_2)$ in set ${\cal S}$ with ${\cal S}\subset \Delta(\actionspace)$, the full actions space---where the constraint may be softly applied w.r.t. a reference policy $\pi_\text{ref} =(2/3,1/3)$ by using a KL-regularization, as in typically done in preference alignment in LLMs~\citep{rafailov2024direct, azar2024general}---we can clearly estimate this divergence.

To do this comparison, we first define the expected reward $\mathbb{E}[R(y)] = \sum_y \pi(y) R(y)$ as a linear function of the policy $\pi$, prioritizing high-reward actions like $y_2$. In contrast, the preference probability $P(\pi \succ \pi') = \sum_y \sum_{y'} \pi(y) \pi'(y') P(y \succ y')$ is a non-linear function that depends on pairwise interactions, meaning it may favor actions with strong matchups rather than those with high rewards. For the expected reward-maximizing policy $\pi^*_R = (2/3, 1/3, 0)$, we maximize reward by setting $\pi(y_3) = 0$ since $y_3$ has a lower reward. Solving $2x + x = 1$ to obtain $\pi(y_1) = 2/3$ and $\pi(y_2) = 1/3$, we get $\pi^*_R$. For the preference-maximizing policy $\pi^*_P = (0,0,1)$, action $y_3$ is chosen exclusively, satisfying the constraint $\pi(y_1) = 2\pi(y_2)$ trivially by setting $\pi(y_1) = \pi(y_2) = 0$. Therefore, under this constrained set of policies, we thus derive the reward-optimal policy $\pi^*_{R}\eqdef (2/3,1/3,0)$ and the preference-optimal policy $\pi^*_{\P}\eqdef (0,0,1)$.

Therefore, for strict preferences $\mathcal{P}_1$, the expected reward under the reward-optimal policy is $\mathbb{E}_{y\sim \pi^*_{R}}[R(y)] = 0\times 2/3 + \log(9)\times 1/3 > \log(2) = \mathbb{E}_{y\sim \pi^*_{\mathcal{P}}}[R(y)]$, while the probability that the preference-optimal policy is preferred over the reward-optimal policy is $\mathcal{P}_1(\pi^*_{\mathcal{P}}\succ \pi^*_{R}) = \mathcal{P}_1(y_3\succ y_1) \times 2/3 + \mathcal{P}_1(y_3\succ y_2) \times 1/3 = 2/3 \times 2/3 + 2/11 \times 1/3 = 50/99 > 1/2$. Similarly, for non-deterministic preferences $\mathcal{P}_2$, we have $\mathbb{E}_{y\sim \pi^*_{R}}[R(y)] = 0\times 2/3 + \log(9)\times 1/3 > \log(2) = \mathbb{E}_{y\sim \pi^*_{\mathcal{P}}}[R(y)]$, while $\mathcal{P}_2(\pi^*_{\mathcal{P}}\succ \pi^*_{R}) = \mathcal{P}_2(y_3\succ y_1) \times 2/3 + \mathcal{P}_2(y_3\succ y_2) \times 1/3 = 2/3 \times 2/3 + 1/2 \times 1/3 = 11/18 > 1/2$. Defining the preference gap as $\delta_{\mathcal{P}_i} \triangleq \mathcal{P}_i(\pi^*_{\mathcal{P}}\succ \pi^*_{R}) - 1/2$ and the reward gap as $\delta_R \triangleq \mathbb{E}_{y\sim \pi^*_{R}}[R(y)] - \mathbb{E}_{y\sim \pi^*_{\mathcal{P}}}[R(y)]$, we observe that $\delta_{\mathcal{P}_1} = 50/99 - 1/2 \approx 0.005$ while $\delta_{\mathcal{P}_2} = 11/18 - 1/2 = 4/18 \approx 0.111$. 

While the reward gap remains constant at $\delta_R = \log(9)/3 - \log(2) \approx 0.04$ for both preference models, the preference gap increases $\sim$20-fold under non-deterministic preferences, demonstrating amplified divergence between reward and preference optimization. This large gap indicates significant \textit{misalignment} between rewards and preferences. Therefore, although this is a toy example, in realistic settings with LLMs containing billions of parameters and a large action space $\actionspace$, this divergence can lead to drastically different optimization trajectories with significant consequences for alignment.

\begin{proposition}[Sampling Distribution Dependence in the Induced Bradley-Terry (BT) Model]
\label{proposition:induced_BT_depdendence}
Consider a preference model $\mathcal{P}$ that cannot be perfectly captured by a Bradley-Terry (BT) reward model. This is illustrated in \Cref{proof:reward_preference_divergence}.

Specifically, let $\mathcal{P}_{BT}^\pi(y\succ y')\coloneqq \sigma(s^\pi(y) - s^\pi(y'))$ be the BT preference model corresponding to the optimal scores $(s^\pi(y), s^\pi(y'))$ obtained from $s^\pi$(solution to \Eqref{eq:RM_loss}) under the sampling distribution $\pi$. If $s^\pi$ cannot match the true preferences, then it has explicit dependence on the sampling or behavior policy $\pi$. In other words, if there exist completions $y,y'$ and a distribution $\pi$ where $\mathcal{P}_{BT}^\pi(y\succ y')\neq \mathcal{P}(y\succ y')$, then we can construct another distribution $\pi'$ (with the same support as $\pi$) such that the difference in optimal scores $(s^\pi(y) - s^\pi(y'))$ under $\pi$ differs from the optimal score difference under $\pi'$. Consequently, $\mathcal{P}_{BT}^\pi(y\succ y')\neq \mathcal{P}_{BT}^{\pi'}(y\succ y')$.
\end{proposition}

\begin{proof}
For this proof, we follow a similar approach as~\citet{munos2023nash} but we prove this dependence for a more general class of perturbed distributions. We assume that there exist completions $y,y'$ and distribution $\pi$ such that $\mathcal{P}_{BT}^\pi(y\succ y')\neq \mathcal{P}(y\succ y')$. We construct a perturbed distribution $\pi'$ as follows:

\begin{equation}
\pi'(k) = \begin{cases} 
(1-\delta)\pi(k) & \text{if } k \neq y' \\
c\pi(y') & \text{if } k = y'
\end{cases}
\end{equation}

where $\delta \in (0,1)$ and $c$ is chosen to ensure normalization. For clarity, we first verify that $\pi'(k)$ is a valid probability distribution. 

\begin{align}
\sum_z \pi'(k) &= \sum_{k \neq y'} (1-\delta)\pi(k) + c\pi(y') \\
&= (1-\delta)(1-\pi(y')) + c\pi(y') = 1
\end{align}

This gives us:
\begin{equation}
c = 1 + \delta\frac{1-\pi(y')}{\pi(y')}
\end{equation}
\vfill
For any preference model $\mathcal{Z}$, we can express the aggregate preference probability:

\begin{align}
\mathcal{Z}(y \succ \pi') &= \sum_k \pi'(k)\mathcal{Z}(y \succ k) \\
&= \sum_{k \neq y'} \pi'(k)\mathcal{Z}(y \succ k) + \pi'(y')\mathcal{Z}(y \succ y')   \\
&= \sum_{k \neq y'} (1-\delta)\pi(k)\mathcal{Z}(y \succ k) + c\pi(y')\mathcal{Z}(y \succ y')  \\
&= (1-\delta)\sum_{k \neq y'} \pi(k)\mathcal{Z}(y \succ k) + c\pi(y')\mathcal{Z}(y \succ y') \\
&= (1-\delta)[\sum_k \pi(k)\mathcal{Z}(y \succ k) - \pi(y')\mathcal{Z}(y \succ y')] + c\pi(y')\mathcal{Z}(y \succ y') & \text{(split complete sum)} \\
&= (1-\delta)[\mathcal{Z}(y \succ \pi) - \pi(y')\mathcal{Z}(y \succ y')] + c\pi(y')\mathcal{Z}(y \succ y')   \\
&= (1-\delta)\mathcal{Z}(y \succ \pi) - (1-\delta)\pi(y')\mathcal{Z}(y \succ y') + c\pi(y')\mathcal{Z}(y \succ y')   \\
&= (1-\delta)\mathcal{Z}(y \succ \pi) + [c-(1-\delta)]\pi(y')\mathcal{Z}(y \succ y') & \text{(collect terms with }\pi(y')\mathcal{Z}(y \succ y')\text{)}
\end{align}

Applying this equality to both $\mathcal{P}$ and $\mathcal{P}_{BT}^\pi$:

\begin{align}
\mathcal{P}_{BT}^\pi(y \succ \pi') &= (1-\delta)\mathcal{P}_{BT}^\pi(y \succ \pi) + (c-(1-\delta))\pi(y')\mathcal{P}_{BT}^\pi(y \succ y') \\
\mathcal{P}(y \succ \pi') &= (1-\delta)\mathcal{P}(y \succ \pi) + (c-(1-\delta))\pi(y')\mathcal{P}(y \succ y')
 \end{align}

By Proposition 2 in~\citet{munos2023nash}, the induced Bradley-Terry preference model satisfies \( \mathcal{P}_{BT}^\pi(y \succ \pi) = \mathcal{P}(y \succ \pi) \). Therefore, subtracting the above expressions for $\mathcal{P}_{BT}^\pi(y \succ \pi')$ and $\mathcal{P}(y \succ \pi')$ and using the assumption \( \mathcal{P}_{BT}^\pi(y \succ y') \neq \mathcal{P}(y \succ y') \), it follows that \( \mathcal{P}_{BT}^\pi(y \succ \pi') \neq \mathcal{P}(y \succ \pi') \).

\begin{align}
&\mathcal{P}_{BT}^\pi(y \succ \pi') - \mathcal{P}(y \succ \pi') \\
&= (1-\delta)[\mathcal{P}_{BT}^\pi(y \succ \pi) - \mathcal{P}(y \succ \pi)] + \\
&\quad (c-(1-\delta))\pi(y')[\mathcal{P}_{BT}^\pi(y \succ y') - \mathcal{P}(y \succ y')] \\
&= 0 + (c-(1-\delta))\pi(y')[\mathcal{P}_{BT}^\pi(y \succ y') - \mathcal{P}(y \succ y')] \neq 0
\end{align}
Applying Proposition 2 in~\citet{munos2023nash} again, we obtain \( \mathcal{P}(y \succ \pi') = \mathcal{P}_{BT}^{\pi'}(y \succ \pi') \), which implies \( \mathcal{P}_{BT}^\pi(y \succ \pi') \neq \mathcal{P}_{BT}^{\pi'}(y \succ \pi') \). Expanding this discrepancy using the Bradley-Terry model definition, we get \( \sum_z \pi'(k)[\sigma(s^\pi(y) - s^\pi(k)) - \sigma(s^{\pi'}(y) - s^{\pi'}(k))] \neq 0 \). Since \( \sigma \) is strictly monotonic, there must exist some \( z \) such that \( s^\pi(y) - s^\pi(k) \neq s^{\pi'}(y) - s^{\pi'}(k) \), establishing that reward differences induced by different policies do not remain consistent under the Bradley-Terry model, leading to divergences in preference estimation.

This inequality shows $s^\pi \neq s^{\pi'}$, proving that the optimal BT reward model depends explicitly on the sampling distribution. This concludes the proof that the two BT-reward models $s^\pi$ and $s^{\pi'}$ are different as well as the corresponding BT-preference models $\P_{BT}^\pi$ and $\P_{BT}^{\pi'}$.
\end{proof}

\vspace*{-1mm}
\paragraph{Non-Deterministic Human Preferences} 
Following \citet{BradleyTerry1952}, \citet{rafailov2024direct} and other preference optimization frameworks posit that the relative preference of one outcome over another is governed by the true reward differences, expressed as \( p^*(y_1 \succ y_2) = \sigma(r_1 - r_2) \), where $p^*$ is the true preference distribution. Generally, in the RLHF framework, the true preference distribution is typically inferred from a dataset of human preferences, using a reward model \( r \) that subsequently guides the optimal policy learning. More importantly, to estimate the rewards and thereby the optimal policy parameters, the critical {\it reward modeling} stage involves human annotators choosing between pairs of candidate answers \( (y_1, y_2) \), indicating their preferences.\footnote{This framework can be extended to rank multiple responses using the Plackett-Luce model.} As such, typical alignment methods assume that $p(y_w \succ y_l | x)$ (the human annotations of preference) is equivalent to $p^*(y_1 \succ y_2 | x)$ or any ranking or choice thereby established with the human decisions. However, prospect theory and empirical studies in rational choice theory suggest that human preferences are often stochastic, intransitive, and can fluctuate across time and contexts \citep{tversky1969intransitivity, von2005welfare, regenwetter2011transitivity}.

Existing direct alignment methods, such as DPO-based supervised alignment, assume access to deterministic preference labels, disregarding the inherent variability in human judgments, \textit{even when popular preference datasets are inherently annotated with such variability, noise, or ``non-deterministic preferences'' given their provenance in human labeling.} More importantly, such implicit trust in the preference data by DPO-like algorithms, without explicit instance-level penalization on the loss, can cause policies that are trained to deviate from true intentions of human preference learning (see \Cref{lemma:dpo_non_deterministic} and \Cref{lemma:edpo_failure} for details). Additionally, in many datasets, a significant proportion of preference pair annotations display low human confidence, or receive similar scores from a third-party reward assignment models (e.g., GPT-4) despite being textually different, indicating that the two responses are likely semantically similar or similar in intent, content or quality. Note that we consider non-deterministic preference samples to be distinct from noise present as flipped labels~\cite{chowdhury2024provably, wang2024secretsrlhflargelanguage}, which is typically resolved using label-smoothing based heuristics, data exclusion or prior knowledge of noise coefficients in the data in preference learning.

As with preference learning, such discrepancies in preference signals can similarly derail reward learning and limit reward models from reaching a consensus, even with majority voting with reward ensembles~\citep{wang2024secretsrlhflargelanguage}. 
These cases reflect the stochastic nature of human choices, and challenge the assumption of stable, deterministic preferences in alignment frameworks.


We now formally define such ``noisy'' or non-deterministic preference labels in offline finite preference data regimes and offer some insights into limitations of current approaches like DPO and e-DPO. For the sake of analysis, we still consider a Bradley-Terry-based modeling to represent such preference signals. All proofs are deferred to the appendix.


\vspace*{-1mm}
\begin{assumption}
\label{assumption:main_offline_assumption}
Let \( \mathcal{D}_{\text{pref}}= \{ (\prompt, \win, \lose) \}_{i=1}^{N} \) be an offline dataset of pairwise preferences with sufficient coverage, where each \( \prompt \) is a prompt, and \( \win \) and \( \lose \) are the corresponding preferred and dispreferred responses, respectively. Let \( r^*(x, y) \in \mathbb{R} \) be an underlying true reward function that is deterministic\footnote{
Deterministic and non-deterministic preferences are only defined on the true preference distribution \( p^* \) and should \textit{not} be confused with the empirical probabilities or confidence assigned by the policy. The use of "deterministic" here is simply to imply that the true reward function \( r^*(x, y) \) is finite and scalar.} and finite everywhere. Let \( \pi_{\theta^*}(y \mid x) \) be the learned model and \( \pi_{\text{ref}}(y \mid x) \) the reference, with \(\text{supp}(\pi_{\text{ref}}) = \mathcal{Y}\). Assume \(\text{supp}(\rho) = \text{supp}(\mu) \times \mathcal{Y} \times \mathcal{Y}\), where \(\mathcal{Y}\) is the space of all responses, \( \rho \) is the data distribution, and \( \mu \) is the prompt or context distribution.
\end{assumption}

\vspace*{-1mm}
\begin{proposition}[Non-Deterministic Preferences]
\label{prop:non_deterministic}
Define the subset \( \mathcal{D}_{\text{nd}} \subset \mathcal{D}_{\text{pref}} \) with non-deterministic preferences as \( \mathcal{D}_{\text{nd}} = \{ (x, y, y') \in \mathcal{D}_{\text{pref}} \mid  P(y \succ y' \mid x) \approx \frac{1}{2} \} \), where \( |\mathcal{D}_{\text{nd}}| \ll N \). Under~\cref{assumption:main_offline_assumption} and antisymmetric preferences~\citep{munos2023nash}, the preference relation is given by \( P(\win \succ \lose \mid \prompt) = 1 - P(\lose \succ \win \mid \prompt) \). Given non-deterministic preferences, i.e., \( (y, y') \in \mathcal{D}_{\text{nd}} \), the Bradley-Terry model assigns \( \Delta r = 0 \), implying \( r^*(x, y) = r^*(x, y') \). Thus, \(\forall (x, y, y') \in \mathcal{D}_{\text{nd}}, \ \Delta r = 0\). See~\Cref{proof:zero_reward_diff} for a complete derivation.\footnote{For clarity, we note that this Proposition 2 is distinct from Proposition 2 from \citet{munos2023nash} that is referenced above.}
\end{proposition}



\vspace*{-1mm}
\begin{lemma} 
\label{lemma:dpo_non_deterministic}
  Under Proposition~\ref{prop:non_deterministic}, a) the DPO implicit reward difference in its objective \( \frac{\pi_{\theta^*}(y) \pi_{\mathrm{ref}}(y')}{\pi_{\theta^*}(y') \pi_{\mathrm{ref}}(y)} \to 1 \), that leads to the policy empirically underfitting the preference distribution. b) For \( |\mathcal{D}_{\text{nd}}| \ll N \) where $N$ is finite, if DPO estimates that $p^*(y \succ y') = 1$, then  \( \frac{\pi_{\theta^*}(y) \pi_{\mathrm{ref}}(y')}{\pi_{\theta^*}(y') \pi_{\mathrm{ref}}(y)} \to \infty  \). c) For all minimizers \( \pi_{\theta^*} \) of the DPO objective (\Eqref{eq:dpo}), it follows that $\pi_{\theta^*}(y_l)  \to 0 $ and \(\pi_{\theta^*}(\mathcal{C}(y_l)^c) \to 1\), where \(\mathcal{C}(y_l)^c\) denotes the complement of the set of dispreferred responses \( \lose \), \(\forall i \in \mathbb{N}\). 
\end{lemma}

Given non-trivial occurrences of non-deterministic preference pairs in typical preference learning datasets, a consequence of \Cref{lemma:dpo_non_deterministic} is that DPO's learned optimal policy can effectively assign non-zero or even very high probabilities to tokens that never appear as preferred in the training data, causing substantial policy degeneracy. Moreover, as noted and shown emprirically in previous work~\citep{azar2023general,pal2024smaug}, DPO effectively underfits the preference distribution because its empirical preference probabilities (RHS of \Eqref{eq:dpo} without the expectation) are only estimates of the true preference probabilities, especially when \( p^*(y \succ y') \in \{0, 1\} \). A noteworthy implication of \Cref{lemma:dpo_non_deterministic} is that this weak regularization effect of DPO can theoretically assign very high probabilities to the complement set of dispreferred tokens that never appear in the training data {\it at all}, especially when \( |N_{\text{nd}}| \ll N \) for finite data regimes. 
In realistic settings where non-deterministic preferences constitute a non-trivial proportion of data, \Cref{lemma:dpo_non_deterministic} additionally implies that DPO leads to unstable updates and inconsistent policy behavior, where the gradient update is effectively cancelled out for these samples since the log probabilities of both the winning and the losing responses are roughly equal ($\Delta r \approx 0$), so the scaled weighting factor (sigmoid of implicit reward differences) does not contribute as much as when \( p^*(y \succ y') \in \{0, 1\} \). As stated in Sec.~\ref{sec:motivation}s, with DPO, $\pi_{\theta^*}$ not only sees less of this type of preference but also fails to adequately regularize when it does.


A solution to the above limitations of DPO within offline settings is to recast its MLE optimization objective into a \textit{regression} task, where the choice of regression target can be the preference labels themselves (as in IPO;~\citet{azar2023general}) or reward differences (as in e-DPO;~\citet{fisch2024robustpreferenceoptimizationreward}). While the former directly utilizes preference labels, regressing the log-likelihood ratio \( \pi_{\text{ratio}}\) to the KL-$\beta$ parameter as defined in \Eqref{eq:dpo}, the latter extends IPO by regressing against the difference in true rewards \( r^*(x, y) \), independent of explicit preference labels and acting as a strict generalization of the IPO framework. Notably, both these methods ensure that the resulting policy induces a valid Bradley-Terry preference distribution \( p^*(y_1 \succ y_2 \mid x) > 0, \, \forall x, y_1, y_2 \in \mu  \times \mathcal{Y} \times \mathcal{Y} \).


However, these approaches have inherent limitations. IPO regresses the log-likelihood difference on a Bernoulli-distributed preference label, failing to capture nuanced strength in relative preferences. Conversely, e-DPO eliminates preference label dependence but sacrifices the granular signals available in preference data, instead over-relying on the quality of reward ensembles, which may still lead to over-optimization~\citep{eisenstein2024helping}.\footnote{Furthermore, the use of reward ensembles in e-DPO introduces significant computational overhead, potentially limiting its broader applicability due to increased resource requirements.} Consider the following lemma that derives from \cref{assumption:main_offline_assumption} and \cref{prop:non_deterministic}:

\vspace*{-1mm}
\begin{lemma} 
\label{lemma:edpo_failure}
Under Proposition~\ref{prop:non_deterministic} and in the spirit of~\citet{fisch2024robustpreferenceoptimizationreward}'s argument, using e-DPO alignment over non-deterministic preference pairs
leads to \(\frac{\pi_{\theta^*}(y) \pi_\mathrm{ref}(y')}{\pi_{\theta^*}(y') \pi_\mathrm{ref}(y)} \to \infty \) for \( (y, y') \in \mathcal{D}_{nd} \) where \( y = \win \)  and \( y' = \lose \), \(\forall i \in \mathbb{N}\). Then, for all minimizers \( \pi_{\theta^*} \) of the e-DPO objective:
\begin{equation}
    \mathcal{L}_{\mathrm{distill}}(r^*, \pi_\theta; \rho) = \mathbb{E}_{\rho(x, y_1, y_2)} \Big[ \Big(r^*(x, y_1) - r^*(x, y_2) - \\ \beta \log \frac{\pi_\theta(y_1\mid x) \piref(y_2 \mid x)}{\pi_\theta(y_2 \mid x) \piref(y_1\mid x)}\Big)^2\Big],
\end{equation} 
it follows that \(\pi_{\theta^*}(\mathcal{C}(y_l)^c) \to 1\) with \( 0 < \pi_{\theta^*}(\win) \leq 1 \), \(\forall i \in \mathbb{N}\), where \(\mathcal{C}(y_l)^c\) denotes the complement of the set of dispreferred responses \( \lose \), \(\forall i \in \mathbb{N}\).
\end{lemma}

Our core insight in proposing DRDO is that modeling relative preference strengths during the {\it policy learning stage, particularly at the extrema of the preference distribution}, is only problematic if one uses a DPO-like MLE loss formulation that maximizes implicit reward differences. On the other hand, the MLE formulation for the {\it reward modeling} stage does not suffer from this limitation precisely because estimated rewards are scalar quantities with no likelihood terms within the log-sigmoid term (as in \Eqref{eq:RM_loss}), provided there is enough coverage in the preference data. Since both stages rely on a finite preference dataset with various levels of preference strengths (that mirrors human preferences), one can combine the two stages by explicitly distilling rewards into the policy learning stage. Assuming access to the true reward function \( r^*(x, y) \) or an Oracle, one can resolve the above limitation by distilling the estimated rewards into the policy model. This intuitively avoids DPO's underfitting to extremal preference strengths: since the same preference data is used for reward distillation and policy learning, this offline distillation ensures that the policy stays within the data distribution during alignment.

\subsection{Proof of Non-Deterministic Preference Relations with Reward Differences}
\label{proof:zero_reward_diff}
\begin{proof}
From \citet{munos2023nash}, we assume the preference relation is antisymmetric: \( P(y_1 \succ y_2) = 1 - P(y_2 \succ y_1) \). Thus, under non-deterministic preference relations, \( P(y_w \succ y_l) \approx P(y_l \succ y_w) \). Since the true preference distribution \( p^* \) is latent and unobserved, we consider the expectation over the subset of non-deterministic samples \( \mathcal{D}_{\text{nd}} \subset \mathcal{D}_{\text{pref}} \) and express this preference relation using a Bradley-Terry (BT)~\citep{BradleyTerry1952} formulation.

Given this , the probability of preferring response \( y_w \) over \( y_l \) given context \( x \) is expressed as:
\begin{equation}
\mathbb{E}_{(x, y_w, y_l) \sim \mathcal{D}_{\text{nd}}}[P(y_w \succ y_l \mid x)] = \mathbb{E}_{(x, y_w, y_l) \sim \mathcal{D}_{\text{nd}}}[\sigma(r(x, y_w) - r(x, y_l))],
\end{equation}
where \( \sigma(\cdot) \) denotes the sigmoid function, and \( r(x, y_w) \) and \( r(x, y_l) \) represent the implicit rewards for the winning and losing responses, respectively. Recall that the BT reward formulation under the  broad Plackett-Luce family models~\citep{plackett1975analysis} is characterized by \textit{underspecification}. As such, it does not impose restrictions on the reward function form, provided it satisfies equivalence relations, i.e., rewards are defined up to a prompt-dependent shift (Definition 1 in~\citet{rafailov2024direct}). Consequently, the expected reward differences still adhere to~\cref{prop:non_deterministic} \textit{without}  necessarily having BT-motivated DPO's implicit reward formulation.

Now, writing the complementary preference probability relation as:
\begin{equation}
\mathbb{E}_{(x, y_w, y_l) \sim \mathcal{D}_{\text{nd}}}[P(y_l \succ y_w \mid x)] = \mathbb{E}_{(x, y_w, y_l) \sim \mathcal{D}_{\text{nd}}}\left[1 - \sigma(r(x, y_w) - r(x, y_l))\right],
\end{equation}
With some algebra, this can be conveniently rewritten as:
\begin{equation}
\mathbb{E}_{(x, y_w, y_l) \sim \mathcal{D}_{\text{nd}}}[P(y_l \succ y_w \mid x)] = \mathbb{E}_{(x, y_w, y_l) \sim \mathcal{D}_{\text{nd}}}\left[\frac{e^{-(r(x, y_w) - r(x, y_l))}}{1 + e^{-(r(x, y_w) - r(x, y_l))}}\right].
\end{equation}

Given the non-deterministic preference, where \( P(y_w \succ y_l) \approx P(y_l \succ y_w) \), equating the expected probabilities yields:
\begin{equation}
\mathbb{E}_{(x, y_w, y_l) \sim \mathcal{D}_{\text{nd}}}\left[\frac{1}{1 + e^{-\Delta r}}\right] = \mathbb{E}_{(x, y_w, y_l) \sim \mathcal{D}_{\text{nd}}}\left[\frac{e^{-\Delta r}}{1 + e^{-\Delta r}}\right],
\end{equation}
where \( \Delta r = r(x, y_w) - r(x, y_l) \).

Solving this relation under the assumption that \( \Delta r \) $\in$ (\( -\infty \), \( \infty \)), we get:
\begin{equation}
\mathbb{E}_{(x, y_w, y_l) \sim \mathcal{D}_{\text{nd}}}\left[1\right] = \mathbb{E}_{(x, y_w, y_l) \sim \mathcal{D}_{\text{nd}}}\left[e^{-\Delta r}\right] \quad \Rightarrow \quad \mathbb{E}_{(x, y_w, y_l) \sim \mathcal{D}_{\text{nd}}}[\Delta r] \approx 0.
\end{equation}

Thus, for non-deterministic preferences, the expected reward difference \( \mathbb{E}[\Delta r] \approx 0 \), indicating that the implicit rewards \( r(x, y_w) \) and \( r(x, y_l) \) are nearly equal, under the expectation, where equality naturally holds at the sample level. 
\end{proof}

\subsection{Proof of Lemma 1a and 1b}
\label{app:proof_lemma_1a_1b}
 \begin{proof}

Let us rewrite the Bradley-Terry preference probability equation in terms of the DPO implicit rewards. The BT model specifies this probability of preferring \( y_w \) over \( y_l \) as:

\begin{equation}
\label{eq:BT_preference}
P(y_w \succ y_l \mid x) = \sigma(r^*(x, y_w) - r^*(x, y_l))
\end{equation} where the rewards can be rewritten in term of the DPO implicit rewards $\hat r_w$ and  $\hat r_l$ as:

\begin{align}
\label{eq:BT_preference_implicit_rewards}
P(y_w \succ y_l \mid x) &= \sigma \left( \underbrace{\beta \log \frac{\pi_{\theta^*}(y_w \mid x)}{\pi_{\text{ref}}(y_w \mid x)}}_{\smash{\scriptstyle\text{($\hat r_w)$}}} - \underbrace{\beta \log \frac{\pi_{\theta^*}(y_l \mid x)}{\pi_{\text{ref}}(y_l \mid x)}}_{\smash{\scriptstyle\text{($\hat r_l$)}}} \right) \\
\label{eq:BT_preference_implicit_rewards_pt2}
&= \sigma \left(\beta \log \left( \frac{\pi_{\theta^*}(y_w \mid x) \pi_{\text{ref}}(y_l \mid x)}{\pi_{\theta^*}(y_l \mid x) \pi_{\text{ref}}(y_w \mid x)} \right)\right) \\
\label{eq:BT_preference_implicit_rewards_pt3}
&= \sigma \left(\beta \log \left( \frac{\pi_{\theta^*}(y \mid x) \pi_{\text{ref}}(y' \mid x)}{\pi_{\theta^*}(y' \mid x) \pi_{\text{ref}}(y \mid x)} \right)\right), \quad \forall (x, y, y') \in \mathcal{D}_{\text{nd}} \subset \mathcal{D}_{\text{pref}}  
\end{align}

where the above equation holds for all $\forall (x, y, y') \in \mathcal{D}_{\text{pref}}$, since DPO does not distinguish between the nature of the true preference relations in estimating the true preference probabilities, assuming $(y, y')$ appear as preferred and dispreferred responses respectively for any context $x$. Since this RHS of \Eqref{eq:BT_preference_implicit_rewards_pt3} is simply the sigmoided difference of implicit rewards assigned by DPO to estimate the true preference probabilities, it is straightforward to see from \cref{prop:non_deterministic} that the RHS i.e., 
\(\frac{\pi_{\theta^*}(y)\pi_{\mathrm{ref}}(y')}{\pi_{\theta^*}(y') \, \pi_{\mathrm{ref}}(y)} \to 1\) and  $P(y_w \succ y_l \mid x)$ $\sim \frac{1}{2}$, as per our definition of non-deterministic preferences. Since the reference model $\pi_{\mathrm{ref}}$ is assumed to have full support over the output space (\(\text{supp}(\pi_{\text{ref}}) = \mathcal{Y}\)) and is not updated and can be set to a uniform prior (\(\pi_{\text{ref}} \sim \mathcal{U}(\mathcal{Y})\))~\cite{xu2024contrastive}, without losing any generality, this implies that \(\frac{\pi_{\theta^*}(y)}{\pi_{\theta^*}(y')}\) must remain close to 1 to satisfy this constraint (\(\frac{\pi_{\theta^*}(y) \, \pi_{\mathrm{ref}}(y')}{\pi_{\theta^*}(y')\pi_{\mathrm{ref}}(y)} \to 1\)). In this case, the policy tends to underfit the preference distribution since the preference signals are weak and policy cannot distinguish between the preferred and the dispreferred response. 

Similar to~\citet{azar2023general}'s argument, we can argue here that in this case when true preference probabilities are $\sim \frac{1}{2}$, i.e., non-deterministic, DPO's empirical reward difference estimates actually tend toward 1 which leads to underfitting of the optimal policy $\pi_{\theta^*}$ during alignment. Indeed, in this case, the $\beta$ parameter does not provide any additional regularization effect to prevent policy underfitting especially under finite data. This completes the proof of \Cref{lemma:dpo_non_deterministic}a. 
\end{proof}

We can similarly prove \Cref{lemma:dpo_non_deterministic}b in the case when \(|\mathcal{D}_{\text{nd}}| \ll N \) where \( N \) is assumed to be finite. In this case, in a similar vein as~\citet{azar2023general},  there is more likelihood that DPO sigmoided reward difference estimates are 1, i.e., 
\(r^*(x, y_w) - r^*(x, y_l) \to \infty\).

As such, from \Eqref{eq:BT_preference}, it is straightforward to see that the DPO's implicit reward difference tends to infinity, regardless of the strength of the $\beta$ parameter, as shown below:

\[
\log \left( \frac{\pi_{\theta^*}(y \mid x) \pi_{\text{ref}}(y' \mid x)}{\pi_{\theta^*}(y' \mid x) \pi_{\text{ref}}(y \mid x)} \right) \to \infty
\] in \Eqref{eq:BT_preference_implicit_rewards_pt3}.This implies that 
\[
\frac{\pi_{\theta^*}(y \mid x) \pi_{\text{ref}}(y' \mid x)}{\pi_{\theta^*}(y' \mid x) \pi_{\text{ref}}(y \mid x)} \to \infty.
\]  


\subsection{Proof of Lemma 1c}
\label{app:proof_lemma_1c}
 \begin{proof}
Our proof follows the argumentation in~\citet{fisch2024robustpreferenceoptimizationreward}. 
Assume for now that all preference samples \((y, y')\) $\in \mathcal{D}_{\text{pref}}$, including non-deterministic preference pairs, are mutually exclusive. Then, for the DPO objective (\Eqref{eq:dpo}) to be minimized, each \(\theta_y\) must correspond uniquely to \(y\),  where \(\theta_y\) are the optimal parameters that minimize the DPO objective in each such disjoint preference pair. This implies that DPO objective over \(\mathcal{D}_\mathrm{pref}\) is convex in the set \(\Lambda = \{\lambda_1, \ldots, \lambda_n\}\), where
\begin{equation}
    \lambda_i = \beta \log \left( \frac{\pi_\theta(y_w^{(i)}) \pi_\mathrm{ref}(y_l^{(i)})}{\pi_\theta(y_l^{(i)}) \pi_\mathrm{ref}(y_w^{(i)})} \right), \quad \forall i \in \mathbb{N}
    \label{eq:lambda_deterministic}
\end{equation}

Now, consider the non-deterministic preference samples indexed by \( j \), which belong to the set \( \mathcal{D}_{\text{nd}} \subset \mathcal{D}_{\text{pref}} \). Let \( j \in \{k+1, \ldots, N\} \) with the assumption \( k \gg (N - k) \). Under the mutual exclusivity assumption, \Eqref{eq:lambda_deterministic} must also hold true for the non-deterministic preference samples. Consequently, we can rewrite \Eqref{eq:lambda_deterministic} as:

\begin{equation}
    \lambda_j = \beta \log \left( \frac{\pi_\theta(y_w^{(j)} \mid x) \pi_\mathrm{ref}(y_l^{(j)} \mid x)}{\pi_\theta(y_l^{(j)} \mid x) \pi_\mathrm{ref}(y_w^{(j)} \mid x)} \right), \quad \forall j \in \mathcal{D}_{\text{nd}}
    \label{eq:lambda_nondeterministic}
\end{equation}

More specifically, for every \(j\), the following holds at the limit for the DPO objective to converge:
\begin{equation}
    \lim_{\lambda_j \to \infty} -\log \left(\sigma\left(\lambda_j\right) \right) = 0,
\end{equation}
 
which implies that \(\Lambda^* = \{\infty\}^N\) induces a set of global minimizers of the DPO objective that includes \(\theta^*\) that are optimal for the set of non-deterministic preference samples, while inducing a parallel set of \(\theta^*\) at convergence for deterministic samples.

Consequently, all global minimizers \(\theta^*\) including those optimal on the non-deterministic samples must satisfy
\begin{equation}
\label{eq:global_minimizer_equation}
    \log \frac{\pi_\theta(y_w^{(j)}) \pi_\mathrm{ref}(y_l^{(j)})}{\pi_\theta(y_l^{(j)}) \pi_\mathrm{ref}(y_w^{(j)})} = \infty.
\end{equation}

    Since \(0 < \pi_{\text{ref}}(y) < 1\) for all \( y \), \(\theta^*\) must satisfy
\begin{equation}
    \frac{\pi_{\theta^*}(y_w^{(j)})}{\pi_{\theta^*}(y_l^{(j)})} = \infty,
\end{equation}
implying \(\pi_{\theta^*}(y_l^{(j)}) = 0\) and \(\pi_{\theta^*}(y_w^{(j)}) > 0\) for all \( i \in N \), given that \(\pi_{\theta^*}(y_w^{(j)}) \leq 1\) for any \( y_w^{(j)} \). Alternatively, let us define the complement of the aggregated representation of all the dispreferred responses $y_l^{(j)}$ i.e., $ \mathcal{\phi}(y_l)^c $, where $\mathcal{\phi}(y_l)$  is the aggregation function. We thereby have, 

\begin{equation}
    \mathcal{\phi}(y_l) = \{ y \colon \exists j \in \mathbb{N} \text{ such that } y_l^{(j)} = y \},
\end{equation}

Under these conditions, it is clear that $\pi_{\theta^*}$ must assign the entire remaining probability mass to $ \mathcal{\phi}(y_l)$ as given below, 
\begin{align}
    \pi_{\theta^*}(\mathcal{C}(y_l)) &= \sum_{y \in \mathcal{\phi}(y_l)} \pi_{\theta^*}(y) = 0 \\
    &\implies \pi_{\theta^*}(\mathcal{\phi}(y_l)^c) = 1.
\end{align}
This completes the proof of \Cref{lemma:dpo_non_deterministic}c and thus \Cref{lemma:dpo_non_deterministic}.

\end{proof}

\subsection{Proof of Lemma 2}
\label{app:proof_lemma2_edpo}
\begin{proof}
    
 Let us first rewrite the e-DPO's distillation objective~\citep{fisch2024robustpreferenceoptimizationreward} over the preference dataset $\mathcal{D}_{\text{pref}}$ and examine how the optimal policy $\pi_\theta$ behaves upon convergence of this objective. For simplicity of analysis, we only consider the point-wise reward based distillation loss in the e-DPO formulation without\footnote{Note that in our empirical experiments we use the full e-DPO objective with a set of three reward models.} considering reward ensembles. Note that the e-DPO objective does not require preference labels and can apply to any response pair.

\begin{align}
\label{eq:edpo_proof_main}
 \mathcal{L}_{\mathrm{distill}}(r^*, \pi_\theta) &= \mathbb{E}_{(x, y_1, y_2) \sim \mathcal{D}_{\text{pref}}} \left[ \left( r^*(x, y_1) - r^*(x, y_2) - \beta \log 
 \frac{\pi_\theta(y_1 \mid x) \pi_{\text{ref}}(y_2 \mid x)}{\pi_\theta(y_2 \mid x) \pi_{\text{ref}}(y_1 \mid x)} \right)^2 \right] \\
 &= \mathbb{E}_{(x, y_1, y_2) \sim \rho}\left[ \left(r^*(x, y_1) - r^*(x, y_2) - \beta \log 
    \frac{\pi_\theta(y_1 \mid x)}{\pi_\theta(y_2 \mid x)} + \beta \log 
    \frac{\pi_{\text{ref}}(y_1 \mid x)}{\pi_{\text{ref}}(y_2 \mid x)}\right)^2\right].
\end{align}

As \(\pi_\theta\) converges to the optimal policy \(\pi_{\theta^*}\), the distillation objective \(\mathcal{L}_{\mathrm{distill}}\) should ideally approach zero and can be expressed as, 
\[
\lim_{\pi_\theta \to \pi_{\theta^*}} \mathcal{L}_{\mathrm{distill}}(r^*, \pi_\theta) = 0
\]

With some slight algebraic rearrangement and substituting the optimal policy \(\pi_{\theta^*}\) for \(\pi_\theta\) at convergence, we get:

\[
r^*(x, y_1) - r^*(x, y_2) = \beta \log \frac{\pi_{\theta^*}(y_1 \mid x)}{\pi_{\theta^*}(y_2 \mid x)} - \beta \log \frac{\pi_{\text{ref}}(y_1 \mid x)}{\pi_{\text{ref}}(y_2 \mid x)}.
\]
Since \((y_1, y_2)\) represents \textit{any} response pair without a preference label, we can substitute them with \((y, y') \in \mathcal{D}_{\text{nd}} \subset \mathcal{D}_{\text{pref}}\), where \(y = y_w\) and \(y' = y_l\). Without losing any generality, we can now rewrite the above equation as, 
 \begin{align}
\label{equation1}
r^*(x, y) - r^*(x, y') &= \beta \log \frac{\pi_{\theta^*}(y \mid x)}{\pi_{\theta^*}(y' \mid x)} - \beta \log \frac{\pi_{\text{ref}}(y \mid x)}{\pi_{\text{ref}}(y' \mid x)} \\
\label{equation2}
&= \beta\log \left( \frac{\pi_{\theta^*}(y \mid x) \pi_{\text{ref}}(y' \mid x)}{\pi_{\theta^*}(y' \mid x) \pi_{\text{ref}}(y \mid x)} \right)  
\end{align}


Now, recall from our proof of \Cref{lemma:dpo_non_deterministic}b where we show that the RHS term of the above equation \(\log \left( \frac{\pi_{\theta^*}(y \mid x) \pi_{\text{ref}}(y' \mid x)}{\pi_{\theta^*}(y' \mid x) \pi_{\text{ref}}(y \mid x)} \right) \to \infty\), which implies that \(\frac{\pi_{\theta^*}(y \mid x) \pi_{\text{ref}}(y' \mid x)}{\pi_{\theta^*}(y' \mid x) \pi_{\text{ref}}(y \mid x)} \to \infty\). 
Indeed, when \(|\mathcal{D}_{\text{nd}}| \ll N \) where \( N \) is finite, unregularized scalar reward estimates of the true preference probabilities can in fact grow exceedingly large in the absence of any other regularization parameters, since $\beta$ by its own does not provide enough regularization as shown in our proof of \Cref{lemma:dpo_non_deterministic}a. Interestingly, similar arguments have also been made in previous works~\citep{azar2023general}. The rest of this proof follows the same argumentation starting \Eqref{eq:global_minimizer_equation} assuming \(0 < \pi_{\text{ref}}(y) < 1\). 
 
This completes the proof of \Cref{lemma:edpo_failure}.
\end{proof}

\subsection{Gradient Derivation of the Focal-Softened Log-Odds Unlikelihood Loss}
\label{app:focal-deriv}

In this section, we derive and analyze DRDO loss gradient and offer insights into how to compared supervised alignment objectives such as DPO~\citep{rafailov2024direct}. Note that we do not analyze the reward distillation component here since it does not directly interact with the focal-softened contrastive log-"unlikelihood" term in training and since it is naturally convex considering its a squared term. 
Let us first rewrite our full DRDO loss, as:

\begin{align} 
\label{eq:distillation_in_proof}
\mathcal{L}_{\mathrm{kd}}(r^*, \pi_\theta) &= \mathbb{E}_{(x, y_1, y_2) \sim \mathcal{D}_{\text{pref}}} \Bigg[ 
\underbrace{\left( r^*(x, y_1) - r^*(x, y_2) - (\hat{r}_1 - \hat{r}_2) \right)^2}_{\text{Reward Difference}} \nonumber \\ 
&\quad - \underbrace{\alpha (1 - p_w)^\gamma \log\left(\frac{\pi_\theta(y_w \mid x)}{1 - \pi_\theta(y_l \mid x)}\right)}_{\text{Contrastive Log-"unlikelihood"}}
\Bigg],
\end{align}

where \( p_w = \sigma(z_w - z_l) = \frac{1}{1 + e^{-(z_w - z_l)}} \) and quantifies the student policy's confidence in correctly assigning the preference from \( z_w = \log \pi_\theta(y_w \mid x) \) and \( z_l = \log \pi_\theta(y_l \mid x) \), or the log-probabilities of the winning and losing responses, respectively.

Without the expectation, consider only the focal-softened log-odds unlikelihood loss given by:

\begin{equation}
 -\alpha \cdot (1 - p_w)^\gamma \cdot \log\left(\frac{\pi_\theta(y_w \mid x)}{1 - \pi_\theta(y_l \mid x)}\right),
\end{equation}  

Taking the gradient of this term with respect to the model parameters $\theta$ and using $\sigma'(x) = \sigma(x) \left( 1 - \sigma(x) \right)$, we derive:

\begin{align}
    \nabla_\theta \mathcal{L}_{\mathrm{kd}} = &\ \alpha \gamma (1 - p_w)^{\gamma - 1} p_w (1 - p_w) (\nabla_\theta z_w - \nabla_\theta z_l) \cdot \log\left(\frac{\pi_\theta(y_w \mid x)}{1 - \pi_\theta(y_l \mid x)}\right) \\
    & - \alpha (1 - p_w)^\gamma \left(\frac{\nabla_\theta \pi_\theta(y_w \mid x)}{\pi_\theta(y_w \mid x)} + \frac{\nabla_\theta \pi_\theta(y_l \mid x)}{1 - \pi_\theta(y_l \mid x)}\right).
\end{align}

\begin{align}
    \nabla_\theta \mathcal{L}_{\mathrm{kd}} = &\ \alpha \gamma (1 - p_w)^{\gamma} p_w   (\nabla_\theta z_w - \nabla_\theta z_l) \cdot \log\left(\frac{\pi_\theta(y_w \mid x)}{1 - \pi_\theta(y_l \mid x)}\right) \\
    & - \alpha (1 - p_w)^\gamma \left(
    \underbrace{\frac{\nabla_\theta \pi_\theta(y_w \mid x)}{\pi_\theta(y_w \mid x)}}_{\text{increase } \pi_\theta(y_w \mid x)}
    + 
    \underbrace{\frac{\nabla_\theta \pi_\theta(y_l \mid x)}{1 - \pi_\theta(y_l \mid x)}}_{\text{decrease } \pi_\theta(y_l \mid x)}
    \right).
\end{align}

While this above equation might appear rather cumbersome, notice that in preference learning in language models, the output token space $\mathcal{Y}$ is exponentially large. Additionally, in typical bandit settings, we consider the entire response itself as the action (summation of log probabilities). Since the modulating term $0 \leq (1 - p_w)^\gamma \leq 1$ and $\alpha$ is typically small $\sim 0.1 $~\citep{lin2018focallossdenseobject, yi2020focal} compared to gradients appearing in likelihood terms appearing above, we can conveniently ignore the first term for the gradient analysis. 

Simplifying the above equation, we get
\begin{equation}
\label{eq:main_drdo_gradient}
    - \alpha (1 - p_w)^\gamma \left(
    \underbrace{\frac{\nabla_\theta \pi_\theta(y_w \mid x)}{\pi_\theta(y_w \mid x)}}_{\text{increase } \pi_\theta(y_w \mid x)}
    + 
    \underbrace{\frac{\nabla_\theta \pi_\theta(y_l \mid x)}{1 - \pi_\theta(y_l \mid x)}}_{\text{decrease } \pi_\theta(y_l \mid x)}
    \right).
\end{equation}

We can now draw some insights and direct comparisons of our approach with Direct Preference Optimization (DPO)~\citep{rafailov2024direct}. As in most contrastive preference learning gradient terms~\citep{rafailov2024direct, Hong2024ORPOMP, xu2024contrastive, meng2024simposimplepreferenceoptimization, Ethayarajh2024KTOMA}, the term \(\frac{\nabla_\theta \pi_\theta(y_w \mid x)}{\pi_\theta(y_w \mid x)}\) in \Eqref{eq:main_drdo_gradient} amplifies the gradient when \(\pi_\theta(y_w \mid x)\) is low, driving up the likelihood of the preferred response \(y_w\). Similarly, \(\frac{\nabla_\theta \pi_\theta(y_l \mid x)}{1 - \pi_\theta(y_l \mid x)}\) penalizes overconfidence in incorrect completions \(y_l\) when $p_w$ is low, encouraging the model to hike preferred response likelihood while discouraging dispreferred ones.

The key insight here is that the modulating term, \((1 - p_w)^\gamma\), strategically amplifies corrections for difficult examples where the probability \( p_w \) of the correct (winning) response is low. Intuitively, unlike DPO's fixed \(\beta\) that is applied across the whole training dataset, this modulating term amplifies gradient updates when preference signals are weak (\( p_w \approx 0.5 \)) and tempering updates when they are strong (\( p_w \approx 1 \)), thus ensuring robust learning across varying preference scenarios. Intuitively, when  (\( p_w \approx 1 \)), the model is already confident of its decision since \( p_w \) remains high, indicating increased model confidence for deterministic preferences. In contrast, when \( p_w \) is small, \((1 - p_w)\) remains near 1, and the term \((1 - p_w)^\gamma\) retains significant magnitude, especially for larger values of \(\gamma\). This allows \(\pi_\theta\) in DRDO to learn from both deterministic and non-deterministic preferences, effectively blending reward alignment with preference signals to guide optimization.

For a deeper intuition, consider the case where the true preference probabilities from $p^* (x,y) \sim \frac{1}{2}$. In this case, since non-deterministic preference samples  are typically low, from \cref{prop:non_deterministic}, DPO would assign zero difference in its implicit rewards, especially for finite preference data. Then as \Cref{lemma:dpo_non_deterministic}(a) suggests, DPO gradients would effectively be nullified, regardless of $\beta$ since the its reward difference range $\in$ $(-\infty, +\infty)$. In this case as \(\pi_\theta\) cannot distinguish between the preference pair and, in turn, effectively misses out on the preference information for such samples.

On the other hand, for \(|\mathcal{D}_{\text{nd}}| \ll N \), if \(\pi_\theta\) in DPO estimates the true preference close to 1 where \(\hat{p} = 1\), as \Cref{lemma:dpo_non_deterministic}(b) suggests, the empirical policy would assign very high probabilities to tokens that do not even appear in the data. This leads to a surprising combination of both underfitting and overfitting, except the overfitting here results in DPO policy generating tokens that are irrelevant to the context.

However, as \Cref{lemma:edpo_failure} suggests, e-DPO does not directly face this limitation \textit{during} training, but the degeneracy manifests upon convergence. Under the same conditions, the DRDO loss still operates at a sample level because if the DRDO estimate of \( p^* \) (via \( p_w \)) is close to its true value of \(\sim \frac{1}{2}\), the modulating factor ensures that gradients do not vanish. This allows DRDO to continue learning from such samples until convergence when winning and losing probabilities are pushed further apart (\Figref{fig:focal_motivation}). Intuitively, the \(\log\left(\frac{\pi_\theta(y_w \mid x)}{1 - \pi_\theta(y_l \mid x)}\right)\) term is minimized precisely under this condition where the modulating term is also close to zero.

\begin{figure}
    \centering
    \includegraphics[width=.5\linewidth,trim={20 15 20 15},clip]{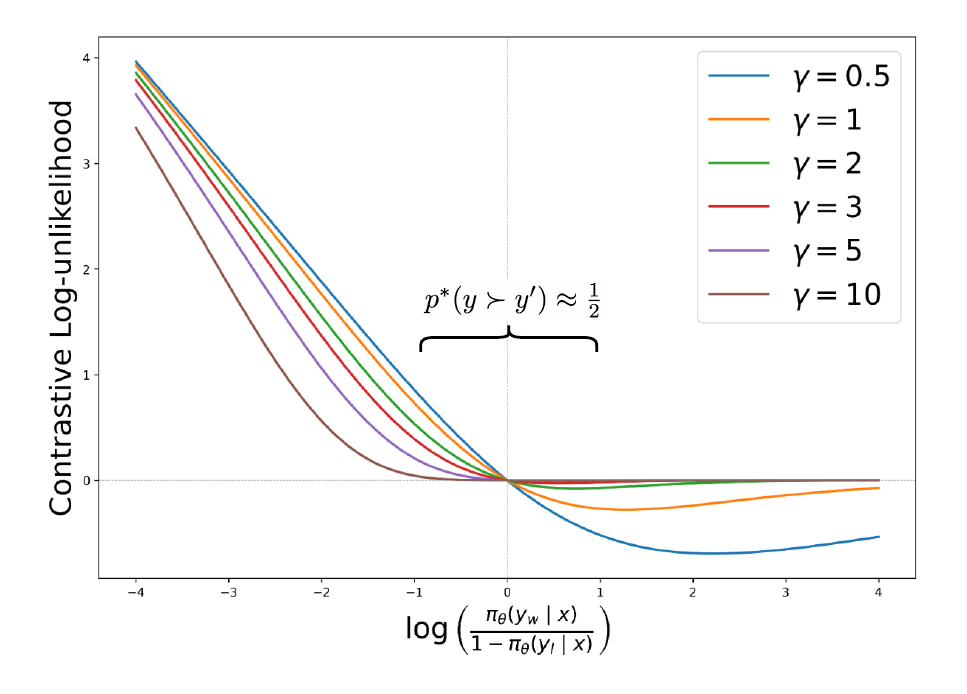}
   \caption{
    Illustration of the DRDO preference loss as a function of the log-unlikelihood ratio 
    across various values of \(\gamma\), the focal modulation parameter. 
    }
    \label{fig:focal_motivation}
\end{figure}

Finally, Table~\ref{tab:drdo_algorithm} provides pseudocode for the DRDO algorithm.

\begin{table}[ht]
\centering
\begin{tabular}{|p{0.95\textwidth}|}
\hline
\textbf{DRDO Algorithm} \\ \hline
\textbf{Input:} Preference dataset \( \mathcal{D}_{\text{pref}} = \{ (\prompt, \win, \lose) \}_{i=1}^{N} \), initialized policy model with reward head \( \pi_{\theta, \theta'} \leftarrow \text{SFT}(\theta) \oplus r_{\theta'} \).\\
\textbf{Output:} Optimized model parameters \( \theta \) in policy \(\pi_{\theta}\).\\
\\
1. Train Oracle \( r_{\phi} \) with loss \(\mathcal{L}_\mathcal{O}(r_{\phi}, \mathcal{D}_{\text{pref}})\) (see Eq.~5).\\
2. For \( t = 1, \ldots, T \):\\
\hspace{5mm} (a) For each \( (\prompt, \win, \lose) \) in \( \mathcal{D}_{\text{pref}} \):\\
\hspace{10mm} i. Compute \( r^*_1 = r_{\phi}(\prompt, \win) \) and \( r^*_2 = r_{\phi}(\prompt, \lose) \).\\
\hspace{10mm} ii. Compute \( \hat{r}_1 = r_{\theta'}(\prompt, \win) \) and \( \hat{r}_2 = r_{\theta'}(\prompt, \lose) \).\\
\hspace{10mm} iii. Compute knowledge distillation loss:
\begin{center}
\scriptsize
$
\mathcal{L}_{\mathrm{kd}}(r^*, \pi_\theta) = \mathbb{E}_{(\prompt, \win, \lose) \sim \mathcal{D}_{\text{pref}}} \Bigg[ 
\underbrace{\left( r^*_1 - r^*_2 - (\hat{r}_1 - \hat{r}_2) \right)^2}_{\text{Reward Difference}} 
- \underbrace{\alpha (1 - p_w)^\gamma \log\left(\frac{\pi_\theta(\win \mid \prompt)}{1 - \pi_\theta(\lose \mid \prompt)}\right)}_{\text{Contrastive Log-"unlikelihood"}}
\Bigg].
$
\end{center}
\\
\hspace{10mm} iv. Update \( \pi_{\theta, \theta'} \) using \(\mathcal{L}_{\mathrm{kd}}\).\\
3. \textbf{Return:} Aligned policy \( \pi_{\theta} \).\\
\hline
\end{tabular}
\caption{DRDO Algorithm steps. We start off with the preference dataset and an SFT-trained policy initialized with an additional linear head parameterized by $\theta'$. Once our oracle is trained, we compute both estimated rewards for each response ($y$) from the initial policy ($\hat{r}$) as well as from the oracle ($r^*$). We then use $\mathcal{L}_{\mathrm{kd}}$ to update both $\theta$ and $\theta'$ in \( \pi_{\theta} \) resulting in our DRDO aligned policies.}
\label{tab:drdo_algorithm}
\end{table}

\section{Further Notes on Experimental Setup}
\label{app:exp-notes}

We provide the following additional explanatory notes regarding the experimental setup:

\begin{itemize}
    \item For non-deterministic and nuanced preferences, note that although we fine-tune all approaches (including DRDO) on \url{https://huggingface.co/datasets/CarperAI/openai_summarize_tldr}, every baseline we use is policy-aligned with only the training data within $\mathcal{D}_{all}$, $\mathcal{D}_{hc,he}$ and $\mathcal{D}_{\ell c,\ell e}$ for a direct comparison.
    \item \citet{pal2024smaug} assume non-determinism of preferences to be correlated to edit-distances between pairwise-samples, but we do not make sure assumptions and consider both the true (oracle) rewards and edit-distances between pairs to verify the robustness of our method. \citet{pal2024smaug}'s theoretical framework brings insights on DPO's suboptimality assumes small edit distance between pairwise samples and they empirically show this primarily for math and reasoning based tasks. In contrast, our evaluation framework is more general in the sense that we consider both the oracle reward difference as well as edit distance in addressing DPO's limitations in learning from non-deterministic preferences and we evaluate on a more diverse set of prompts apart from math and reasoning tasks.
    \item For reward distillation w.r.t to model size, the version of Ultrafeedback we used can be found at \url{https://huggingface.co/datasets/argilla/ultrafeedback-binarized-preferences-cleaned}.
    \item For SFT on both experiments, use the TRL library implementation (\url{https://huggingface.co/docs/trl/en/sft_trainer}) for SFT training on all initial policies for all baselines.
\end{itemize}

\subsection{DRDO and e-DPO Specifics}
\label{app:drdo-edpo}

Although DRDO requires an explicit reward Oracle \( \mathcal{O} \), we fix only one model (based on parameter size and model family) for each experiment. We use Phi-3-Mini-4K-Instruct and OPT 1.3B causal models initialized with a separate linear reward head while retaining the language modeling head weights.\footnote{Similar to~\citet{yang2024regularizinghiddenstatesenables}, we found better generalization in reward learning when our Oracle reward learning loss is regularized with the SFT component (second term in \Eqref{eq:RM_loss_grm}) with an $\alpha$ of 0.01.} 
Fixing the size of \( \mathcal{O} \) allows us to evaluate the extent of preference alignment to smaller models, as in classic knowledge distillation~\citep{gou2021knowledge}. To reproduce the e-DPO~\citep{fisch2024robustpreferenceoptimizationreward} baseline, we train three reward models using \Eqref{eq:RM_loss} with the mentioned base models but with different random initialization on the reward heads.

\subsection{Choice of Evaluation Datasets}
\label{app:dataset-motivation}

As mentioned in \Secref{sec:exp}, our experiments were designed to balance robustness and thoroughness with research budget constraints, and to account for properties of the task being evaluated relative to the data. The rationale for evaluating DRDO's robustness to non-deterministic or ambiguous preferences is straightforward: the TL;DR dataset is annotated with human confidence labels, which enables the creation of the \( \mathcal{D}_{hc,he} \) and \( \mathcal{D}_{\ell c,\ell e} \) splits requires to test the different non-determinism settings. 

Testing the effect of model size on reward distillation does not require human confidence labels, and as the TL;DR training data size is 1.3M rows, testing distillation across 5 models became computationally intractable given available resources. Thus we turned to Ultrafeedback, which at a train size of 61.1k rows made this experiment much more feasible. Additionally, Ultrafeedback is rather better suited to the preference distillation problem because Ultrafeedback was specifically annotated and cleaned \citep{cui2024ultrafeedbackboostinglanguagemodels}, for the purposes of evaluating open source models for distillation. Ultrafeedback also provides a high quality dataset with GPT-4 or scores on both straighforward summarization instruction following {\it and} multiple preference dimensions such as honesty, truthfulness, and helpfulness. Previous work like Zephyr \citep{tunstall2023zephyr} utilize this dataset for evaluating preference distillation. However, their method is more of a data-augmentation method and does not provide any novel distillation algorithms for preferences since they use the DPO loss but with cleaner and diverse feedback data. They only test student models of $\sim$7B parameters, which are still relatively large models that require additional compute for training without including some form of PEFT. As such, Ultrafeedback is best suited to evaluate distillation-based methods and our novel contribution in this area included evaluation of smaller distilled models.

\subsection{TL;DR Summarization Dataset Splits}
\label{sec:tldr_splits}

\begin{table}[h!]
\centering
\caption{\label{tab:tldr_splits}Mean confidence and normalized edit distance statistics our TL;DR preference generalization experiment.}
\begin{tabular}{@{}lcccc@{}}
\toprule[0.7pt]  
\textbf{Split} & \textbf{Conf (Mean)} & \textbf{Edit Dist (Mean)} & \textbf{Train} & \textbf{Validation} \\ 
\midrule
\textbf{{$\mathcal{D}_{all}$}}      & 5.01    & 0.12 & 44,709 & 86,086 \\
{$\mathcal{D}_{hc,he}$}  & 7.31    & 0.15 & 10,136 & 1,127 \\
\textbf{$\mathcal{D}_{\ell c,\ell e}$}  & 2.67    & 0.09 & 10,000 & 1,112 \\
\bottomrule[0.7pt]  
\end{tabular}
\end{table}

Table~\ref{tab:tldr_splits} shows the mean confidence and normalized edit distance statistics in the TL;DR dataset which we used to compute the deterministic and non-determinisitic splits. $\mathcal{D}_{all}$ represents the full data~\citep{stiennon2020learning}. $\mathcal{D}_{hc,he}$ and \textbf{$\mathcal{D}_{\ell c,\ell e}$} represent subsets created by splitting at the 50th percentile of human labeler confidence and edit distance values. The range of labeler confidence values for the full training data range is $[1,9]$. Additionally, segmenting the training data based on the \textit{combination} of confidence and edit distance thresholds do not make the splits roughly half of the full training data. This is because there are many samples that do not simultaneously satisfy the 50th percentile threshold for each metric. In reality, this choice is intentional to more robustly evaluate DRDO under more difficult preference data settings. 
Additionally, previous theoretical as well as empirical work~\citep{pal2024smaug} has shown that supervised methods like DPO fail to learn optimal policies when the token-level similarity is high in the preference pairs, especially in the beginning of the response. Therefore, we apply this combined thresholding for all our experiments on TL;DR summarization dataset.

\vfill

\section{Preference Likelihood Analysis}
 
Fig.~\ref{fig:drdo_eval_plots} shows preference optimization method performance vs. training steps, according to Bradley-Terry (BT) implicit reward accuracies (Fig.~\ref{fig:bt_reward}), Oracle reward advantage (Fig.~\ref{fig:oracle_advantage}), preferred log-probabilities (Fig.~\ref{fig:preferred_probs}) and dispreferred log-probabilities (Fig.~\ref{fig:rejected_probs}). Although all baselines show roughly equal performance in increasing the likelihood of preferred responses ($y_w$), DRDO is particularly efficient in penalizing dispreferred responses ($y_l$) as Fig.~\ref{fig:rejected_probs} suggests.

\begin{figure}[H]
    \centering
    \begin{subfigure}[b]{0.45\textwidth}
        \centering
        \includegraphics[width=\textwidth]{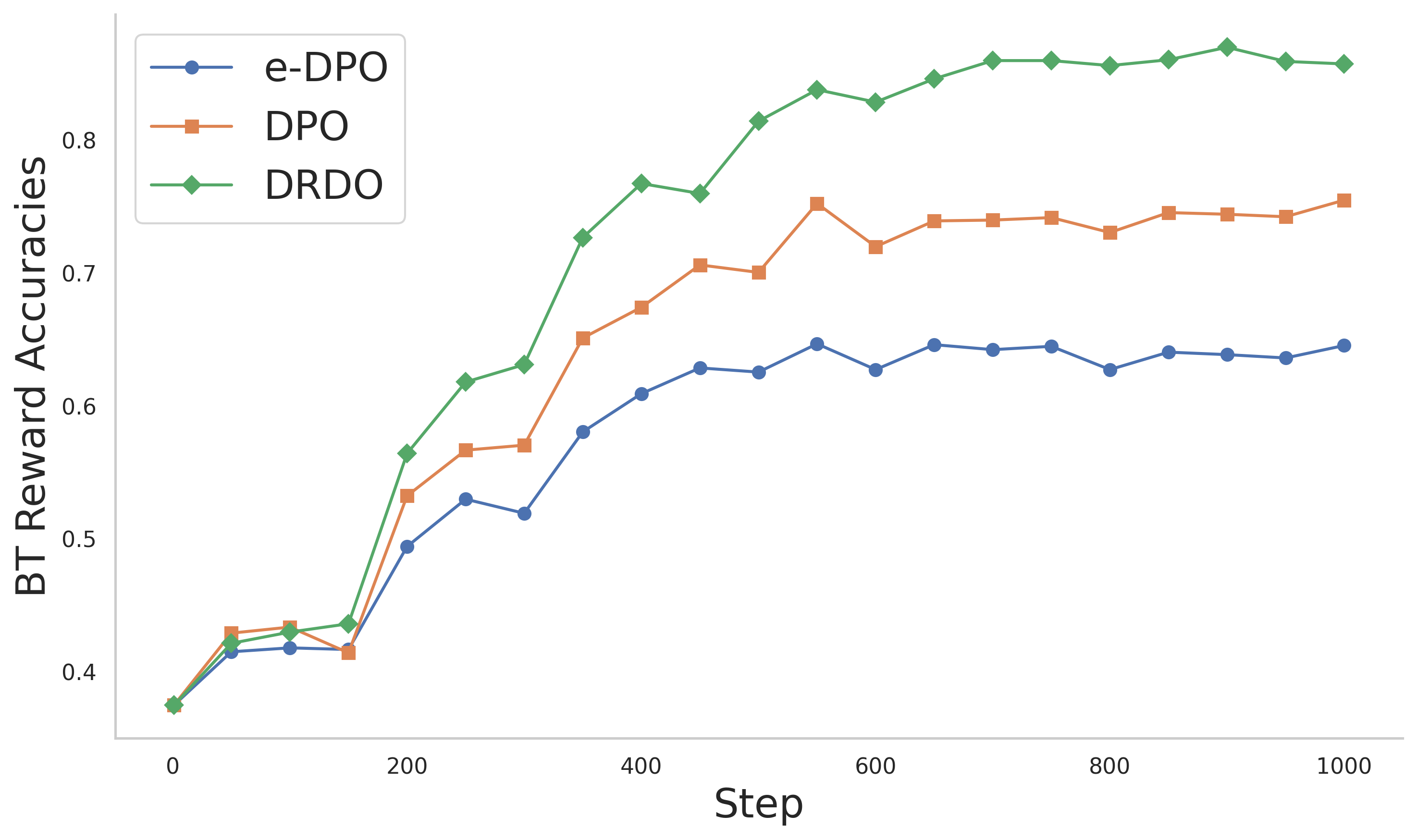}
        \caption{}
        \label{fig:bt_reward}
    \end{subfigure}
    \hfill
    \begin{subfigure}[b]{0.45\textwidth}
        \centering
        \includegraphics[width=\textwidth]{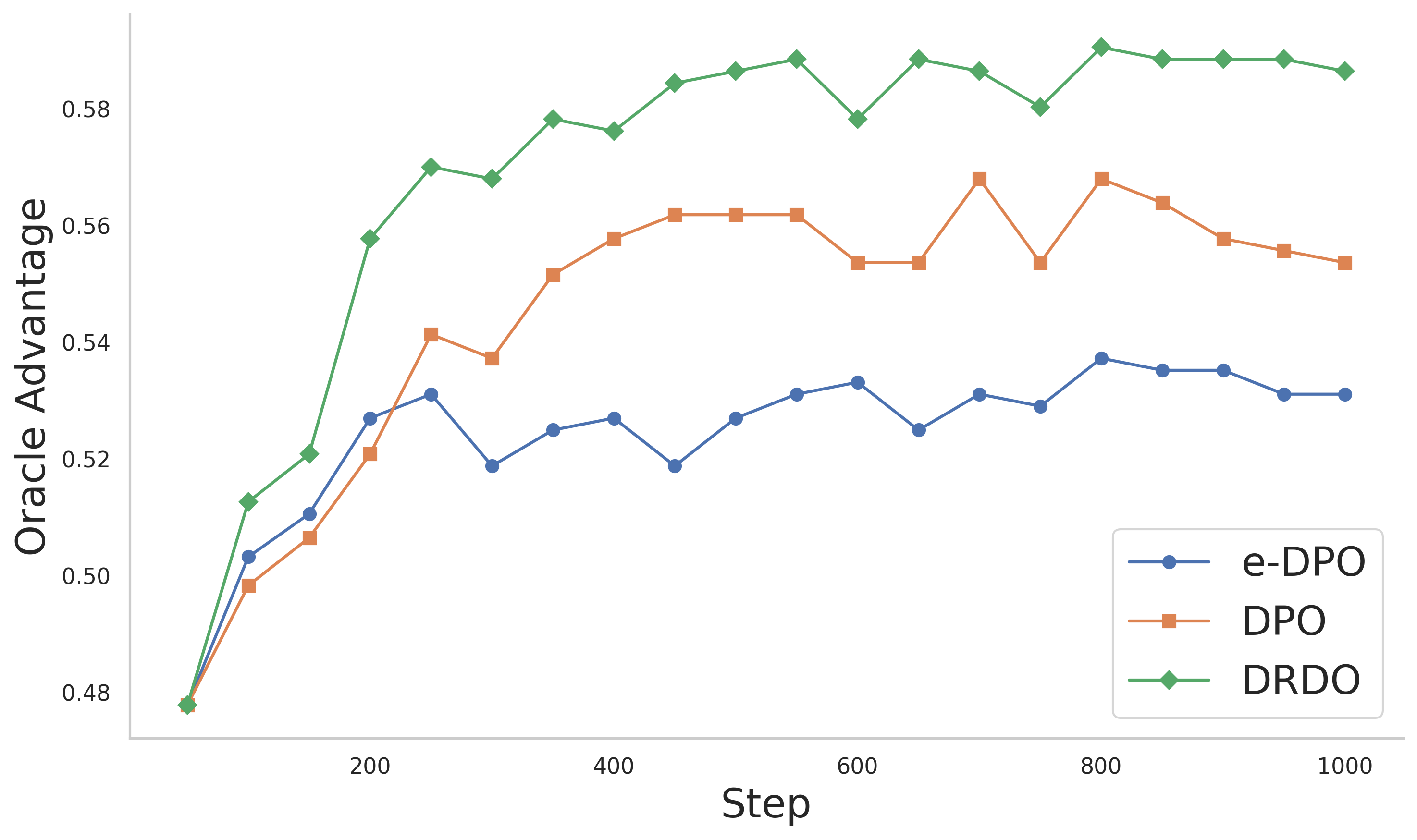}
        \caption{}
        \label{fig:oracle_advantage}
    \end{subfigure}

    \begin{subfigure}[b]{0.45\textwidth}
        \centering
        \includegraphics[width=\textwidth]{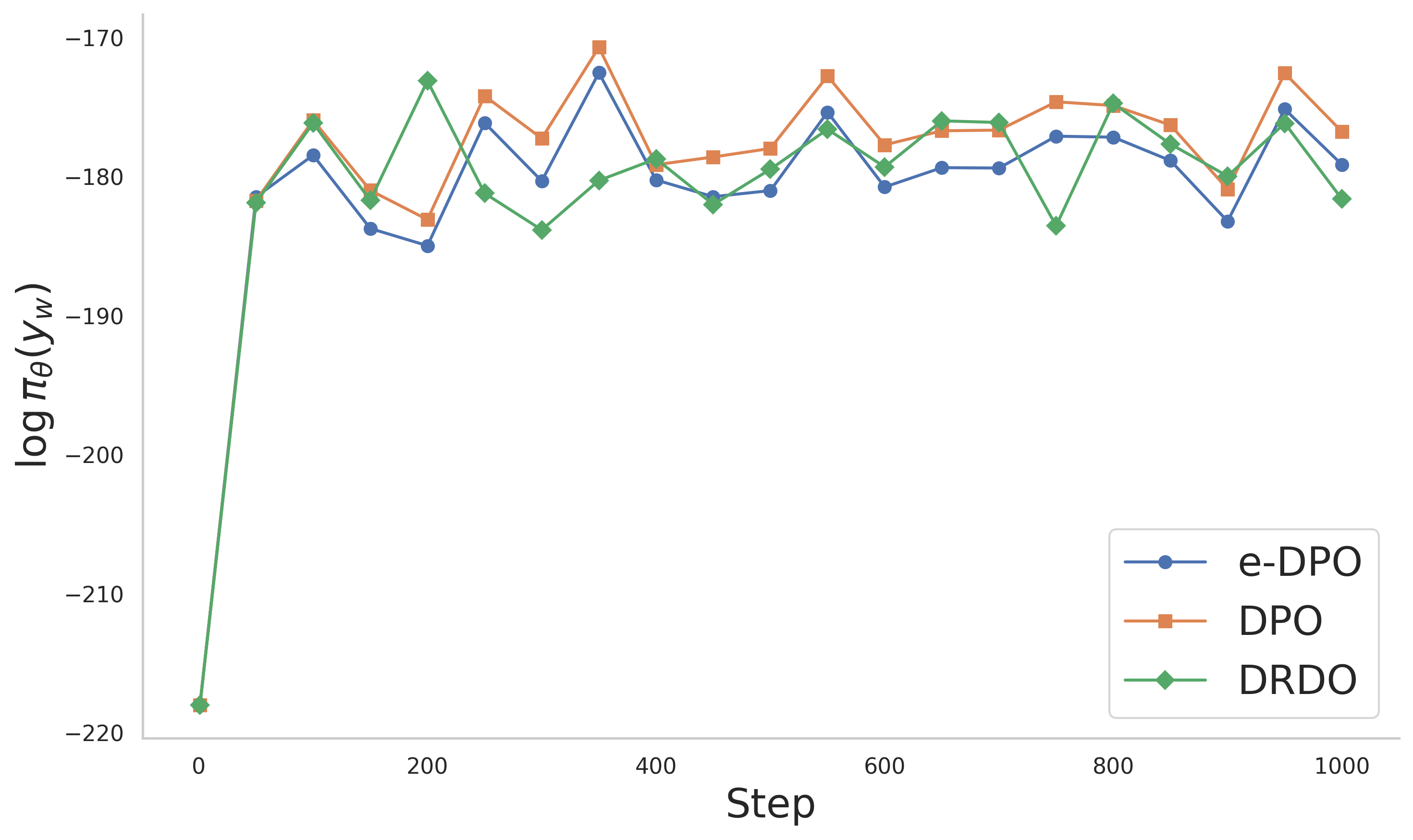}
        \caption{}
        \label{fig:preferred_probs}
    \end{subfigure}
    \hfill
    \begin{subfigure}[b]{0.45\textwidth}
        \centering
        \includegraphics[width=\textwidth]{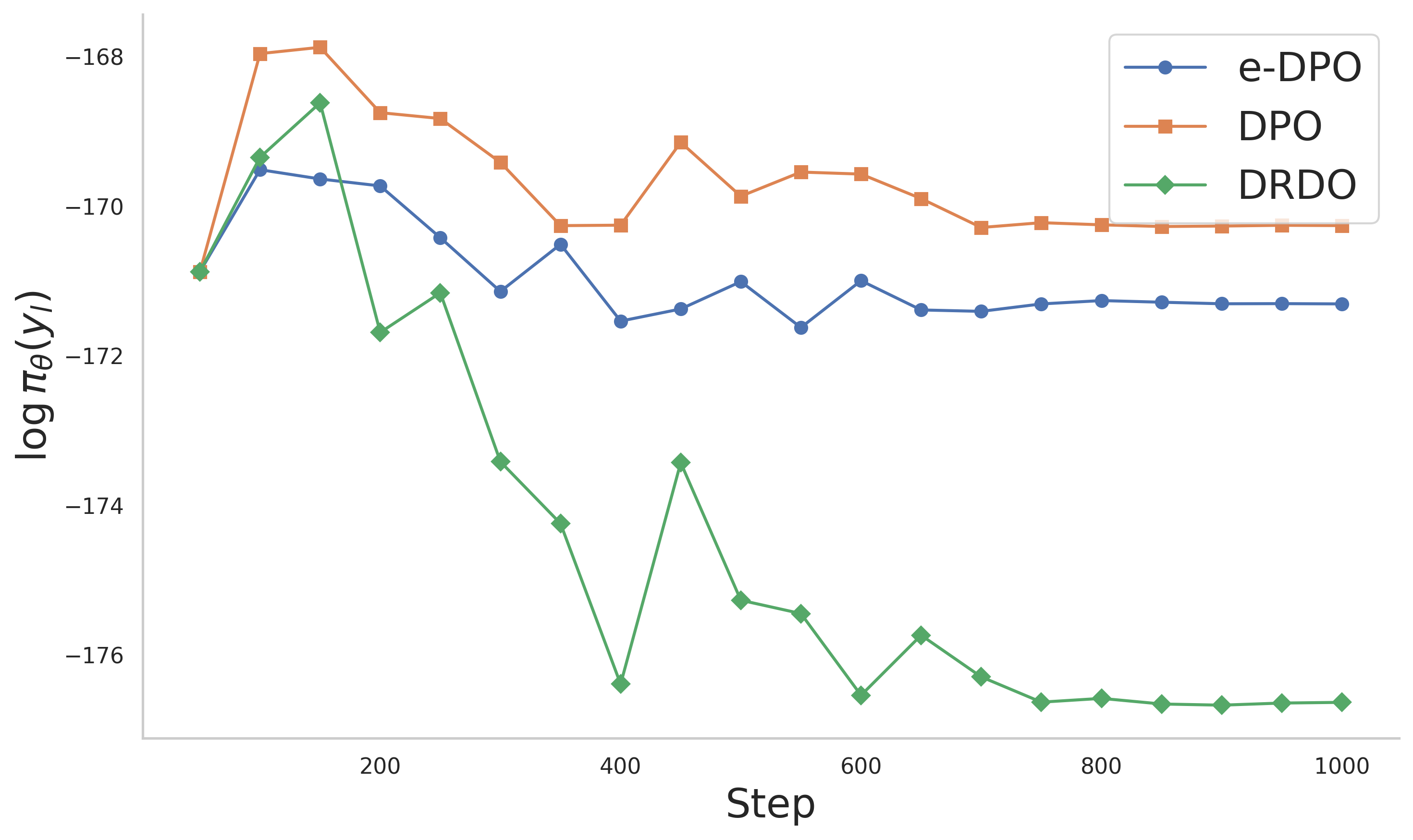}
        \caption{}
        \label{fig:rejected_probs}
    \end{subfigure}

    \caption{Top: DRDO performance evolution during OPT 1.3B training compared to DPO and e-DPO on the evaluation set of Ultrafeedback~\citep{cui2024ultrafeedbackboostinglanguagemodels}, and randomly sampled generations to compute the reward advantage against the preferred reference generations.}
    \label{fig:drdo_eval_plots}
\end{figure}




\section{Hyperparameters}
\label{sec:hyperparameters}

\begin{table}[h!]
\centering
\caption{Model Configuration and Full set of hyperparameter used for DRDO Training}
\label{tab:hyperparameters}
\begin{tabular}{@{}lc@{}}
\toprule
\textbf{Parameter}                        & \textbf{Default Value}       \\ \midrule
learning\_rate                            & 5e-6                         \\
lr\_scheduler\_type                       & cosine                       \\
weight\_decay                             & 0.05                         \\
optimizer\_type                           & paged\_adamw\_32bit          \\
loss\_type                                & DRDO                         \\
per\_device\_train\_batch\_size           & 12                           \\
per\_device\_eval\_batch\_size            & 12                           \\
gradient\_accumulation\_steps             & 4                            \\
gradient\_checkpointing                   & True                         \\
gradient\_checkpointing\_use\_reentrant   & False                        \\
max\_prompt\_length                       & 512                          \\
max\_length                               & 1024                         \\
max\_new\_tokens                          & 256                          \\
max\_steps                                & 20                           \\
logging\_steps                            & 5                            \\
save\_steps                               & 200                          \\
save\_strategy                            & no                           \\
eval\_steps                               & 5                            \\
log\_freq                                 & 1                            \\
 
$\alpha$                   & 0.1                        \\
$\gamma$                         & 2                         \\
                      \\ \bottomrule
\end{tabular}
\end{table}

We only use full-parameter training for all our policy models. We train all our student policies for 1k steps with an effective batch size of 64, after applying an gradient accumulation step of 4. Specifically, both OPT series and Phi-3-Mini-4K-Instruct model were optimized using DeepSpeed ZeRO 2~\citep{rasley2020deepspeed} for faster training. All models were trained on 2 NVIDIA A100 GPUs, except for certain runs that were conducted on an additional L40 gpu. For the optimizer, we used AdamW \citep{loshchilov2017fixing} and paged AdamW \citep{dettmers2024qlora} optimizers with learning rates that were linearly warmed up with a cosine-scheduled decay. For both datasets, we filter for prompt and response pairs that are $<$ 1024 tokens after tokenization. This allows the policies enough context for coherent generation. Apart from keeping compute requirement reasonable, this avoids degeneration during inference since we force the model to only generate upto 256 new tokens not including the prompt length in the maximum token length.  

For our DRDO approach, we sweep over $\alpha \in \{.1, 1\}$ and $\gamma \in \{0,1,2, 5\}$ but found the most optimal combination to be  $\alpha$ = 0.1 and  $\gamma$ = 2, since a higher $\gamma$ tends to destabilize training due to the larger penalties induced on DRDO loss. This is consistent with optimal $\gamma$ values found in the literature, albeit for different tasks~\citep{yi2020focal, lin2018focallossdenseobject}. For Oracle trained for DRDO, we use the same batch size as for policy training with a slightly larger learning rate of 1e-5 and train for epoch. For consistency, we use a maximum length of 1024 tokens after filtering for pairs with prompt and responses < 1024 tokens. For all SFT training, we use the TRL library\footnote{\url{https://huggingface.co/docs/trl/en/sft_trainer}} with a learning rate of 1e-5 with a cosine scheduler and 100 warmup steps. 

For the DPO baselines, we found the implementation in DPO Trainer\footnote{\url{https://huggingface.co/docs/trl/main/en/dpo_trainer} to be the  most stable and build off most of our DRDO training pipline and configuration files based on their trainer.} For optimal parameter selection, we sweep over $\beta \in \{.1, 0.5, 1, 10\}$ but we found the default value of  $\beta$ = 0.1 to be most optimal based on the validation sets during training. Also, we found the default learning rate of 5e-6 to be optimal after validation runs. For e-DPO, we restrict the number of reward ensembles to 3 but use the same Oracle training hyperparameters mentioned above.

Table~\ref{tab:hyperparameters} provides a full list of model configurations and hyperparameters used during trainng of DRDO models.


\section{Ablation Studies}

\subsection{Ablations on reward distillation and contrastive log-unlikelihood}
\label{app:ablation-r-c}

For a more robust evaluation using a much more high-capacity oracle (GPT-4o), we randomly sampled 40 prompts from the CNN/Daily Mail test set and compute win-rates and reward margins of samples generated with a top-$p$ of 0.8 and $T=0.7$ for all baselines vs. DRDO policies trained on Reddit TL;DR. We use the Phi-3-Mini-4K-Instruct model for this experiment. We include the IPO baseline~\citep{azar2024general} with $\beta = 0.1$ (or $\tau$ in their paper), the baselines without the distillation (shown as DRDO (-R)), and without the contrastive component (shown as DRDO (-C)). Additionally, we include the baseline where reward distillation term in DRDO is replaced by the DPO loss but keep the contrastive log-unlikelihood component (shown as DPO (+C)). Table~\ref{tab:comparison_metrics_gpt4o_oracle} shows results of this experiment. Note that due to compute constraints, we only train these policies from the SFT checkpoint for 500 steps with an effective batch size of 128. For the reward estimates using GPT-4o, we only add one additional condition to the prompt shown in Fig. 5 to get scalar rewards (between 0 and 1). Fig.~\ref{fig:post_submission_gpt_oracle_ratings} provides the prompt format used for this experiment.

\begin{table}[ht]
    \centering
    \small
    {
    \begin{tabular}{lcccccc}
        \toprule
        \textbf{Comparison} & \textbf{WR A (\%)} & \textbf{WR B (\%)} & \textbf{Reward A} & \textbf{Reward B} & \textbf{Margin A} & \textbf{Margin B} \\
        \midrule
        DRDO vs. DRDO (-R) & 85.0 & 15.0 & 0.24$_{\pm 0.26}$ & 0.07$_{\pm 0.08}$ & 0.22$_{\pm 0.22}$ & 0.09$_{\pm 0.04}$ \\
        DRDO vs. DRDO (-C) & 90.0 & 10.0 & 0.25$_{\pm 0.23}$ & 0.05$_{\pm 0.07}$ & 0.23$_{\pm 0.22}$ & 0.06$_{\pm 0.03}$ \\
        DRDO vs. IPO       & 65.0 & 35.0 & 0.21$_{\pm 0.17}$ & 0.21$_{\pm 0.24}$ & 0.09$_{\pm 0.08}$ & 0.16$_{\pm 0.16}$ \\
        DRDO vs. DPO (+C)  & 80.0 & 20.0 & 0.18$_{\pm 0.16}$ & 0.14$_{\pm 0.15}$ & 0.11$_{\pm 0.08}$ & 0.22$_{\pm 0.21}$ \\
        \bottomrule
    \end{tabular}}
    \caption{Full DRDO policies (denoted ``A'') compared against various baselines (denoted ``B''), including DRDO without reward distillation and DRDO without contrastive log-unlikelihood loss. Win-rates (WR) are computed using average of reward comparison for each sample and then averaged. Margins are computed using the difference of rewards.}
    \label{tab:comparison_metrics_gpt4o_oracle}
\end{table}

We see that in all cases, full DRDO is the clear winner in terms of win rate. We also see that in this sample the reward distillation component contributes slightly more to the overall performance than the contrastive log-unlikelihood loss, but given the small sample size this is not a significant difference; DRDO's performance can be attributed to a combination of both. This is reinforced by DRDO's 80-20 performance against contrastive log-unlikelihood combined with DPO loss instead of DRDO reward distillation.

\subsection{Extent of Out-of-distribution data}

In order to comprehensively evaluate DRDO's performance against baseline methods under increasing out-of-distribution (OOD) conditions, we randomly sample 1,000 prompts from the CNN/DailyMail dataset and segment these prompts into bins of 50 tokens based on prompt-token counts, spanning the full range of token lengths. Since the CNN/DailyMail dataset represents a previously unseen input distribution, this evaluation effectively measures the OOD generalization capabilities of the policies as prompt lengths (and corresponding news article lengths) increase. For this automatic evaluation, we use GPT-4o as a high-capacity judge, consistent with Sec.~\ref{sec:exp} (prompt used is shown in Fig.~\ref{fig:prompt-reddit}). For response sampling, we use top-$p$ of 0.8 and temperature of 0.7 for DRDO and all baselines. The trends in Fig.~\ref{fig:win_rate_comparison_tldr_postsubmission} suggest that as OOD composition (with prompt-token lengths as proxy) increases, on average DRDO policies tend to have relatively larger win-rates compared to shorter prompts over baselines like DPO and e-DPO. In contrast, we find that DPO and e-DPO win-rates tend to decrease with increase in prompt-lengths.

\section{Win-Rate Evaluation Prompt Formats}
\label{app:eval-prompts}

\Figref{fig:prompt-reddit} and \Figref{fig:prompt-cnn} show the prompt format used for GPT-4o evaluation of policy generations compared to human summaries provided in the evaluation data of Reddit TL;DR (CNN Daily Articles). \Figref{fig:prompt-reddit} specifically provides the human-written summaries as reference in GPT's evaluation of the baselines.  In contrast, \Figref{fig:prompt-cnn} shows the prompt that was used to  evaluate policy generated summaries in direct comparison to human-written summaries. Note that in both prompts, we swap order of provided summaries to avoid any positional bias in GPT-4o's automatic evaluation. 

\begin{figure}[ht]
    \centering
    \begin{subfigure}[b]{0.7\textwidth}
        \centering
        \includegraphics[width=\textwidth]{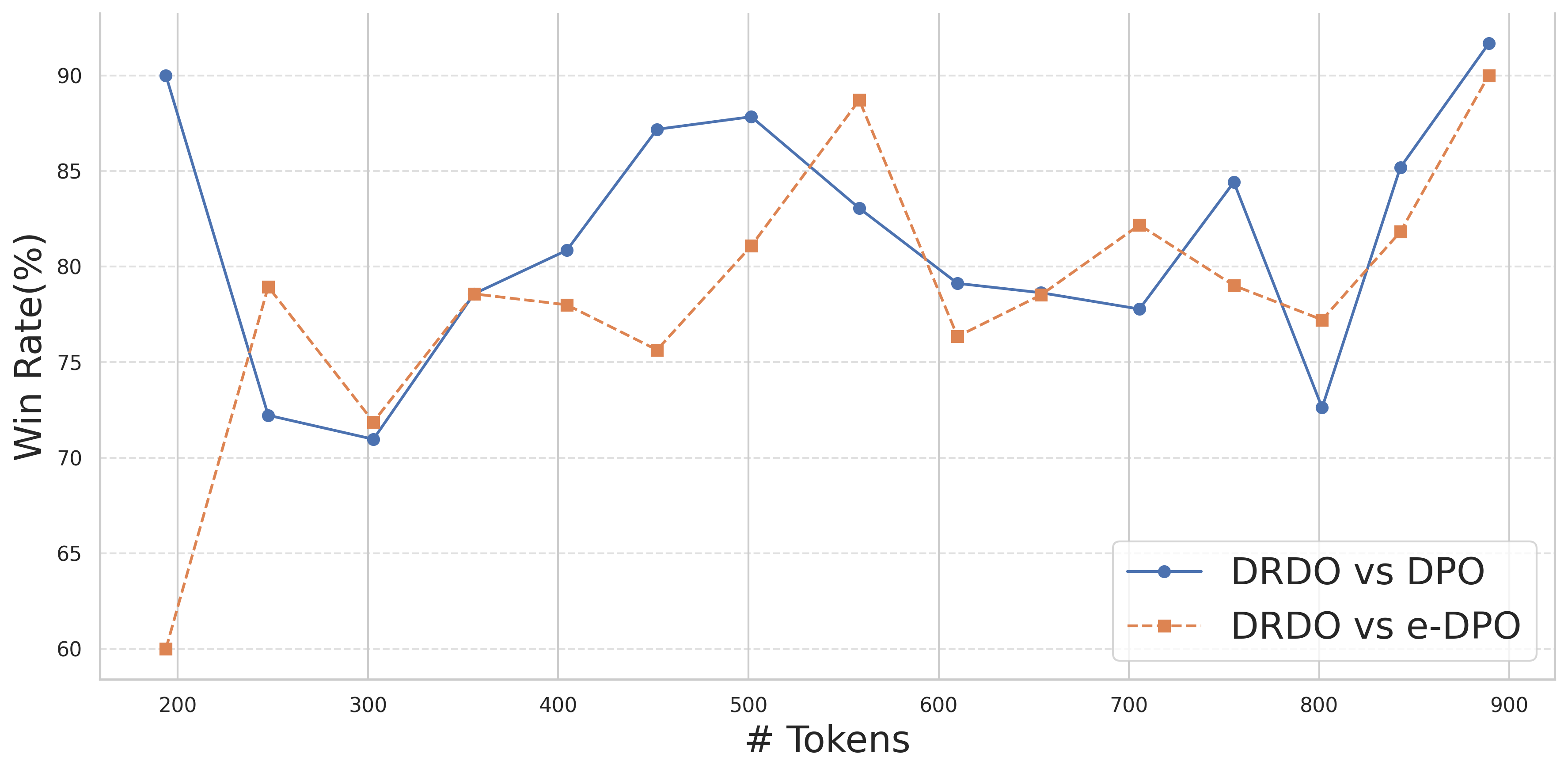}
        
        \label{fig:win_rate_tldr_post_submission}
    \end{subfigure}
  
    \begin{subfigure}[b]{0.7\textwidth}
        \centering
        \includegraphics[width=\textwidth]{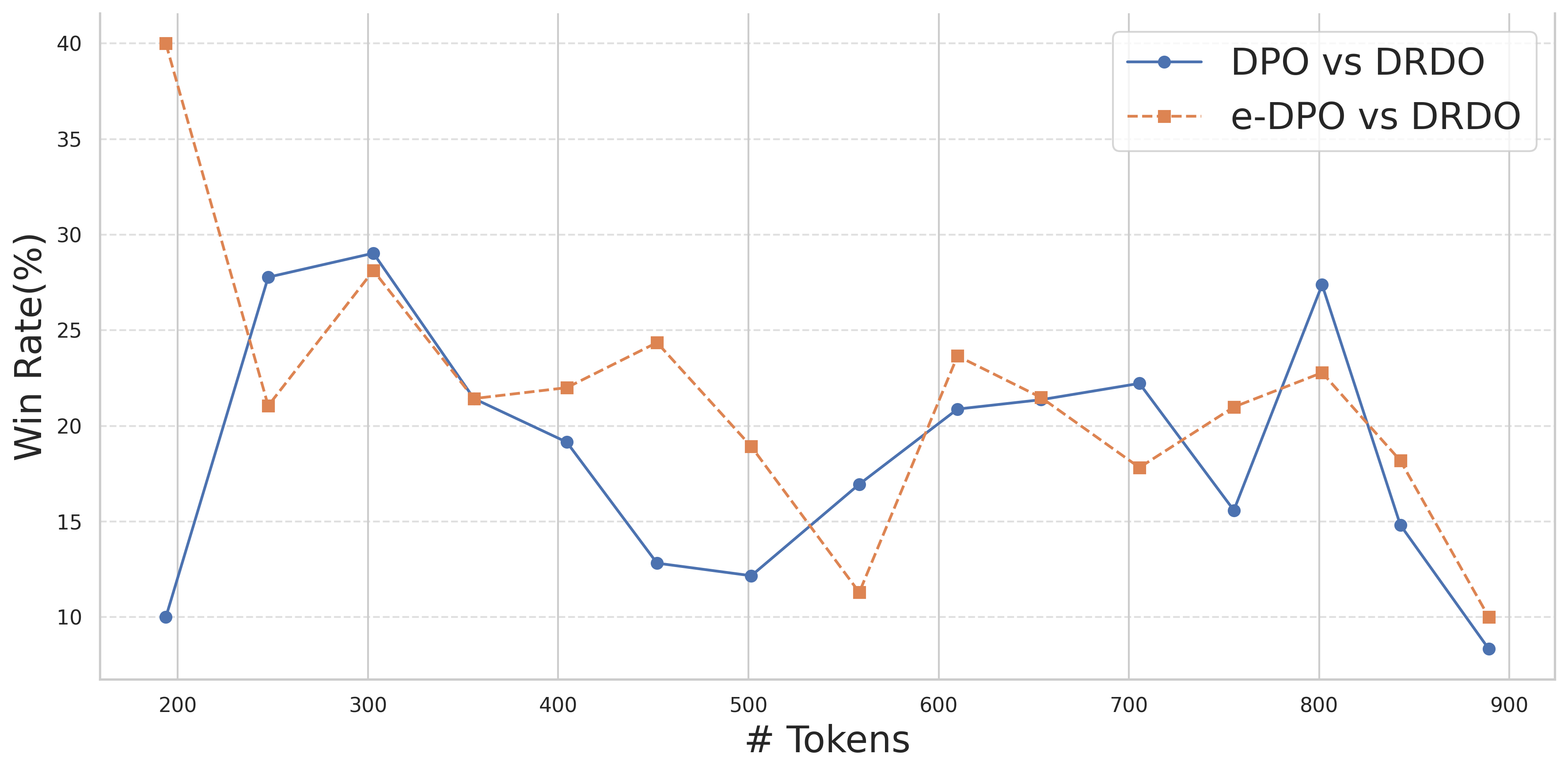}
        
        \label{fig:win_rate_tldr_post_submission_reverse}
    \end{subfigure}
    \caption{Comparison of win-rates as a function of the extent of out-of-distribution (OOD) data on the CNN daily article dataset. Win-rates (y-axis) of DRDO vs DPO and e-DPO (top) and competitor win-rates (bottom) are plotted against the increasing prompt lengths (number of tokens) over 1000 randomly sampled prompts for evaluation. DRDO is more robust to OOD settings, on average, compared to baselines like DPO and e-DPO as seen in the upward trend in win-rates over prompt-tokens.}
    \label{fig:win_rate_comparison_tldr_postsubmission}
\end{figure}

\begin{figure}[h!]
    \centering
    \includegraphics[width=\linewidth]{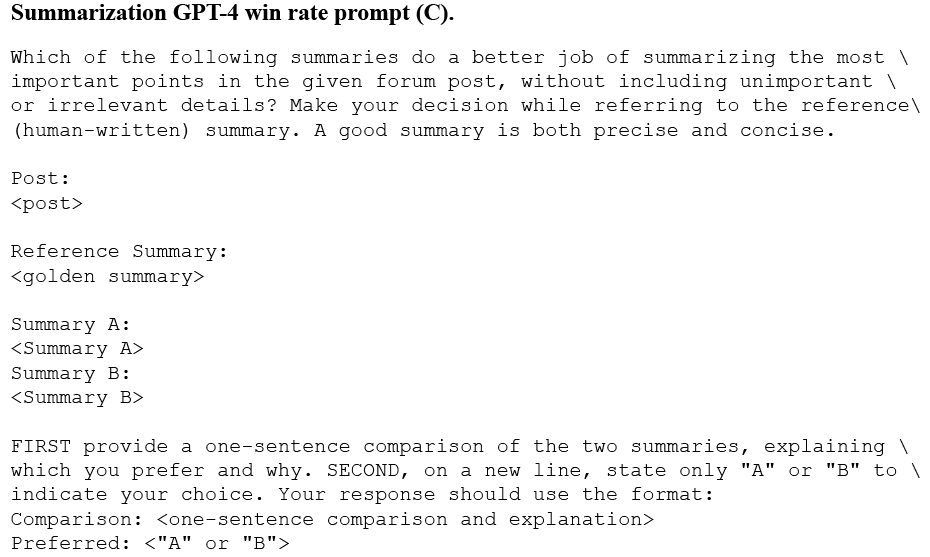}
    \caption{Prompt format for Reddit TL;DR (CNN/Daily Mail)}
    \label{fig:prompt-reddit}
\end{figure}

\begin{figure}[h!]
    \centering
    \includegraphics[width=\linewidth]{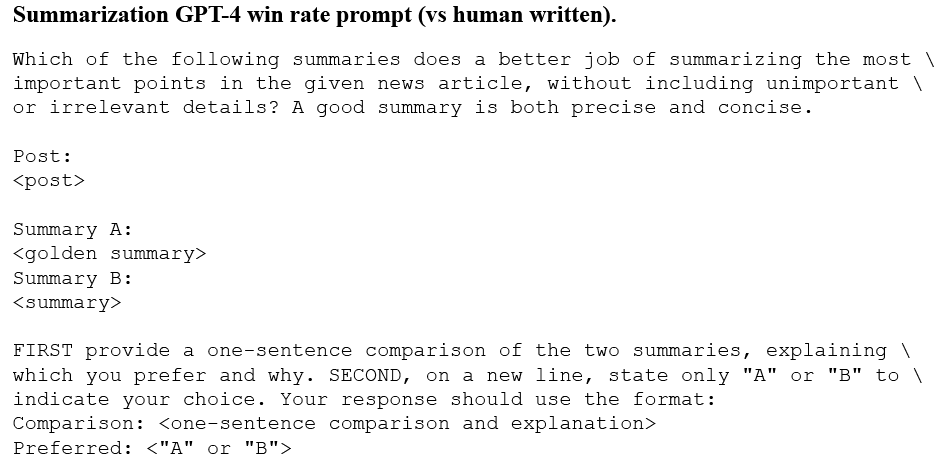}
    \caption{Prompt format for Reddit TL;DR (CNN/Daily Mail)}
    \label{fig:prompt-cnn}
\end{figure}

\begin{figure}[ht]
    \centering
    \fbox{%
        \begin{minipage}{0.9\linewidth} 
            \textbf{Summarization GPT-4o win rate prompt (C).}

            \vspace{0.2cm} 
            Which of the following summaries do a better job of summarizing the most important points in the given forum post, without including unimportant or irrelevant details? Make your decision while referring to the reference (human-written) summary. A good summary is both precise and concise.

            \vspace{0.3cm} 
            \textbf{Post:} \\
            \texttt{<post>}

            \vspace{0.3cm} 
            \textbf{Reference Summary:} \\
            \texttt{<golden summary>}

            \vspace{0.3cm} 
            \textbf{Summary A:} \\
            \texttt{<Summary A>} \\
            \textbf{Summary B:} \\
            \texttt{<Summary B>}

            \vspace{0.3cm} 
            FIRST, provide a one-sentence comparison of the two summaries, explaining which you prefer and why.\\ 
            SECOND, on a new line, state only \textbf{"A"} or \textbf{"B"} to indicate your choice. \\
            THIRD, on a new line, provide your ratings (a real reward score between 0 to 1 where 1 is highest and 0 is lowest in quality) for the summaries.\\
            Your response should use the format: \\
            \texttt{Comparison: <one-sentence comparison and explanation>} \\
            \texttt{Preferred: <"A" or "B">}\\
            \texttt{Score for Summary A: <score>}\\
            \texttt{Score for Summary B: <score>}
        \end{minipage}%
    }
    \caption{Prompt format for Reddit TL;DR.}
    \label{fig:post_submission_gpt_oracle_ratings}
\end{figure}

\section{Computational Efficiency}

e-DPO requires training reward ensembles to form a confidence set for training the policy. In our experiments, we use 3 reward models to construct this set which makes it roughly thrice as expensive as DRDO training. Like DPO, e-DPO requires a separate reference model to be kept in memory, further increasing compute requirements. DRDO, on the other hand, only requires a trained oracle for distillation. The expected oracle rewards can be precomputed once and a separate reference model does not need to be kept in memory (as shown in \Eqref{eq:distillation}). During training, DRDO does require one additional linear head on top of the base LM to predict the reward estimates. This adds a negligible 0.003\% more trainable parameters (relative to the language modeling head of base LM Phi-3-Mini-4K-Instruct). During inference, DRDO trained policies do not require this head.

\section{DRDO vs. Pluralistic Preferences}
In certain circumstances, non-deterministic preferences, as reflected in low labeler confidence or equal rewards, could be a consequence of innate pluralistic tendencies of human preferences. However, DRDO is not motivated directly by pluralistic preferences, where there are multiple annotations (or preferences) for a single $(x, y_1, y_2)$, but by the diversity of preference strength for paired samples. Typically, pluralistic approaches require multiple reward models, such as reward soups~\cite{rame2023rewardedsoupsparetooptimalalignment}, e-DPO~\cite{fisch2024robustpreferenceoptimizationreward}, MaxMin-RLHF~\cite{chakrabortymaxmin}, or conditioned policy~\cite{wang2024conditional} to model such preferences. This is computationally expensive and \textit{assumes rewards over multiple dimensions can be linearly interpolated}. We do not make any such assumptions. Our main argument as stated in Sec.~\ref{sec:motivation} is that non-deterministic preferences likely constitute a non-trivial amount of paired samples in popular preference datasets and as such, DRDO provides an efficient alignment method under such conditions. Our only strong assumption in the modeling is that the Oracle reward model, given sufficient data, should reasonably approximate human preferences using any standard reward-modeling approach. Furthermore, given such an Oracle, we directly regress on the rewards and, unlike e-DPO, do not need to find additional optimal parameters like $\beta$ in the regression or confidence set in policies or reward model ensembles. Thus, our DRDO approach does not need to learn a variety of models each unique to specific viewpoints expressed in the data, and thus our results that best the competitor baselines reflect that we are able to fit better to non-deterministic preferences while still maintaining an ability to fit to deterministic preferences and the data distribution at large.

\section{Sensitivity of DRDO's $\gamma$ vs. DPO's $\beta$ w.r.t KL-divergence from SFT model}
\label{app:ablation_kl_sensitivity}
We ran an additional experiment to compare the sensitivity of model-specific hyperparameters (DRDO's $\gamma$ vs DPO's $\beta$). Keeping $\alpha$ as 0.1 for all DRDO policies, we compute the KL-divergence during training on sampled generations on 40 randomly sampled evaluation prompts in the held-out set of Ultrafeedback with top-p of 0.8 and temperature of 0.7 with various $\gamma$ values in DRDO ($\alpha = 0.1$) and with different KL-$\beta$ values in DPO using the SFT-trained Phi-3-Mini-4K-Instruct model. Table~\ref{tab:kl_divergence_dpo_drdo_new} shows expected KL-divergence (averaged over tokens) over the 40 completions at every 100 steps of training. The expected reward accuracies (win-rates) over the SFT model completions with the same hyperparameters over these samples (after 400 training steps) are shown in Table~\ref{tab:expected_rewards_kl} below.}

These results suggest that while DRDO does not explicitly regularize its policy w.r.t. reference-model based KL-regularization, it still outperforms DPO in oracle-assigned expected reward accuracies (win-rates) on sampled generations as long as the $\gamma$ parameter is carefully chosen. In particular, as previously observed in~\citet{meng2024simposimplepreferenceoptimization} and \citet{rafailov2024direct}, smaller $\beta$ in DPO tends to increase KL-divergence with respect to the baseline SFT model. However, a relatively larger KL divergence in DRDO on average does not necessarily impede preference learning but larger $\gamma$ values tend to degrade expected rewards.

\begin{table}[h!]
    \centering
    \small
    {
    \begin{tabular}{cccccc}
        \toprule
        \textbf{Step} & \textbf{DRDO ($\gamma = 5$)} & \textbf{DRDO ($\gamma = 2$)} & \textbf{DRDO ($\gamma = 1$)} & \textbf{DPO ($\beta = 0.01$)} & \textbf{DPO ($\beta = 0.1$)} \\
        \midrule
        100 & 0.63 & 0.44 & 0.82 & 0.64 & 0.38 \\
        200 & 0.35 & 1.51 & 1.67 & 0.81 & 0.39 \\
        300 & 0.59 & 1.67 & 1.82 & 1.17 & 0.42 \\
        400 & 0.64 & 1.63 & 1.71 & 1.35 & 0.44 \\
        \bottomrule
    \end{tabular}}
    
    \caption{KL-divergence during training on sampled generations on 40 randomly sampled evaluation prompts in the held-out set of Ultrafeedback with top-p of 0.8 and temperature of 0.7 with various $\gamma$ values in DRDO ($\alpha = 0.1$) and with different KL-$\beta$ values in DPO using the Phi-3-Mini-4K-Instruct model.}
    \label{tab:kl_divergence_dpo_drdo_new}
\end{table}

\begin{table}[h!]
    \centering
  {
    \begin{tabular}{lc}
        \toprule
        \textbf{Model} & \textbf{Expected Oracle Reward} \\
        \midrule
        DPO ($\beta = 0.1$)  & 0.775 ($\pm$ 0.42) \\
        DPO ($\beta = 0.01$) & 0.675 ($\pm$ 0.47) \\
        DRDO ($\gamma = 1$)  & 0.750 ($\pm$ 0.44) \\
        DRDO ($\gamma = 2$)  & 0.825 ($\pm$ 0.38) \\
        DRDO ($\gamma = 5$)  & 0.600 ($\pm$ 0.50) \\
        \bottomrule
    \end{tabular}}

      \caption{DRDO vs DPO expected reward accuracies (win-rates) over the SFT-model completions computed using the OPT 1.3B oracle model.}
          \label{tab:expected_rewards_kl}
\end{table}

As for the exponential parameter $\gamma$, $\gamma$ = 2 appears to be a reasonable choice, as previously found in in~\citet{lin2018focallossdenseobject, yi2020focal} in the focal loss literature. A larger  $\gamma$ can harshly penalize the loss when the policy is uncertain ($p_w <<1$ ) while a smaller $\gamma = 0$ may not adequately penalize and impact its adaptive nature. In our experiments including the above experiment, we find that the optimal reward is achieved for $\gamma$ = 2 while too low or too high a $\gamma$ can affect performance as seen in Tab.~\ref{tab:expected_rewards_kl} . Note that, although we find $\gamma$ = 2 to be optimal across datasets, a reasonable way to find the right $\gamma$ would vary case by case–--if the baseline policy at the start of alignment training has not undergone or in off-policy settings, a lower $\gamma$ could be ideal since a higher $\gamma$ might apply harsher penalties in this case. However, if the policy is initialized with SFT model (as in DRDO) or in on-policy (where $p_w$ is likely to be higher already) alignment settings, a higher $\gamma$ could be optimal.  In practice though, empirical validation on a held-out set can be an efficient alternative, similar to how an optimal $\beta$ can be determined in algorithms like DPO.


\end{document}